\documentclass{article}

\usepackage{geometry}
\usepackage[utf8]{inputenc} 
\usepackage[T1]{fontenc}    
\usepackage[hidelinks]{hyperref}       
\usepackage{url}            
\usepackage{booktabs}       
\usepackage{amsfonts}       
\usepackage{nicefrac}       
\usepackage{microtype}      
\usepackage{xcolor}         
\usepackage[round]{natbib}
\usepackage{multirow}

\hypersetup{
    colorlinks=true,
    citecolor=blue
}

\usepackage[toc,page,header]{appendix}
\usepackage{minitoc}

\usepackage{amsmath,bm,dsfont,amssymb,amsthm,mathtools}
\usepackage{algorithm}
\usepackage{algorithmicx}
\usepackage{algpseudocode}

\usepackage{bigints}

\newcommand{\I}{\mathcal{I}}
\newcommand{\J}{\mathcal{I}^c}
\newcommand{\R}{\mathbb{R}}
\newcommand{\E}{\mathbb{E}}
\newcommand{\Prob}{\mathbb{P}}
\newcommand{\thetabf}{\bm{\theta}}
\newcommand{\varthetabf}{\bm{\vartheta}}

\newcommand{\MGF}{\mathbb{M}}
\newcommand{\MGFto}{\xrightarrow{\MGF}}

\DeclareMathOperator*{\argmax}{argmax}
\DeclareMathOperator*{\argmin}{argmin}

\usepackage{enumitem}

\usepackage{framed}

\usepackage{thm-restate}

\newtheorem{theorem}{Theorem}
\newtheorem{proposition}[theorem]{Proposition}
\newtheorem{lemma}[theorem]{Lemma}
\newtheorem{corollary}[theorem]{Corollary}
\newtheorem{definition}{Definition}

\newtheorem{remark}{Remark}
\newtheorem{example}{Example}

\newcommand*{\email}[1]{\texttt{#1}}
\newcommand*\samethanks[1][\value{footnote}]{\footnotemark[#1]}

\usepackage{authblk}

\title{Information-Directed Selection for Top-Two Algorithms}

\author[1]{Wei You\thanks{Equal contributions: \email{weiyou@ust.hk} and \email{cq2199@columbia.edu}}\thanks{Accepted for presentation at the Conference on Learning Theory (COLT) 2023}}
\author[2]{Chao Qin\samethanks[1]}
\author[1]{Zihao Wang}
\author[1]{Shuoguang Yang}
\affil[1]{Department of IEDA, The Hong Kong University of Science and Technology}
\affil[2]{Columbia Business School, Columbia University}

\begin{document}

\doparttoc 
\faketableofcontents 

\maketitle

\begin{abstract}
We consider the best-$k$-arm identification problem for multi-armed bandits,
where the objective is to select the \textit{exact} set of $k$ arms with the highest mean rewards by sequentially allocating measurement effort. 
We characterize the necessary and sufficient conditions for the optimal allocation using dual variables.
Remarkably these optimality conditions lead to the extension of \emph{top-two algorithm} design principle \citep{russo_simple_2020}, initially proposed for best-arm identification. 
Furthermore, our optimality conditions induce a simple and effective selection rule dubbed \emph{information-directed selection} (IDS) that selects one of the top-two candidates based on a measure of information gain.
As a theoretical guarantee, we prove that integrated with IDS, \emph{top-two Thompson sampling} is (asymptotically) optimal for Gaussian best-arm identification, 
solving a glaring open problem in the pure exploration literature \citep{russo_simple_2020}.
As a by-product, we show that for $k > 1$, top-two algorithms cannot achieve optimality even when the algorithm has access to the unknown ``optimal'' tuning parameter.
Numerical experiments show the superior performance of the proposed top-two algorithms with IDS and considerable improvement compared with algorithms without adaptive selection.
\end{abstract}


\section{Introduction}

This paper studies the best-$k$-arm identification problem in a stochastic multi-arm bandit with $K$ arms.
The goal is to identify precisely the set of $k$ arms with the largest mean by sequentially allocating measurement effort.
It serves as a fundamental model with rich applications. 
For example, in the COVID-19 vaccine race, the goal is to identify a set of best-performing vaccines as fast and accurately as possible and push them to mass production.
The best-performing alternatives are determined based on costly sampling in this type of application. 
Algorithms that require fewer samples to reach the desired confidence level or accuracy guarantee are invaluable.

We focus on the simple top-two algorithm design principle \citep{russo_simple_2020}, originally mainly proposed for the best-arm identification (BAI) problem with $k = 1$. 
A top-two algorithm follows a two-step procedure to measure an arm at each time step. 
It first identifies a pair of top-two candidates, usually dubbed as the \textit{leader} and the \textit{challenger}, and then flips a potentially biased coin to decide which of the candidates to sample.
We refer to the second step as the \textit{selection step}. 
Various top-two algorithms have been proposed for BAI, including top-two Thompson sampling (TTTS) \citep{russo_simple_2020}, top-two expected improvement (TTEI) \citep{Qin2017} and top-two transportation cost (T3C) \citep{Shang2019}.
The literature on top-two algorithms predominately focuses on studying the first step of determining the top-two candidates, whereas the second step of selecting among the top-two candidates is primarily simplified.
Existing top-two algorithms mostly require a tuning parameter $\beta$ and sample the leader with a fixed probability $\beta$.
Despite enjoying the good empirical performance, top-two algorithms' theoretical analyses are usually tailored to a weaker notion called $\beta$-optimality, namely, achieving the optimal problem complexity under the additional constraint that the best arm receives a $\beta$ proportion of the sampling effort.
It is worth mentioning that \cite{russo_simple_2020} shows the robustness of choosing $\beta = 1/2$, showing that the problem complexity for a subset of algorithms that asymptotically allocate $1/2$ proportion to the best arm is at most two times (achievable) that of the (optimal) problem complexity. 
Existing literature on top-two algorithms further proposes adaptive procedures to adjust the tuning parameter $\beta$ by solving the instance complexity optimization problem with plug-in mean estimators.
We refer to this procedure of adjusting $\beta$ as adaptive \textit{$\beta$-tuning}.

Moreover, although we can extend the top-two algorithms (designed for BAI) to tackle best-$k$-arm identification, we provide an example showing that for $k>1$, top-two algorithms can fail to achieve the optimality even if the value of $\beta$ is set to the \emph{unknown} optimal value, let alone with the potential generalization of the aforementioned adaptive $\beta$-tuning procedures for $k>1$.
Indeed, we present a structural analysis of best-$k$-arm identification and show that $k>1$ is surprisingly much more complicated than $k = 1$.
Consequently, the optimality conditions widely used to design BAI algorithms, e.g., those in \cite{GarivierK16,chen2023balancing}, are no longer sufficient for $k >1$, so how to optimally select among the top-two candidates remains open.

\paragraph{Our contributions.}
We reformulate KKT conditions of the instance complexity optimization problem to characterize asymptotic optimality. The crucial feature of our approach is the inclusion of \emph{dual variables} in the optimality conditions, which allows us to overcome the challenges due to the much more complex optimality conditions for $k>1$.
Based on the \textit{complementary slackness conditions}, we provide a novel interpretation of top-two design principle, which leads us to extending the existing top-two algorithms and designing new ones.
We propose an adaptive selection rule dubbed \emph{information-directed selection} (IDS) that wisely selects among the top-two candidates based on the \textit{stationarity conditions}. 
We show that integrated with IDS, \emph{top-two Thompson sampling} is optimal for Gaussian BAI, which solves a glaring open problem in \citet{russo_simple_2020}.
A key feature of IDS is its adaptivity to the proposed top-two candidates.
This differs from adaptive $\beta$-tuning procedures, which use the same value of $\beta$ regardless of the proposed candidates, even though it may be updated over time.
As a by-product, we show that (surprisingly) top-two algorithms with adaptive $\beta$-tuning cannot achieve the notion of $\beta$-optimality for $k > 1$.
Finally, we demonstrate the superior performance of top-two algorithms with IDS in extensive numerical experiments.

\paragraph{Related work.}
Various algorithms for BAI have been developed, including Track-and-Stop \citep{GarivierK16} and the aforementioned TTTS \citep{russo_simple_2020}, TTEI \citep{Qin2017} and T3C \citep{Shang2019}.
Algorithms dedicated to the best-$k$-arm identification problem includes OCBA-m \citep{chen2008efficient}, OCBAss \citep{gao2015note}, LUCB \citep{Kalyanakrishnan2012}, 
KL-LUCB \citep{kaufmann2013information}, LUCB++ \citep{simchowitz2017simulator},
SAR \citep{bubeck2013multiple}, and unified gap-based exploration \citep{gabillon2012best}.

Some algorithms can be adapted to the best-$k$-arm identification problem, including gradient-ascent-based
\citep{menard2019gradient}, elimination-based \citep{tirinzoni2022elimination}, and algorithms tailored to linear bandits
\citep{reda2021top,reda2021dealing}.
Specifically, \cite{reda2021top} proposes several statistically efficient Gap-Index Focused Algorithms (GIFA), which indeed share a similar spirit with the top-two design principle. Due to the combinatorial structure, most of the proposed algorithms in \cite{reda2021top} are not computationally efficient, while we focus instead on computationally efficient top-two algorithms.
\cite{Degenne2019} and \cite{degenne2019non} consider general pure exploration problems and propose meta-algorithms that require oracle solvers.
Specifically, \cite{Degenne2019} necessitates an optimal allocation solver, while \cite{degenne2019non} takes a game theoretical view of pure exploration and requires a best response oracle algorithm for the game's players.
\cite{wang2021fast} proposes Frank-Wolfe-based Sampling (FWS), an algorithm that can be adapted to a wide range of pure exploration problems.
Our approach is considerably different in the following aspects:
1. Each step of FWS requires solving a time-consuming linear programming;
2. We utilize the detailed structure of the optimality conditions to design intuitive algorithms, while FWS treats it as a generic maximin problem; and
3. Our algorithms can be easily implemented, while FWS requires characterization of the $r$-subdifferential subspace, where $r$ is a tuning parameter.

\section{Preliminary}
\paragraph{Problem formulation.} We study the \emph{best-$k$-arm identification} problem a stochastic multi-armed bandit with $[K]\triangleq\{1,\ldots,K\}$ arms. The goal of a decision maker is to confidently identify among these $K$ arms the best $k$ arms where $k\in[K]$ is fixed and known to her.
At each timestep $t\in\mathbb{N}_0\triangleq\{0,1,\ldots\}$, the decision maker selects an arm $I_t\in [K]$, and observes a reward $Y_{t+1, I_t}$. 
The rewards from arm $i\in [K]$ are independent and identically distributed random variables drawn from a distribution $p(\cdot|\theta_i)$ where $\theta_i\in\mathbb{R}$, and rewards from different arms are mutually independent. 
Throughout the paper, let $p(\cdot|\theta_i)$ be a one-parameter exponential family, parameterized by its mean $\theta_i$. A problem instance is denoted as $\bm{\theta}= (\theta_1,\ldots,\theta_K)\in\mathbb{R}^K$, which is  \emph{fixed but unknown} to the decision maker and can only be estimated through the so-called \emph{bandit feedback} in the sense that only the rewards associated with the selected arms are observed. In this paper, we focus on problem instances with a \emph{unique} set of the best $k$ arms,
\begin{equation}
	\I({\bm{\theta}}) \triangleq \argmax_{\mathcal{A} \subseteq [K], |\mathcal A| = k}\left\{\sum_{i \in \mathcal A} \theta_i\right\} ,\label{eq:empirical_best}
\end{equation}
where $|\mathcal{A}|$ is the cardinality of $\mathcal A$. 
The problem space consisting of all such problem instances is
\[
\Theta \triangleq \left\{\bm\theta\in\mathbb{R}^K \,:\, \min_{i\in \I(\bm{\theta})}\theta_i > \max_{j\in \I^c(\bm{\theta})}\theta_j \right\}.
\]
We refer to arms in $\I(\bm\theta)$ as the top arms and arms in $\I^c(\bm{\theta}) = [K]\backslash \I(\bm\theta)$ as the bottom arms.

\paragraph{Generic top-two algorithm.}
This paper focuses on top-two algorithms with adaptive selection in the form of Algorithm \ref{alg:top_two}.
At each time step, the algorithm first identifies a pair of candidates and then flips a biased coin to decide which candidate to sample. 

\begin{algorithm}
	\caption{Top-two algorithm}
	\label{alg:top_two}
	\begin{algorithmic}[1]
		\renewcommand{\algorithmicrequire}{\textbf{Input:}}
		\Require{Sampling sub-routine and tuning sub-routine.}
		\For{$t = 0, 1, \dots$}
		\State{Run sampling sub-routine to propose top-two candidates $\left(I_t^{(1)}, I_t^{(2)}\right)$.} 
		\State{Run selection sub-routine obtain selection probability $h_t$.} 
		\State{Play arm $I_t  =
			\begin{cases}
				I_t^{(1)} & \text{with probability } h_t,\\
				I_t^{(2)}  & \text{with probability } 1-h_t.
			\end{cases}$
		}
		\State{Observe reward $Y_{t+1,I_t}$ and update mean parameters.} 
		\EndFor
	\end{algorithmic}
\end{algorithm}

\section{Optimality conditions for best-\texorpdfstring{$k$}{k}-arm identification}

In this section, we provide a structural analysis of the optimal allocation problem related to the problem complexity of the best-$k$-arm identification problem.
We propose KKT-based optimality conditions that directly incorporate the dual variables, which serve as a basis for our unified framework for top-two algorithms bundled with adaptive selection.

\paragraph{Problem complexity.}
In pure exploration literature, a classical setting is called \emph{fixed-confidence setting} where a confidence level $\delta$ is given to the player. Besides a \textit{sampling rule} and a \textit{decision rule}, she needs to specify a \textit{stopping rule} $\tau_\delta$ to ensure that the probability of the recommendation $\hat{\I}_{\tau_\delta}$ being incorrect is no more than $\delta$ and to simultaneously minimize the expected number of samples $\mathbb{E}_{\bm\theta}[\tau_\delta]$. An algorithm is said to be $\delta$-correct if $\mathbb{P}_{\bm\theta}(\tau_\delta <\infty,\hat{\I}_{\tau_\delta}\neq \I(\bm\theta))\leq \delta$ for any problem instance $\bm \theta\in\Theta$. Denote by $\mathcal{A}$ the class of $\delta$-correct algorithms.
We define the problem complexity of $\thetabf$ as
\begin{equation}
	\label{eq:instance-complexity}
	\kappa(\thetabf) \triangleq \inf_{ \mathtt{Alg} \in \mathcal{A}} \, \limsup_{\delta\to 0} \frac{\E_{\bm \theta}[\tau_\delta]}{\log(1/\delta)}.
\end{equation}

\paragraph{Generalized Chernoff information.} The key to our analysis is a generalization of Chernoff information, a classical quantity in hypothesis testing \citep{cover2006elements}. Let $\mathcal{S}_K\subset \R^K$ denote the probability simplex and let $\bm p \in \mathcal{S}_K$ denote a generic allocation vector, indicating the proportion of samples allocated to each arm. 
Let $d(\cdot,\cdot)$ denote the Kullback-Leibler (KL) divergence of two reward distributions.
The \emph{generalized Chernoff information} measures the strength of evidence of distinguishing $\theta_i$ and $\theta_j$, defined as\footnote{When the context is clear, we suppress $\bm \theta$ in the notation.}
\begin{equation}
	\label{eq:def_C_ij}
	C_{i,j}(\bm p) = C_{i,j}(\bm p;\bm\theta)  \triangleq p_i d(\theta_i, \bar{\theta}_{i,j}) + p_j d(\theta_j, \bar{\theta}_{i,j}), \quad \text{where} \quad \bar{\theta}_{i,j} = \bar{\theta}_{i,j}(\bm p) \triangleq \frac{p_i \theta_i + p_j \theta_j}{p_i + p_j}.
\end{equation}
Given an allocation $\bm{p}\in\mathcal{S}_K$, the minimal strength of evidence of distinguishing an arm in the top set $\I$ and another arm in the bottom set $\I^c$ can be calculated as
\[
\Gamma_{\bm\theta}(\bm p) \triangleq \min_{(i,j) \in \I\times\I^c }  C_{i,j}(\bm p).
\]
The decision maker aims to optimize this quantity over the possible allocations, i.e.,
\begin{equation}
	\label{eq:optimal_allocation}
	\Gamma^*_{\thetabf} \triangleq \max_{\bm p \in\mathcal{S}_K} \Gamma_{\bm\theta}(\bm p).
\end{equation}
The following result shows that the inverse of this value is exactly the fixed-confidence problem complexity, which is closely related to those studied in \cite{GarivierK16}.
\begin{theorem}[Problem complexity]
	\label{thm:explicit_opt}
	For best-$k$-arm identification, $\kappa(\thetabf) = (\Gamma^*_{\thetabf})^{-1}$.
\end{theorem}
We defer the detailed proof to Appendix \ref{app:fixed_confidence_LB}. 

\subsection{Structure of the optimal solution and common approaches in the literature}\label{sec:existing}

To start, we show that the optimal solution to \eqref{eq:optimal_allocation} is unique.
To the best of our knowledge, this is the first uniqueness result for best-$k$-arm identification\footnote{\cite{Degenne2019} shows that it is possible to have non-unique optimal solutions for general pure exploration problems. For example, \cite{Degenne2019} and \cite{jedra2020optimal} imply that for best-arm identification in linear bandits, there may be multiple optimal allocations. Our result indicates that this is impossible for the best-$k$-arm identification problem in unstructured bandits.}. 
We postpone the proof to Appendix \ref{app:proof_uniqueness}.
Additionally, we report two additional properties of the optimal solution, positivity and monotonicity, in Appendix \ref{sec:properties_optimal_solution}.
\begin{restatable}[Uniqueness]{lemma}{uniqueness}
\label{lm:uniqueness}
	The optimal solution to \eqref{eq:optimal_allocation} is unique.
\end{restatable}


Next, we show that common approaches for BAI cannot be easily extended to address the best-$k$-arm identification problem.
We collect additional supporting materials in Appendix \ref{app:optiaml_allocation_problem}.

\paragraph{Design principles based on optimality conditions only dependent on $\bm p$.}
The optimal allocation problem \eqref{eq:optimal_allocation} has been widely studied in BAI.
Sufficient conditions for optimality are known in the fixed-confidence setting, e.g., \citet[Lemma 4]{GarivierK16}, and in the large deviation rate formulation\footnote{There is a switch of arguments in the KL divergence under the large deviation rate formulation, but the structure of the problem remains the same, and our analysis extends.}, e.g., \citet[Theorem 1]{Glynn2004}, and \citet[Lemma 1]{chen2023balancing}. These conditions are elegant and popular because (i) they are sufficient and necessary for BAI, (ii) they involve only two types of balancing equations, i.e., balancing allocations among individual sub-optimal arms and balancing between the best arm and the sub-optimal arms; and (iii) they involve only the allocation vector $\bm p$, making it straightforward to track the balancing conditions.

Given their simplicity and popularity, it may be intriguing to consider a direct generalization of these conditions for $k>1$ and to hope that they remain sufficient, as in \citet[Theorem 2]{gao2015note} and \citet[Theorem 3]{zhang2021asymptotically}.
Unfortunately, these conditions fail to remain sufficient.

We now summarize the generalized conditions.
The first proposition is an extension of a well-known result for BAI, e.g., \citet[Proposition 8]{russo_simple_2020}. 
\begin{proposition}[A necessary condition -- information balance]
	\label{prop:information_balance}
	For general reward distributions, the optimal solution $\bm p^*$ to (\ref{eq:optimal_allocation}) satisfies
	\begin{equation}
		\label{eq:information_balance}
		\Gamma_{\bm\theta}^* 
		= \min_{j' \in \J} C_{i,j'}(\bm p^*) \quad \forall i\in\I \quad\text{and} \quad \Gamma_{\bm\theta}^* = \min_{i' \in \I} C_{i',j}(\bm p^*) \quad \forall j\in\J. 
	\end{equation}
\end{proposition}

The next necessary condition states that the optimal solution must balance the overall allocation to the top-arm and bottom-arm sets.
For ease of exposition, we present Proposition \ref{prop:overall_balance} for Gaussian bandits, which recovers the observations in \cite{GarivierK16}.
We note that our result can also be extended to the general exponential family, which extends the results in \cite{Glynn2004}; see Appendix \ref{app:overall_balance_details}.

\begin{proposition}[A necessary condition for Gaussian -- overall balance]
	\label{prop:overall_balance}
	For Gaussian bandits, the optimal solution $\bm p^*$ to (\ref{eq:optimal_allocation}) satisfies
	\begin{equation}
		\label{eq:overall_balance}
		\sum_{i\in\mathcal{I}}(p^{*}_i)^2 = \sum_{j\in\I^c}(p^{*}_j)^2.
	\end{equation}
\end{proposition}

For BAI (i.e., $k=1$), $\I = \{I^*\}$, where $I^* = I^*(\bm\theta) \triangleq \argmax_{i\in[K]}\theta_i$ is the best arm, and \eqref{eq:information_balance} reduces to $\Gamma_{\bm\theta}^* = C_{I^*,j}(\bm p^*)$ for all $j \neq I^*$.
The intuition behind the information balance property is that the optimal allocation should gather \textit{equal evidence} to distinguish any sub-optimal arms from the best arm.

However, the balancing of evidence becomes significantly more complex when $k > 1$.
The information balance \eqref{eq:information_balance} suggests that the optimal allocation should balance the statistical evidence gathered to distinguish any top arm $i$ from the most confusing bottom arm 
\[j_i \triangleq\argmin_{j'\in\J} C_{i,j'}(\bm p^*),\]
and similarly, any bottom arm from the most confusing top arm.
But, it is a priori unclear which pairs of $(i,j)$ are the most confusing under the optimal allocation; see Appendix \ref{app:optimality_condition_on_p}.

Indeed, these simplistic balancing conditions fail to be sufficient when $k > 1$; see Remark \ref{rmk:SS_overall} and Example~\ref{ex:failure_of_sufficiency}. 
Our result indicates that algorithms that rely only on overall balance is highly likely to fail.
We also discuss the complete characterization of the necessary and sufficient condition (that depends only on $\bm p$) for general reward distributions in Appendix \ref{app:optimality_condition_on_p}.
Our analysis shows that to obtain a sufficient condition, the overall balance property \eqref{eq:overall_balance} needs to be decomposed into finer balancing conditions among subgroups of arms (Propositions \ref{prop:SS_balance} and \ref{prop:general_SS}).
Unfortunately, the optimal decomposition relies on the true $\bm\theta$, which is unavailable.
Due to the \emph{combinatorial} number of scenarios, we caution that these conditions are too complicated to yield any meaningful algorithm.

\paragraph{Adaptive adjustment of the tuning parameter in top-two algorithms.}
An alternative approach to designing an optimal algorithm is to adaptively update the tuning parameter $\beta$ used in top-two algorithms, e.g., \cite{russo_simple_2020}.
For BAI, this is relatively straightforward: (1) the optimal tuning parameter $\beta^*$ equals the optimal allocation to the best arm, which can be solved using standard bisection search as in \cite{GarivierK16}; and (2) the optimal solution is uniquely determined by \eqref{eq:information_balance} and $p_{I^*} = \beta^*$.

\begin{figure}[!ht]
	\centering
	\includegraphics[width=0.6\textwidth]{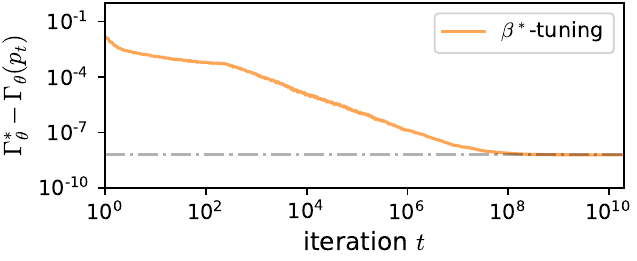}
	\caption{Gaussian bandits with $\bm\theta = (.51,.5,0,-.01,-.092)$, $\sigma=0.5$ and $k = 2$. We plot the sample path of $\Gamma_{\bm\theta}(\bm p_t)$, averaged across 100 replications, where $\bm p_t$ is the allocation under TS-KKT($0$) (Algorithm \ref{alg:TS-KKT}) with $\beta^*$-tuning, i.e., let $h_t = \beta^* \triangleq p_1^* + p_2^*$ for all $t$.  For demonstration, we let the algorithm have access to $\bm\theta$, i.e., we replace empirical/sampled mean vectors by the true $\bm \theta$. Surprisingly, it fails to achieve the optimal $\Gamma_{\bm\theta}^*$.}
	\label{fig:fail_main}
\end{figure}

However, such an approach does not work for the best-$k$-arm problem.
First, $\beta^*$ is now interpreted as the total allocation to the \textit{set} of best-$k$-arm\footnote{Assuming sufficient exploration, the top set $\I$ is accurately estimated. Then precisely one candidate will be selected from each of $\I$ and $\J$. Hence, $\beta^*$ is asymptotically the total allocation to the top set $\I$.}.
Second, solving the optimal allocation no longer reduces to a one-dimensional optimization. 
Most importantly, \eqref{eq:information_balance} and $\sum_{i \in \I}p_{i} = \beta^*$ do not guarantee the uniqueness of the solution.  
To demonstrate the failure of top-two algorithms with a fixed tuning parameter, we present a counterexample in Figure \ref{fig:fail_main}; see Example \ref{ex:failure_of_fixed_tuning} for details.
This example suggests that top-two algorithms with adaptive $\beta$-tuning (but non-adaptive to the current candidates) are not guaranteed to be optimal even if the tuning parameter $\beta$ is set to the optimal value $\beta^* = \sum_{i \in \I}p^*_{i}$.

\subsection{Necessary and sufficient conditions via KKT} \label{sec:KKT}
The major challenge in extending results from $k = 1$ to $k >1 $ lies in the significantly more complex structure in the information balance \eqref{eq:information_balance}.
To address these challenges, we propose including dual variables, which encode the structure of the optimal information balance.
We present a novel necessary and sufficient condition for optimality based on a reformulated Karush–Kuhn–Tucker (KKT) conditions for (\ref{eq:optimal_allocation}).
With a proper interpretation of the dual variables, we will see that our optimality conditions naturally induce top-two algorithms bundled with adaptive selection.

To start, note that the optimal allocation problem \eqref{eq:optimal_allocation} can be reformulated as 
\begin{subequations}
	\label{eq:reformulation}
	\begin{align}
		\Gamma^*_{\thetabf} = \max_{\phi, \bm p} 
		& \hspace{10pt} \phi \\
		\mathrm{s.t.}
		& \hspace{10pt} \sum_{i\in[K]} p_i - 1=0, \label{eq:convex_formulation_01}\\
		& \hspace{10pt} p_i \geq 0, \quad \forall i\in[K],\label{eq:convex_formulation_03}\\	
		& \hspace{10pt} \phi - C_{i,j}(\bm p)   \le 0, \quad\forall (i,j) \in  \I\times \I^c. \label{eq:convex_formulation_02}
	\end{align}
\end{subequations}
It can be verified that \eqref{eq:reformulation} is a convex optimization problem.
Slater's condition holds since $\bm p = \bm 1/K$ and $\phi = 0$ is a feasible solution such that all inequality constraints hold strictly, where $\bm 1$ is a vector of $1$'s.
Hence, the KKT conditions are necessary and sufficient for global optimality.

Let $\lambda \in \R$, $\bm \iota = \{\iota_i: i\in[K]\}$ and $\bm \mu = \{\mu_{i,j}: i \in \I, j \in \I^c\} $ be the Lagrangian multipliers corresponding to \eqref{eq:convex_formulation_01}, \eqref{eq:convex_formulation_03} and \eqref{eq:convex_formulation_02}, respectively.
Define the following \textit{selection functions}
\begin{equation}\label{eq:selection_function}
	h_{i,j}(\bm p) = h_{i,j} (\bm p; \bm\theta) \triangleq \frac{p_i d(\theta_i, \bar{\theta}_{i,j})}{C_{i,j}(\bm p)} \quad \text{and} \quad h_{j,i}(\bm p) = h_{j,i} (\bm p; \bm\theta) \triangleq \frac{p_j d(\theta_j, \bar{\theta}_{i,j})}{C_{i,j}(\bm p)}. 
\end{equation}
For Gaussian bandits, we have $h_{i,j}(\bm p) = p_{j}/(p_i + p_j)$ and $h_{j,i}(\bm p) = p_{i}/(p_i + p_j)$.
The next theorem presents our reformulated KKT conditions, which guide our algorithm design.
The key step here is to eliminate the dual variables $\lambda$ and $\bm\iota$, while leaving $\bm \mu$ as is.
\begin{theorem}[Necessary and sufficient conditions]
	\label{thm:KKT}
	A feasible solution $(\phi, \bm p)$ to (\ref{eq:reformulation}) is optimal if and only if $\phi= \min_{(i',j')\in\I\times\I^c} C_{i',j'}(\bm p)$ and there exist dual variables $\bm\mu \in \mathcal{S}_{k(K-k)}$ such that 
	\begin{subequations}
		\label{eq:KKT_equiv}
		\begin{align}
			&p_i = \sum_{j'\in\I^c} \mu_{i,j'} h_{i,j'}(\bm p),\quad\forall i\in\I,
			\quad p_j = \sum_{i'\in \I} \mu_{i',j} h_{j,i'}(\bm p),\quad\forall j\in\I^c, \quad\text{and}
			\label{eq:KKT_equiv_stationarity}\\
			&\mu_{i,j}(\phi - C_{i,j}(\bm p))= 0, \quad\forall (i,j)\in \I\times\I^c.\label{eq:KKT_equiv_complementary_slackness} 
		\end{align}
	\end{subequations}
\end{theorem}

\section{Top-two algorithms and information-directed selection}

Note that the KKT conditions consist of two parts, the stationarity conditions (\ref{eq:KKT_equiv_stationarity}) and the complementary slackness conditions \eqref{eq:KKT_equiv_complementary_slackness}.
These two sets of conditions serve distinctive roles in algorithm design.
In Section \ref{sec:top_two_CS}, we discuss the connection between \eqref{eq:KKT_equiv_complementary_slackness} and top-two algorithms.
In Section \ref{sec:selection_stationarity}, we discuss how \eqref{eq:KKT_equiv_stationarity} leads to information-directed selection in top-two algorithms.

\subsection{Top-two algorithms and complementary slackness conditions}
\label{sec:top_two_CS}

We show that complementary slackness conditions lead to top-two algorithms, including TTTS \citep{russo_simple_2020} and T3C \citep{Shang2019}.
The key here is interpreting the dual variables $\bm\mu$ as the long-run proportion of time $(i,j)$ are selected as the candidates.

Let $N_{t,i}$ denote the number of samples drawn from arm $i$ before time $t$ and define $p_{t,i} = N_{t,i}/t$. Furthermore, let $\mu_{t,i,j}$ be the proportion of time that $(i,j)$ is selected as candidates before time $t$. 
The values of $\bm p_{0}$ and $\bm \mu_{0}$ do not matter, and can be set to any values.
For top-two algorithms to be optimal, a necessary condition is that $(\bm p_{t}, \bm \mu_t)$ asymptotically solves \eqref{eq:KKT_equiv}.

Complementary slackness \eqref{eq:KKT_equiv_complementary_slackness} states that the optimal solution $(\bm p^*, \bm \mu^*)$ satisfies $\mu_{i,j}^* = 0$ if $C_{i,j}(\bm p^*) > \Gamma_{\bm\theta}^*.$
This suggests that, to design an optimal top-two algorithm, we should propose $(i,j)$ as candidates asymptotically if $C_{i,j}(\bm p_t;\bm \theta_t) = \Gamma_{\bm\theta_t}(\bm p_t)$.
This is indeed the case for many existing top-two algorithms, as the following example shows.
\begin{example}[Connection to T3C  and TTTS ]\label{ex:TTTS}
	T3C selects the first candidate by Thompson Sampling and then selects the second candidate by searching for the pair with the smallest transportation cost.
	With sufficient exploration, the first candidate coincides with the best arm with high probability, and their transportation cost is asymptotically equivalent to our $C_{i,j}(\bm p)$ here.
	As a result, the pair of candidates selected by T3C asymptotically satisfies \eqref{eq:KKT_equiv_complementary_slackness}.
	
	TTTS samples from the posterior distribution of the mean rewards until two distinct best-arm candidates are sampled. 
	One concern for TTTS is that it might take enormous samples until the challenger differs from the leader when the posterior is concentrated.
	In that case, approximations can be used to avoid repeated sampling.
	For example, we can sample second arm $I^{(2)}_t = j$ with probability proportional to $\Prob_{\tilde{\bm\theta}\sim\Pi_t}\left(\tilde{\theta}_j > \tilde{\theta}_{I^{(1)}_t}\right)$, where $\Pi_t$ is the posterior distribution of $\bm\theta$.
	This roughly equals sampling with probability proportional to $\exp\left(-t C_{t,I^{(1)}_t,j}\right)$ in the asymptotic sense, see \citet[Proposition 5]{russo_simple_2020}.
	As $t \to \infty$, this becomes a hard minimum as in \eqref{eq:KKT_equiv_complementary_slackness}.
\end{example}

\subsection{Adaptive selection and stationarity conditions}
\label{sec:selection_stationarity}

To completely specify a top-two algorithm (Algorithm \ref{alg:top_two}), we need a selection step.
Previous top-two algorithms require a non-adaptive tuning parameter $\beta$ and set $h_t = \beta$.
Recall our discussions in Section \ref{sec:existing} and Example \ref{ex:failure_of_fixed_tuning}, such algorithms are not guaranteed to converge to the optimal allocation even if $\beta$ is tuned so that it converges to the optimal $\beta^*$.

We now discuss information-directed selection based on the stationarity conditions \eqref{eq:KKT_equiv_stationarity}.
The key is to interpret the selection functions $h_{i,j}$ as the probability that arm $i$ is played given candidates $(i,j)$.
For simplicity of demonstration, we consider a top-two algorithm that guarantees sufficient exploration so that the estimation of the set of the best-$k$-arm is accurate. We further assume\footnote{For BAI, we only need $\bm{p}_\infty$ since $\bm{\mu}_\infty$ can be explicitly expressed by $\bm{p}_\infty$.} that the sample allocation $\bm p_t$ converges to some $\bm p_{\infty}$ and $\bm \mu_t$ converges to $\bm \mu_{\infty}$.
We then focus on adaptive selection such that $(\bm p_{\infty}, \bm \mu_{\infty})$ satisfies the stationarity conditions \eqref{eq:KKT_equiv_stationarity}.
Recall that $\mu_{t,i,j}$ is the proportion of time that $(i,j)$ is selected as candidates before time $t$.
By the convergence of $(\bm p_t, \bm \mu_t)$, we have 
$p_{\infty,i} = \sum_{j \in \I^c} \mu_{\infty,i,j}h_{i,j}(\bm p_{\infty}).$
Similarly, for any arm $j \in \I^c$, we have $p_{\infty,j} = \sum_{i \in \I} \mu_{\infty,i,j}h_{j,i}(\bm p_{\infty}).$
Hence, the stationarity conditions \eqref{eq:KKT_equiv_stationarity} hold asymptotically.

Consequently, we propose the information-directed selection for top-two algorithms in Algorithm \ref{alg:IDS}. 
Here, we explicitly allow $h_{i,j}$ to depend on the mean vector $\bm \theta$ because the decision maker does not know $\bm \theta$ and the selection function must be estimated using an estimate of $\bm \theta$.

\begin{algorithm}[!ht]
	\caption{Information-directed selection}
	\label{alg:IDS}
	\begin{algorithmic}[1]
		\renewcommand{\algorithmicrequire}{\textbf{Input:}}
		\Require{Sample allocation $\bm p_t$, mean estimator $\bm \theta_t$ and candidates $\left(I_t^{(1)},I_t^{(2)}\right)$.}
		\State{
			Return $h_t = h_{I_t^{(1)},I_t^{(2)}} (\bm p_t; \bm\theta_t).$  
		}
	\end{algorithmic}
\end{algorithm}

\begin{remark}[Information-directed selection]
	In interpreting our adaptive selection, we write
	\begin{equation}\label{eq:PDE}
		C_{i,j}(\bm p) = p_i\frac{\partial C_{i,j}(\bm p)}{\partial p_i} + p_j \frac{\partial C_{i,j}(\bm p)}{\partial p_j},
	\end{equation}
	which follows from \eqref{eq:def_C_ij} and the fact that $\frac{\partial C_{i,j}(\bm p)}{\partial p_i} = d(\theta_i, \bar{\theta}_{i,j})$ and $\frac{\partial C_{i,j}(\bm p)}{\partial p_j} = d(\theta_j, \bar{\theta}_{i,j})$.
	Note that $C_{i,j}(\bm p)$ measures the amount of information collected per unit sample under allocation vector $\bm p$ to assert whether $\theta_i \ge \theta_j$ or not. 
	Hence, $\frac{\partial C_{i,j}(\bm p)}{\partial p_i}$ denotes the information gain per unit allocation to arm $i$ and $p_i\frac{\partial C_{i,j}(\bm p)}{\partial p_i}$ denotes the information gain contributed by allocating $p_i$ proportion samples to arm $i$.
	The selection function $h_{i,j}$ suggests that we should select proportional to the information gain when deciding between a pair of candidates. Note that $h_{i,j}$ involves unknown oracle quantities. In the implementation, we estimate it using empirical quantities.
	
    To prevent any confusion, we note that our selection rule used in top-two algorithms (for pure exploration problems) shares the same abbreviation with the algorithm design principle proposed in \cite{russo2018learning} (for regret minimization problems), but they are not related.
 
\end{remark}

\begin{remark}[Selection versus tuning] 
	We remark that our information-directed selection is fundamentally different from the standard approach of adaptive $\beta$-tuning where the same parameter $\beta$ is used regardless of the proposed top-two candidates, even though the value of $\beta$ may be updated over time.
	Hence, we regard this method as hyperparameter tuning.
	In contrast, our information-directed selection selects from the proposed candidates using the respective selection function $h_{I_t^{(1)},I_t^{(2)}} (\bm p_t; \bm \theta_t)$.  
    Given different choices of top-two candidates, the selection functions are determined differently. 
    Hence, the selection functions cannot be regarded as the same parameter.
\end{remark}

\subsection{The TTTS algorithm for best-\texorpdfstring{$k$}{k}-arm identification}

This section presents a TTTS sub-routine for best-$k$-arm identification in Algorithm \ref{alg:TTTS-k}, whose essential idea has appeared in \cite{russo_simple_2020}.
Our TTTS algorithm, when applied with IDS (Algorithm \ref{alg:IDS}), follows the essential algorithm design principles from our KKT analysis.
As a sanity check of IDS, we prove the fixed-confidence optimality of the proposed algorithm for Gaussian BAI.  
This solves an open problem in the pure exploration literature, e.g., Section 8 of \cite{russo_simple_2020}.

We will propose several Bayesian algorithms for best-$k$-arm identification that start with a prior distribution $\Pi_0$ with density $\pi_0$ over a set of parameters $\tilde{\Theta}$ that contains $\bm\theta$. Conditional on the history $\mathcal{H}_{t} \triangleq \sigma\left(I_0, Y_{1,I_0},\ldots, I_{t-1}, Y_{t, I_{t-1}}\right)$, the posterior distribution $\Pi_t$ has density 
\[
\pi_t({\bm\vartheta}) = \frac{\pi_0({\bm\vartheta})\prod_{\ell=0}^{t-1}p(Y_{\ell+1, I_{\ell}}|{\vartheta}_{I_{\ell}})}{\int_{\tilde{\Theta}}\pi_0(\tilde{\bm\vartheta})\prod_{\ell=0}^{t-1}p(Y_{\ell+1, I_{\ell}}|\tilde{\vartheta}_{I_{\ell}})\mathrm{d}\tilde{\bm\vartheta}},   \quad \forall {\bm{\vartheta}}=(\vartheta_1,\ldots,\vartheta_K)\in\tilde{\Theta}.
\]

At each time step, our TTTS sub-routine repeatedly samples $\tilde{\bm\theta}$ from the posterior distribution $\Pi_t$ until two distinct empirical best-$k$-arm sets appear.
The top-two candidates are then selected from the symmetric difference of the two distinct best-$k$ arm sets, breaking tie arbitrarily. 
By the same argument in Remark \ref{ex:TTTS}, TTTS asymptotically follows from the complementary slackness condition \eqref{eq:KKT_equiv_complementary_slackness}, i.e., $(i,j)$ is selected as the candidates if and only if the empirical estimation $C_{i,j}(\bm p_t;\bm \theta_t)$ is the smallest in the asymptotic sense as $t \to \infty$.
\begin{algorithm}[!ht]
	\caption{TTTS (for best-$k$-arm identification) sub-routine}
	\label{alg:TTTS-k}
	\begin{algorithmic}[1]
		\renewcommand{\algorithmicrequire}{\textbf{Input:}}
		\Require{Posterior distribution $\Pi_t$.}
		\State{Sample $\tilde{\bm\theta} \sim \Pi_t$ and define $\I_t = \I(\tilde{\bm{\theta}})$}
		\State{Repeatedly sample $\tilde{\bm\theta}' \sim \Pi_t$ until $\I_t' = \I(\tilde{\bm{\theta}'})$ differs from $\I_t$}
		\State{Return any $\left(I_t^{(1)},I_t^{(2)}\right) \in (\I_t \backslash \I_t')\times(\I_t' \backslash \I_t)$.}
	\end{algorithmic}
\end{algorithm}

Combining the TTTS sub-routine (Algorithm \ref{alg:TTTS-k}) and the information-directed selection sub-routine (Algorithm \ref{alg:IDS}), we have the TTTS-IDS algorithm.

\paragraph{Stopping and recommendation rule.}
For notational convenience, we define $C_{t,i,j} \triangleq C_{i,j}(\bm p_t;\bm \theta_t)$ for any $(i,j)\in [K]\times[K]$. As shown \citet{GarivierK16}, the Chernoff stopping rule is given by
\begin{equation}\label{eq:Chernoff_stopping}
	\tau_\delta = \inf \left\{
	t: Z_t \triangleq \min_{(i, j)\in \hat{\I}_t\times\hat{\I}_t^c} t \cdot C_{t, i,j} >\gamma(t, \delta)
	\right\},
\end{equation}
where $\hat{\I}_{t} = \I(\bm\theta_t)$ and $\gamma(t, \delta)$ is a threshold determined by the decision maker.
Upon stopping, we recommend $\hat{\I}_{\tau_{\delta}} = \I(\bm\theta_{\tau_{\delta}})$. 
We postpone the discussion of \eqref{eq:Chernoff_stopping} to Appendix \ref{app:stopping}.

\paragraph{Approximate TTTS algorithms.}

Despite its simplicity, the TTTS algorithm could be computationally inefficient when the posterior concentrates due to the resampling step .
Hence, we implement two approximate versions of TTTS called TS-PPS (with PPS referring to Posterior Probability Sampling) and TS-KKT($\rho$) (which takes $\rho$ as a hyperparameter).

Here we follow and extend the naming convention in \cite{jourdan2022top} for best-$k$-arm identification. For an algorithm with name \texttt{(top\_set)-(candidates)}, \texttt{(top\_set)} is a sub-routine calculating the estimated top-set and \texttt{(candidates)} is a sub-routine proposing the top-two candidates.

\begin{algorithm}[!ht]
	\caption{TS-PPS sub-routine}
	\label{alg:ATTTS-k}
	\begin{algorithmic}[1]
		\renewcommand{\algorithmicrequire}{\textbf{Input:}}
		\Require{Posterior distribution $\Pi_t$.}
		\State{Sample $\tilde{\bm\theta} \sim \Pi_t$ and define $\I_t = \I(\tilde{\bm{\theta}})$}
		\State{Sample and return $\left(I_t^{(1)}, I_t^{(2)}\right) = (i,j)\in \I_t \times \I_t^c$ with probability proportional to $\Prob_{\tilde{\bm\theta}\sim\Pi_t}\left(\tilde\theta_j > \tilde\theta_i\right)$.}
	\end{algorithmic}
\end{algorithm}

\begin{algorithm}[!ht]
	\caption{TS-KKT($\rho$) sub-routine}
	\label{alg:TS-KKT}
	\begin{algorithmic}[1]
		\renewcommand{\algorithmicrequire}{\textbf{Input:}}
		\Require{Posterior distribution $\Pi_t$, mean estimator $\bm\theta_t$, and hyperparameter $\rho$.}
		\State{Sample $\tilde{\bm\theta} \sim \Pi_t$ and define $\I_t = \I(\tilde{\bm{\theta}})$}
		\State{Return $\left(I_t^{(1)}, I_t^{(2)}\right) \in \arg\min_{(i, j)\in \I_t\times\I_t^c} \left(C_{t, i,j} - \frac{\rho}{t} \log(1/N_{t,i} + 1/N_{t,j})\right)$.} 
	\end{algorithmic}
\end{algorithm}

These two algorithms are closely related to TTTS. Essentially they recover the complementary-slackness conditions in the asymptotic sense. 
Given any $i \in \I_t$, for TTTS to sample a different arm $j\in \I_t^{c}$ into $\I_t'$, we need $\tilde{\theta}_{j} > \tilde{\theta}_{i} $, which occurs with probability $\Prob_{\tilde{\bm\theta}\sim\Pi_t}\left(\tilde\theta_j > \tilde\theta_i\right)$.
This is the motivation behind TS-PPS.
We note that $\Prob_{\tilde{\bm\theta}\sim\Pi_t}\left(\tilde\theta_j > \tilde\theta_i\right)$ is asymptotically equivalent to $\exp(-tC_{t,i,j})$ by \citet[Proposition 5]{russo_simple_2020}. As $t \to \infty$, this becomes a hard minimum in \eqref{eq:KKT_equiv_complementary_slackness}. TS-PPS samples the top-two candidates proportional to posterior probabilities, naturally encouraging exploration.
The TS-KKT($\rho$) algorithm includes an additional term to encourage measuring under-sampled pairs of candidates to reduce the persistent effect of rare events at early stages. The design of proposing the top-two candidates in TS-KKT($\rho$) is inspired by the challenger used in TTEI \citep{Qin2017}, and shares a similar spirit with the Transportation Cost Improved (TCI) challenger proposed in \citet{jourdan2022top}.

\subsection{Optimality for Gaussian best-arm identification}

As a theoretical guarantee, we show that TTTS with IDS achieves the optimality in the fixed-confidence setting for Gaussian BAI.

\begin{theorem}[Fixed-confidence optimality]
	\label{thm:main}
	Consider any problem instance $\bm{\theta}$ with unique arm means. Let prior distribution be uninformative and independent across arms. Let $\delta\in(0,1)$ and $\alpha>1$.  
	Using the Chernoff stopping rule \eqref{eq:Chernoff_stopping} with\footnote{The constant $C = C(\alpha, K)$ is given in \citet[Proposition 12]{GarivierK16}.} threshold $\gamma(t, \delta) = \log\left(Ct^\alpha/\delta\right)$, for TTTS integrated with IDS, 
	\[
	\limsup_{\delta\to0} \,\,\frac{ \E_{\bm{\theta}}[\tau_\delta]}{\log(1/\delta)} \leq \frac{1}{\Gamma^*_{\thetabf}}.
	\]
\end{theorem}
By Theorems \ref{thm:main} and \ref{thm:explicit_opt}, TTTS with IDS achieves the problem complexity defined in Equation \eqref{eq:instance-complexity}. It also implies the optimal rate of posterior convergence studied in \citet{russo_simple_2020}.

\begin{remark}
	The analysis of TTTS might be the most involved among the top-two algorithms since its leader and challenger are both random (given the history). Our proof strategy also works for other top-two algorithms coupled with IDS, e.g., TS-PPS, TS-KKT($\rho$) and those in \citet{russo_simple_2020, Qin2017, Shang2019, jourdan2022top}.
\end{remark}

\section{Numerical experiments}

In this section, we numerically validate the convergence to optimal allocation for $k > 1$, demonstrate the advantage of IDS over $\beta$-tuning, and present extensive comparisons with existing algorithms\footnote{\url{https://github.com/zihaophys/topk_colt23}}.

For numerical experiments, we consider the problem instances listed below; some of them are also used in \cite{russo_simple_2020}. For cases 1 to 4, we consider Bernoulli bandits. For cases 1 to 6, we consider Gaussian bandits with variance $\sigma^2 = 1$. We adopt an uninformative prior for our algorithms.
\begin{table}[!ht]
	\centering
	\resizebox{\columnwidth}{!}{
		\begin{tabular}{lll|lll}
			\toprule
			ID &  $K,k$ & $\bm\theta$ &  ID &  $K,k$ & $\bm\theta$ \\ \midrule
			1 &  $5,2$ & $(0.1, 0.2, 0.3, 0.4, 0.5)$ & 4 &  $50,25$ & $\theta_i = 1, \forall i \in [25]; \theta_i = 0.5, \forall i \in [40]\backslash[25]; \theta_i = 0, \forall i\in[50]\backslash [40]$ \\ 
			2 & $15,1$ & $(0.3,\dots,  0.3, 0.5)$ & 5 &  $10,3$ & $(-1.9, -0.6, -0.5, -0.4, -0.3, -0.1, 0, 0.1, 0.4, 1.8)$\\
			3 & $100,5$ & $\theta_i = 0.3,\forall i\in [95];\theta_i = 0.7,\forall i\in [100]\backslash[95]$ & 6 &  $10,5$ & $(-1.9, -0.6, -0.5, -0.4, -0.3, -0.1, 0, 0.1, 0.4, 1.8)$\\
			\bottomrule
		\end{tabular}
	}
\end{table}

\begin{figure}[!ht]
	\centering
	\includegraphics[width = 0.24\textwidth]{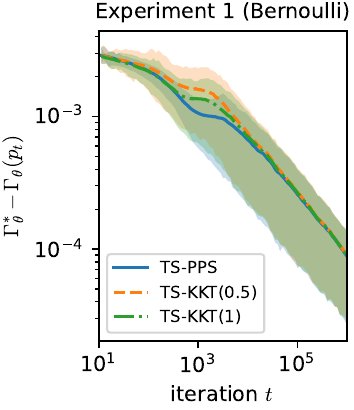}
	\includegraphics[width = 0.24\textwidth]{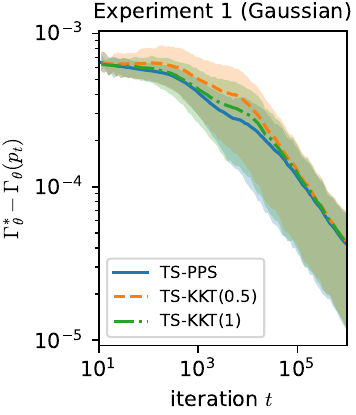}
	\includegraphics[width = 0.24\textwidth]{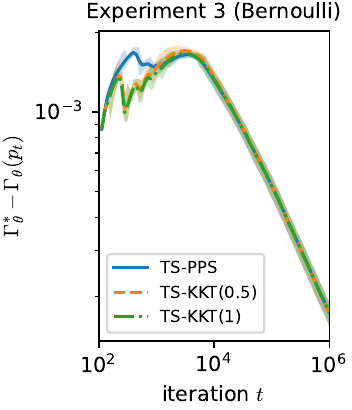}
	\includegraphics[width = 0.24\textwidth]{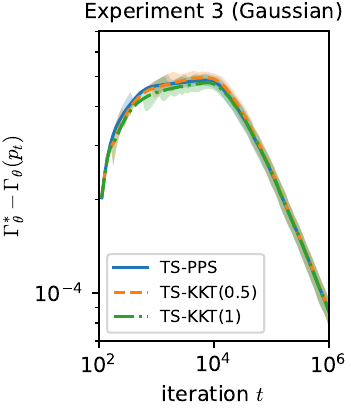}
	\caption{The optimality gap for TS-PPS and TS-KKT($\rho$), averaged across 1000 replications. The shaded areas indicate the first and third quartiles.}\label{fig:conv}
\end{figure}

\paragraph{Convergence to the optimal allocation.}
We numerically validate that $\Gamma_{\bm\theta}(\bm p_t)$ under TS-PPS and TS-KKT($\rho$) converges to $\Gamma_{\bm \theta}^*$.
Figure \ref{fig:conv} reports the optimality gap $\Gamma_{\bm \theta}^* - \Gamma_{\bm\theta}(\bm p_t)$ in logarithm scale under cases 1B, 1G, 3B, and 3G where B and G refer to Bernoulli and Gaussian, respectively.

\paragraph{Advantage of IDS over $\beta$-tuning.}
To isolate the effect of the selection rule, we fix the sampling subroutine to TS-PPS in Algorithm \ref{alg:TS-KKT} and compare the selection subroutine IDS in Algorithm \ref{alg:IDS} with $0.5$-tuning, i.e., set $\beta = 0.5$.
We consider the slippage configuration where the mean $\bm\theta$ is given by $\theta_{i} = 0.75, i \in [k]$ and $\theta_{j} = 0.5, j \notin [k]$ for both Gaussian ($\sigma^2 = 1$) and Bernoulli bandits. We label the instances by $(K,k)$-distribution. For all fixed confidence experiments, we adopt the Chernoff stopping rule \eqref{eq:Chernoff_stopping} with a heuristic threshold $\gamma(t,\delta) = \log \left((\log (t) + 1)/\delta\right)$.
Table \ref{tab:num_IDS_vs_tuning} shows a clear advantage of IDS in terms of fixed-confidence problem complexity.
\begin{table}[!htb]
	\scriptsize
	\centering
	\resizebox{\columnwidth}{!}{
	\begin{tabular}{
			lccccccccc}
		\toprule
		$(K,k)$-dist & $(500,1)$-G & $(500,2)$-G & $(500,3)$-G & $(500,5)$-G  &  $(500,1)$-B & $(500,2)$-B & $(500,3)$-B & $(500,5)$-B  \\ 
		\midrule
		TS-PPS-IDS & 199374 & 216668  & 223860 & 241869 &  44640 & 47044 & 50436 & 52767 \\
		TS-PPS-$0.5$ & 285823 & 285591  & 283401 & 283180  &  62596 & 61914 & 62660 & 62734  \\
		\midrule
		Increase & 43.4\% & 31.8\% & 26.6\% & 17.0\% & 40.2\% & 31.6\% & 24.2\% & 18.9\% \\
		\bottomrule
	\end{tabular}
 }
	\caption{Fixed-confidence problem complexity with $\delta = 0.001$. Averaged over 100 replications.}\label{tab:num_IDS_vs_tuning}
\end{table}

\paragraph{Fixed-confidence performance.}
We now compare our proposed algorithms with existing algorithms, which we briefly summarize in Appendix \ref{app:existing_algs}.
We use the $m$-LinGapE and MisLid algorithms implemented by \cite{tirinzoni2022elimination}, which are designed for linear bandits and hence not implemented for Bernoulli bandits.
To make fair comparisons, we choose the exploration rate as $\beta({t, \delta}) = \log \left((\log (t) + 1)/\delta\right)$ in UGapE and KL-LUCB, coupled with their own stopping rules. Table \ref{tab:fixed_confidence} reports the fixed confidence problem complexity, which shows the superior performance of our proposed algorithms. 
Comparing TS-KKT($0.5$) and TS-KKT($1$) provides intuitions for selecting the tuning parameter $\rho$. In particular, the larger the $\rho$, the more exploration is encouraged. More exploration helps correct the persistent effect of rare samples (e.g., in Example 3G, $\delta = 0.001$) at early stages but will slightly increase the average problem complexity.
The FWS algorithm achieves a comparable performance at the cost of a much heavier computational complexity.

\begin{table}[!htb]
	\centering
	\resizebox{\columnwidth}{!}{
		\begin{tabular}{@{}ll|rrrrrrrrrrrrrr@{}}\toprule
			Case   & $\delta$  & TS-KKT(0.5) & TS-KKT(1) &  TS-PPS & FWS   & $m$-LinGapE & MisLid  & LMA      & KL-LUCB & UGapE  & Uniform \\ 
			\midrule 
			
			& 0.1       & \textbf{786}   & 816  & 892  & 879  & -- & -- & 1128 & 1653 & 1638 & 1254 \\
			1 (Bernoulli)   & 0.01      & \textbf{1289}  & 1301 & 1398 & 1359 & -- & -- & 1673 & 2646 & 2651 & 2266  \\
			& 0.001     & 1881  & \textbf{1826} & 1863 & 1871 & -- & -- & 2209 & 3594 & 3586 & 3324  \\ 
			\midrule                   
			
			& 0.1       &  1146 & 1237 & 1410 & \textbf{1102} & -- & -- & 1895 & 2161 & 2199 & 2437  \\
			2 (Bernoulli)   & 0.01      & 1791  & 1844 & 2057 & \textbf{1735} & -- & -- & 2647 & 3398 & 3550 & 3599 \\
			& 0.001     &  2408 & 2452 & 2636 & \textbf{2340} & -- & -- & 3435 & 4678 & 5020 & 4618 \\
			\midrule                   
			
			& 0.1       & \textbf{2301}  & 2401 & 3106 & 2380 & -- & -- & 6816 & 3393 & 4110 & 8238   \\
			3 (Bernoulli)   & 0.01      & \textbf{3120}  & 3244 & 4060 & 3343 & -- & -- & 9046 & 5365 & 6615 & 10688    \\
			& 0.001     & \textbf{4108}  & 4232 & 5065 & 4378 & -- & -- & 11169 & 7400 & 9160 & 12931  \\
			\midrule  
			
			& 0.1       &  \textbf{3699} & 3747 & 3928  & 3763  & 4394  & 4395 & 4864 & 7547  & 7641 & 5885   \\
			1 (Gaussian)    & 0.01      & \textbf{5772}  & 6095 & 6159  & 5977 & 6732 & 6861 & 7246 & 11865 & 11791  & 9800   \\
			& 0.001     &  \textbf{8206} & 8441 & 8554  & 8219 & 9004 & 9525 & 9975 & 16152 & 16350 & 14574    \\
			\midrule  
			
			& 0.1       & \textbf{5082}  & 5359  & 6008 & 5406 & 7337 & 8471 & 7614 & *12295  & *11084 & 10876  \\
			2 (Gaussian)    & 0.01      & \textbf{7634}  & 7901  & 8604 & 8028 & 10422 & 12407 & 10922 & 14548 & 15404 & 15367  \\
			& 0.001     & \textbf{10257} & 10513 & 11217 & 10607 & 13452 & 15807 & 14235 & 19688 & 21436 & 19975  \\
			\midrule  
			
			& 0.1       & 10671 & \textbf{10189} & 12820 & 14789 & 17757 & 22308 & 22342 & *65185 & *58544 & 35139  \\
			3 (Gaussian)    & 0.01      & \textbf{13645} & 13820 & 16977 & 18518 & 23394 & 28672 & 29471 & *23694 & *28765 & 45507 \\
			& 0.001     & 18654 & \textbf{17766} & 21068 & 22884 & 28926 & 34696 & 35882 & 30825 & 38126 & 55412  \\
			\midrule
			
			& 0.1       & \textbf{3680}  & 3786 & 4735 & 4911 & 5419 & 7594 & 7105 & *14682 & *19918 & 11223   \\
			4 (Gaussian)    & 0.01      & \textbf{4974}  & 4981 & 6265 & 6321 & 7156 & 9998 & 9598 & *9657 & *11708 & 14603   \\
			& 0.001     & \textbf{6366}  & 6530 & 7844 & 7563 & 8739 & 12096 & 11860 & 11502 & 11852 & 17753   \\
			\midrule  
			
			& 0.1       & 3580  & 3618  & 3764 & \textbf{3546}  & 4225  & 4724  & 6344 & 7203  & 7283 & 11818  \\
			5 (Gaussian)    & 0.01      & \textbf{5770}  & 5820  & 6049 & 5807  & 6412 & 6993 & 8906 & 11653 & 11691 & 20670 \\
			& 0.001     & 7884  & 8029  & 8073 & \textbf{7802} & 8506 & 9582 & 11372 & 15692 & 15840 & 30608  \\
			\midrule  
			
			& 0.1       & \textbf{1267}  & 1271 & 1393  & 1294 & 1631 & 1933 & 1968 & 2332 & 2345  & 3490  \\
			6 (Gaussian)    & 0.01      & 1943  & 1988 & 2114 & \textbf{1921} & 2362 & 2796 & 2817 & 3667 & 3699 & 5477 \\
			& 0.001     & \textbf{2636}  & 2707 & 2813 & 2652 & 3119 & 3590 & 3663 & 4950 & 5022 & 7830  \\
			\midrule\midrule
			
			& $(K,k)$       & \multicolumn{9}{c}{Execution time per iteration in milliseconds}  \\ 
			\midrule
			1 (Gaussian)  & (5,2)       & 5.5   &  5.4 & 4.7    & 293.8 & 13.0        & 15.8   & 13.8 &  2.1      & 2.2   & 0.82 \\
			6 (Gaussian)  & (10,5)      & 18.0  & 17.8 & 13.4   & 306.1 & 43.2        & 41.5   & 35.5 &  2.6      & 3.9   & 0.99 \\
			3 (Gaussian)  & (100,5)     & 311.2  &  314.0 & 202.3  & 13782 & 4203        & 3392   & 1563 &  5.6      & 125.7 & 2.59 \\
			\bottomrule
		\end{tabular}
	}
	\caption{Fixed-confidence problem complexity, averaged over 1000 replications. We marked with $*$ the cases where the algorithm is not stopped in $10^6$ steps for some replications, in which cases the problem complexity is recorded as $10^6$.}\label{tab:fixed_confidence}
\end{table}

\section{Conclusion}

We provide optimality conditions for best-$k$-arm identification via KKT analysis and show that such conditions naturally lead to top-two algorithms coupled with the proposed information-directed selection rule that determines how to select from top-two candidates. As a theoretical guarantee,
we prove that TTTS with IDS is optimal in the fixed-confidence setting for Gaussian BAI, solving an open problem in the pure exploration literature \citep{russo_simple_2020}.
Numerical experiments demonstrate convergence to the optimal allocation under top-two algorithms with IDS for best-$k$-arm identification, a considerable improvement over algorithms without adaptive selection and leading performance in the fixed-confidence setting.
In future works, we plan to establish optimality for the case of $k > 1$.
Furthermore, it would be interesting to apply our KKT analysis and algorithm design to broader pure exploration problems in unstructured and structured bandits that satisfy the assumptions in \cite{wang2021fast}.

\clearpage

\section*{Acknowledgments}

Wei You received support from the Hong Kong Research Grants Council [ECS Grant 26212320 and GRF Grant 16212823]. Shuoguang Yang received support from the Hong Kong Research Grants Council [ECS Grant 26209422].

\bibliography{refs_TopK}
\bibliographystyle{plainnat}

\clearpage
\appendix

\addcontentsline{toc}{section}{Appendix} 

\part{Appendix} 

\parttoc 

\clearpage


\newpage

\section{Proof of Theorem \ref{thm:main}}\label{app:TTTS_optimality}

\subsection{Preliminary}
\subsubsection{Notation}
For Gaussian BAI with reward distributions $N(\theta_i,\sigma^2)$ for $i\in[K]$,
the unique best arm is denoted by $I^* = I^*(\bm\theta) = \argmax_{i\in[K]}\theta_i$.  
The generalized Chernoff information can be calculated as follows,
\begin{equation}\label{eq:def_C_ij_Gaussian}
	C_{i,j}(\bm p) = \frac{\left(\theta_{i} - \theta_j\right)^2}{2\sigma^2\left(p_{i}^{-1} + p_j^{-1}\right)},
	\quad \forall j\neq i.
\end{equation}
and 
the selection functions specialize to
\begin{equation} \label{eq:hij_Gaussian}
	h_{i,j} = h_{i,j}(\bm p) \triangleq \frac{p_{i}^{-1}}{p_{i}^{-1} + p_{j}^{-1}} = \frac{p_{j}}{p_{i} + p_{j}}, \quad \forall j \neq i.
\end{equation}
The empirical versions of $C_{i,j}(\cdot)$ and $h_{i,j}$ can be obtained by plugging in sample proportion $\bm p_t$ and mean estimators $\bm \theta_t$.

\paragraph{Minimum and maximum gaps.} We define the minimum and maximum values between the expected rewards of two different arms:
\[
\Delta_{\min} \triangleq \min_{i\neq j} \left\lvert\theta_i-\theta_j \right\rvert \quad\text{and}\quad \Delta_{\max} \triangleq \max_{i,j\in [K]} \left(\theta_i-\theta_j \right).
\] 
By the assumption that the arm means are unique,
we have $\Delta_{\max}\geq\Delta_{\min} > 0$.

\paragraph{Another measure of cumulative effort and average effort.}
We have introduced one measure of cumulative effort and average effort: the number of samples and the proportion of total samples allocated allocated to arm $i\in [K]$ before time $t\in\mathbb{N}_0$ are denoted as
\[
N_{t,i} = \sum_{\ell = 0}^{t-1} \mathds{1}(I_{\ell} = i) 
\quad\text{and}\quad
p_{t,i} = \frac{N_{t,i}}{t}.
\]
To analyze randomized algorithms (such as TTTS), 
we introduce the probability of sampling arm $i$ at time $t$ as 
\[
\psi_{t,i} \triangleq \mathbb{P}(I_t = i |  \mathcal{H}_{t}),
\]
where the history $\mathcal{H}_{t}$ is the sigma-algebra generated by the observations $\left(I_0, Y_{1,I_0},\ldots, I_{t-1}, Y_{t, I_{t-1}}\right)$ available up to time $t$. Then we define another measure of cumulative effort and average effort:
\[
\Psi_{t,i} \triangleq \sum_{\ell=0}^{t-1} \psi_{\ell, i}
\quad\text{and}\quad
w_{t,i}\triangleq \frac{\Psi_{t,i}}{t}.
\]

\paragraph{Posterior means and variances under independent uninformative priors.}
Under the independent uninformative prior, the posterior mean and variance of the unknown mean of arm $i\in[K]$ are
\[
\theta_{t,i} = \frac{1}{N_{t,i}}\sum_{\ell = 0}^{t-1} \mathds 1(I_{\ell} = i)Y_{\ell+1,I_{\ell}}
\quad\text{and}\quad
\sigma_{t,i}^2 = \frac{\sigma^2}{N_{t,i}}.
\]

\subsubsection{Definition and properties of strong convergence}
The proof of our Theorem \ref{thm:main} requires bounding time until quantities reaches their limits. 
Following \cite{qin2022electronic}, we first introduce a class of random variables that are  ``light-tailed".
Let the space $\mathcal{L}^p$ consist of all measurable $X$ with $\| X\|_p <\infty$ where $\|  X \|_{p} \triangleq \left( \E\left[ | X|^p \right]   \right)^{1/p}$ is the $p$-norm. 

\begin{definition}[Definition 1 in \cite{qin2022electronic}]
	\label{def:MGF}
	For a real valued random variable $X$, we say $X\in \MGF$ if and only if $\| X\|_p <\infty$ for all $p\geq 1$. Equivalently, $\MGF=\cap_{p\geq 1}  \mathcal{L}_p$. 
\end{definition}

Then we introduce the so-called \emph{strong convergence} for random variables, which is a stronger notion of convergence than almost sure convergence.

\begin{definition}[Definition 2 in \cite{qin2022electronic}]\label{def: strong convergence}
	For a sequence of real valued random variables $\{X_t\}_{t\in \mathbb{N}_0}$ and a scalar $x\in \mathbb{R}$,
	we say $X_t \MGFto x$ if 
	\[
	\text{for all } \epsilon>0 \, \text{there exists } \, T \in \MGF \, \text{ such that for all } t\geq T,\,\,   |X_t - x|\leq \epsilon.
	\]
	We say $X_t \MGFto \infty$ if
	\[
	\text{for all } c>0 \, \text{there exists } \, T \in \MGF \, \text{ such that for all } t\geq T,\,\,   X_t \geq c,
	\]
	and similarly, we say $X_t \MGFto -\infty$ if $-X_t \MGFto \infty$. \\
	For a sequence of random vectors $\{\bm X_t\}_{t\in \mathbb{N}_0}$ taking values in $\mathbb{R}^d$ and a vector $\bm x\in \mathbb{R}^d$ where $\bm X_t\triangleq (X_{t,1},\ldots,X_{t,d})$ and $\bm x = (x_1,\ldots,x_d)$, we say $\bm X_t \MGFto \bm x$ if $ X_{t,i} \MGFto x_{i}$ for all $i\in [d]$.
\end{definition}

\citet{qin2022electronic} also introduces a number of properties of the class $\MGF$ (Definition \ref{def:MGF}) and strong convergence (Definition \ref{def: strong convergence}).

\begin{lemma}[Lemma 1 in \citet{qin2022electronic}: Closedness of $\MGF$]
	\label{lem: closedness of MGF}
	Let $X$ and $Y$ be non-negative random variables in $\MGF$.
	\begin{enumerate}
		\item $aX+bY\in \MGF$ for any scalars $a,b\in \mathbb{R}$.
		\item $X Y \in \MGF$.
		\item $\max\{X,Y\} \in \MGF$. 
		\item $X^q \in \MGF$ for any $q\geq 0$.
		\item If $g: \mathbb{R} \to \mathbb{R}$ satisfies $\sup_{x\in -[c,c]} |g(x)| < \infty$ for all $c\geq 0$ and $|g(x)| = O(|x|^q)$ as $|x|\to \infty$ for some $q\geq 0$, then $g(X)\in \MGF$. 
		\item If $g: \mathbb{R} \to \mathbb{R}$ is continuous and $|g(x)| = O(|x|^q)$ as $|x|\to \infty$ for some $q\geq 0$, then $g(X)\in \MGF$.
	\end{enumerate}
\end{lemma}

As mentioned in \citet{qin2022electronic}, many of the above properties can be extended to any finite collection of random variables in $\MGF$. A notable one is that if $X_i \in \MGF$  for each $i\in [d]$ then
\[
X_1+\cdots + X_d \in \MGF \quad \text{and} \quad  \max\{X_1, \ldots, X_d \} \in \MGF.
\]

\citet{qin2022electronic} also shows an equivalence between pointwise convergence and convergence in the maximum norm and a continuous mapping theorem 
under strong convergence (Definition \ref{def: strong convergence}).

\begin{lemma}[Lemma 2 in \citet{qin2022electronic}]
	\label{lem:pointwise and maximum norm convergences}
	For a sequence of random vectors $\{X_n\}_{n\in \mathbb{N}}$ taking values in $\mathbb{R}^d$ and a vector $x\in \mathbb{R}^d$,
	if $X_n \MGFto x$, then $\| X_n - x\|_{\infty} \MGFto 0$. Equivalently, if $X_{n,i}\MGFto x_i$ for all $i\in [d]$, then for all $\epsilon>0$, there exists $N\in \MGF$ such that $n\geq N$ implies $|X_{n,i}-x_i|\leq \epsilon$ for all $i\in [d]$.  
\end{lemma}

\begin{lemma}[Lemma 3 in \citet{qin2022electronic}: Continuous Mapping]
	\label{lem:continuous-mapping} 
	For any  $x$ taking values in a normed vector space and any random sequence $\{X_n \}_{n\in \mathbb{N}}$:
	\begin{enumerate}
		\item If $g$ is continuous at $x$, then $X_n\MGFto x$ implies $g(X_n) \MGFto g(x)$.
		\item  
		If the range of function $g$ belongs to $\mathbb{R}$ and $g(y)\to \infty$ as $y\to \infty$, then $X_n \MGFto \infty$ implies $g(X_n)\MGFto \infty$. 
	\end{enumerate}
\end{lemma}

\subsubsection{Maximal inequalities}
Following \citet{Qin2017,Shang2019}, we introduce the following path-dependent random variable to control the impact of observation noises:
\begin{equation}
	\label{eq:W1}
	W_1 = \sup_{(t,i)\in\mathbb{N}_0\times [K]} \sqrt{\frac{N_{t,i}+1}{\log(N_{t,i}+e)}}\frac{|\theta_{t,i}-\theta_i|}{\sigma}.
\end{equation}
In addition, to control the impact of algorithmic randomness, we introduce the other path-dependent random variable:
\begin{equation}
	\label{eq:W2}
	W_2 = \sup_{(t,i)\in\mathbb{N}_0\times [K]} \frac{|N_{t,i}-\Psi_{t,i}|}{\sqrt{(t+1)\log(t+e^2)}}.
\end{equation}
The next lemma ensures that these maximal deviations have light tails.
\begin{lemma}[Lemma 6 of \cite{Qin2017} and Lemma 4 of \cite{Shang2019}]
	\label{lem:W1 and W2}
	For any $\lambda > 0$, 
	\[
	\E[e^{\lambda W_1}] < \infty 
	\quad\text{and}\quad 
	\E[e^{\lambda W_2}] < \infty.
	\]
\end{lemma}

The definition of $W_2$ in Equation \eqref{eq:W2} and Lemma \ref{lem:W1 and W2} imply the following corollary.
\begin{corollary}
	\label{cor:W2}
	There exists $T\in \MGF$ such that\footnote{The exponent $0.6$ is not essential in the sense that it can be any constant in $(0.5,1)$. } for any $t\geq T$,
	\[
	N_{t,i} - \Psi_{t,i} \leq W_2t^{0.6}, \quad \forall  i\in[K].
	\]
	This implies
	\[
	p_{t,i}-w_{t,i}\MGFto 0, \quad \forall  i\in[K].
	\]
\end{corollary}

\subsection{Proof outline}
\paragraph{Stopping rule and $\delta$-correct guarantee.} 
When $k = 1$, our Chernoff stopping rule specializes to the one used in \cite{GarivierK16} for BAI.
The $\delta$-correct guarantee is provided in \citet[Proposition 12]{GarivierK16}.
\begin{proposition}[Proposition 12 in \cite{GarivierK16}]\label{prop:delta_correct}
	Consider the best-arm identification problem. Let $\delta\in (0, 1)$ and $\alpha > 1$. There exists a constant\footnote{The constant $C = C(\alpha, K)$ is given in the proof of \citet[Proposition 12]{GarivierK16}.} $C = C(\alpha, K)$ such that for any sampling rule, using the Chernoff stopping rule \eqref{eq:Chernoff_stopping} with threshold $\gamma(t, \delta) = \log\left(Ct^\alpha/\delta\right)$ and the recommendation rule $\hat{\I}_{\tau_{\delta}} = \I(\bm\theta_{\tau_{\delta}})$ ensures that for any $\bm\theta\in\Theta$, $\mathbb{P}_{\bm \theta}(\tau_\delta < \infty, \hat{\I}_{\tau_{\delta}}  \neq \I(\bm\theta) ) \leq \delta$.
\end{proposition}

\paragraph{Sufficient conditions for optimality in fixed-confidence setting.}

Theorem 3 in \citet{Qin2017} shows a sufficient condition of sampling rules that  ensures the optimality in fixed-confidence setting.
\begin{theorem}[Theorem 3 in \citet{Qin2017}: A sufficient condition]\label{thm:sufficient_for_optimality}
	Consider any problem instance $\bm\theta\in\Theta$. Let $\delta\in (0, 1)$ and $\alpha > 1$. Using the Chernoff stopping rule \eqref{eq:Chernoff_stopping} with threshold $\gamma(t, \delta) = \log\left(Ct^\alpha/\delta\right)$, for any sampling rule,
	\[
	\bm p_t\MGFto \bm p^*
	\quad\implies\quad
	\limsup_{\delta\to0} \,\,\frac{ \E_{\bm{\theta}}[\tau_\delta]}{\log(1/\delta)} \leq \frac{1}{\Gamma^*_{\thetabf}}.
	\]
\end{theorem}

Sometimes it is easier to study the following alternative sufficient conditions to prove optimality of sampling rules.
\begin{proposition}[Alternative sufficient conditions]
	\label{prop:ratio strong convergence}
	The following conditions are equivalent:
	\begin{enumerate}
		\item $\bm p_t\MGFto \bm p^*$;
		\item $\frac{p_{t,j}}{p_{t,I^*}} \MGFto \frac{p_{j}^*}{p_{I^*}^*}$ for any $j\neq I^*$;
		\item $\bm w_t\MGFto \bm p^*$; and
		\item $\frac{w_{t,j}}{w_{t,I^*}} \MGFto \frac{p_{j}^*}{p_{I^*}^*}$ for any $j\neq I^*$.
	\end{enumerate}
\end{proposition}
\begin{proof}
	By Corollary \ref{cor:W2}, conditions 1 and 3 are equivalent. 
	
	Now we are going to show conditions 1 and 2 are equivalent. By Lemma \ref{lem: closedness of MGF}, condition 1 implies condition 2. On the other hand,
	applying Lemma \ref{lem: closedness of MGF} to condition 3 gives
	\[
	\frac{1- p_{t,I^*}}{p_{t,I^*}} = \sum_{j\neq I^*} \frac{p_{t,j}}{p_{t,I^*}} \MGFto \sum_{j\neq I^*} \frac{p^*_{j}}{p^*_{I^*}} = \frac{1- p^*_{I^*}}{p^*_{I^*}},
	\]
	Using Lemma \ref{lem: closedness of MGF} again yields
	$
	p_{t,I^*}\MGFto p^*_{I^*},
	$
	and then condition 2 gives
	$
	p_{t,j} \MGFto p_{j}^*
	$
	for any $j\neq I^*$
	Hence, condition 2 implies condition 1, and thus they are equivalent. Similarly, conditions 2 and 4 are equivalent. This completes the proof.
\end{proof}

To prove our Theorem \ref{thm:main}, we are going to show that TTTS with IDS satisfies the last sufficient condition in Proposition \ref{prop:ratio strong convergence}:
\begin{proposition}
	\label{prop:sufficient condition}
	Under TTTS with IDS, 
	\[
	\frac{w_{t,j}}{w_{t,I^*}} \MGFto \frac{p_{j}^*}{p_{I^*}^*}, \quad \forall j\neq I^*.
	\]
\end{proposition}

The rest of this appendix is devoted to proving Proposition \ref{prop:sufficient condition}.

\subsection{Empirical overall balance under TTTS with IDS}
Previous papers \citep{Qin2017,Shang2019,jourdan2022top} establishes $\beta$-optimality by proving either $w_{t,I^*}\MGFto \beta$ or $p_{t,I^*}\MGFto \beta$ where $\beta$ is a fixed tuning parameter taken as an input to the algorithms, while
the optimal $\beta^* = $ (i.e., $p^*_{I^*}$) is unknown a priori.
In our information-directed selection rule, at each time step, $h_t$ is adaptive to the top-two candidates, so novel proof techniques are required.
In this subsection, we first show that the ``empirical'' version of overall balance \eqref{eq:overall_balance} holds.

\begin{proposition}
	\label{prop:empirical overall balance}
	Under TTTS with IDS, 
	\label{prop:overall_balance_Psi}
	\[
	w_{t,I^*}^2 - \sum_{j\neq I^*}w_{t,j}^2\MGFto 0
	\quad\text{and}\quad
	p_{t,I^*}^2 - \sum_{j\neq I^*}p_{t,j}^2\MGFto 0.
	\]
\end{proposition}

To prove this proposition, we need the following supporting result.
\begin{lemma}
	\label{lem:psi_bounds}
	Under TTTS with IDS, for any $t\in\mathbb{N}_0$,
	\[
	\alpha_{t,I^*}\left(\sum_{\ell\neq I^*} \frac{\alpha_{t,\ell}}{1-\alpha_{t,I^*}}\frac{p_{t,\ell}} {p_{t,I^*}+p_{t,\ell}}\right)
	\leq \psi_{t,I^*} \leq 
	\alpha_{t,I^*}\left(\sum_{\ell\neq I^*} \frac{\alpha_{t,\ell}}{1-\alpha_{t,I^*}}\frac{p_{t,\ell}} {p_{t,I^*}+p_{t,\ell}}\right) +  (1-\alpha_{t,I^*}),
	\]
	and for any $j\neq I^*$,
	\[
	\alpha_{t,I^*}\frac{\alpha_{t,j}}{1-\alpha_{t,I^*}} \frac{p_{t,I^*}}{p_{t,I^*}+p_{t,j}}
	\leq \psi_{t,j} \leq
	\alpha_{t,I^*}\frac{\alpha_{t,j}}{1-\alpha_{t,I^*}} \frac{p_{t,I^*}}{p_{t,I^*}+p_{t,j}} + (1-\alpha_{t,I^*}).
	\]
\end{lemma}
\begin{proof}
	Fix $t\in\mathbb{N}_0$. For a fixed $j\in [K]$ and any $\ell \neq j$, TTTS obtains the top-two candidates
	\begin{equation*}
		\left(I_t^{(1)}, I_t^{(2)}\right) = 
		\begin{cases}
			(j,\ell), & \text{with probability } \alpha_{t,j}\frac{\alpha_{t,\ell}}{1-\alpha_{t,j}},\\
			(\ell,j), & \text{with probability } \alpha_{t,\ell}\frac{\alpha_{t,j}}{1-\alpha_{t,\ell}},
		\end{cases} 
	\end{equation*}
	and when TTTS is applied with IDS,
	\begin{align*}
		\psi_{t,j} = \alpha_{t,j}\left(\sum_{\ell\neq j} \frac{\alpha_{t,\ell}}{1-\alpha_{t,j}}\frac{p_{t,\ell}} {p_{t,j}+p_{t,\ell}}\right) + \sum_{\ell\neq j}\alpha_{t,\ell}\frac{\alpha_{t,j}}{1-\alpha_{t,\ell}}\frac{p_{t,\ell}}{p_{t,j} + p_{t,\ell}}.
	\end{align*}
	
	For $j = I^*$, we have 
	\begin{align*}
		\psi_{t,I^*} 
		&= \alpha_{t,I^*}\left(\sum_{\ell\neq I^*} \frac{\alpha_{t,\ell}}{1-\alpha_{t,I^*}}\frac{p_{t,\ell}} {p_{t,I^*}+p_{t,\ell}}\right) + \sum_{\ell\neq I^*}\alpha_{t,\ell}\frac{\alpha_{t,I^*}}{1-\alpha_{t,\ell}}\frac{p_{t,\ell}}{p_{t,I^*} + p_{t,\ell}},
	\end{align*}
	so
	\begin{align*}
		\psi_{t,I^*} 
		&\geq \alpha_{t,I^*}\left(\sum_{\ell \neq I^*} \frac{\alpha_{t,\ell}}{1-\alpha_{t,I^*}}\frac{p_{t,\ell}} {p_{t,I^*}+p_{t,\ell}}\right)
	\end{align*}
	and
	\begin{align*}
		\psi_{t,I^*} 
		&\leq \alpha_{t,I^*}\left(\sum_{\ell\neq I^*} \frac{\alpha_{t,\ell}}{1-\alpha_{t,I^*}}\frac{p_{t,\ell}} {p_{t,I^*}+p_{t,\ell}}\right) + \sum_{\ell\neq I^*}\alpha_{t,\ell}\\
		&= \alpha_{t,I^*}\left(\sum_{\ell\neq I^*} \frac{\alpha_{t,\ell}}{1-\alpha_{t,I^*}}\frac{p_{t,\ell}} {p_{t,I^*}+p_{t,\ell}}\right) +  (1-\alpha_{t,I^*}),
	\end{align*}
	where the inequality uses that for any $\ell\neq I^*$, $\frac{\alpha_{t,I^*}}{1-\alpha_{t,\ell}}\frac{p_{t,\ell}}{p_{t,I^*}+p_{t,\ell}} \leq \frac{\alpha_{t,I^*}}{1-\alpha_{t,\ell}} = \frac{\alpha_{t,I^*}}{\sum_{\ell'\neq \ell}\alpha_{t,\ell'}}\leq 1$. 
	
	For $j\neq I^*$, we have
	\begin{align*}
		\psi_{t,j} &= \sum_{\ell\neq j}\alpha_{t,\ell}\frac{\alpha_{t,j}}{1-\alpha_{t,\ell}}\frac{p_{t,\ell}}{p_{t,j}+p_{t,\ell}} 
		+ \alpha_{t,j}\left(\sum_{\ell\neq j} \frac{\alpha_{t,\ell}}{1-\alpha_{t,j}}\frac{p_{t,\ell}} {p_{t,j}+p_{t,\ell}}\right)\\
		&= \alpha_{t,I^*}\frac{\alpha_{t,j}}{1-\alpha_{t,I^*}} \frac{p_{t,I^*}}{p_{t,j}+p_{t,I^*}} 
		+ \sum_{\ell\neq j,I^*}\alpha_{t,\ell}\frac{\alpha_{t,j}}{1-\alpha_{t,\ell}}\frac{p_{t,\ell}}{p_{t,\ell}+p_{t,j}} 
		+ \alpha_{t,j}\left(\sum_{\ell\neq j} \frac{\alpha_{t,\ell}}{1-\alpha_{t,j}}\frac{p_{t,\ell}} {p_{t,j}+p_{t,\ell}}\right),
	\end{align*}
	so
	\begin{align*}
		\psi_{t,j} 
		&\geq \alpha_{t,I^*}\frac{\alpha_{t,j}}{1-\alpha_{t,I^*}} \frac{p_{t,I^*}}{p_{t,j}+p_{t,I^*}}
	\end{align*}
	and 
	\begin{align*}
		\psi_{t,j} 
		&\leq   \alpha_{t,I^*}\frac{\alpha_{t,j}}{1-\alpha_{t,I^*}} \frac{p_{t,I^*}}{p_{t,j}+p_{t,I^*}}  
		+ \left(\sum_{\ell\neq j,I^*}\alpha_{t,\ell}\right) + \alpha_{t,j}   \\
		&= \alpha_{t,I^*}\frac{\alpha_{t,j}}{1-\alpha_{t,I^*}} \frac{p_{t,I^*}}{p_{t,j}+p_{t,I^*}}  + (1-\alpha_{t,I^*}),
	\end{align*}
	where the inequality uses that
	\begin{enumerate}
		\item for any $\ell\neq j,I^*$, $\frac{\alpha_{t,j}}{1-\alpha_{t,\ell}}\frac{p_{t,\ell}}{p_{t,\ell}+p_{t,j}}\leq \frac{\alpha_{t,j}}{1-\alpha_{t,\ell}} = \frac{\alpha_{t,j}}{\sum_{\ell' \neq \ell}\alpha_{t,\ell'}}\leq 1$; and
		\item $\sum_{\ell\neq j} \frac{\alpha_{t,\ell}}{1-\alpha_{t,j}}\frac{p_{t,\ell}} {p_{t,j}+p_{t,\ell}}\leq \sum_{\ell\neq j} \frac{\alpha_{t,\ell}}{1-\alpha_{t,j}}=1$.
	\end{enumerate}
	This completes the proof.
\end{proof}

Now we are ready to complete the proof of Proposition \ref{prop:empirical overall balance}.

\begin{proof}[Proof of Proposition \ref{prop:empirical overall balance}]
	By Corollary \ref{cor:W2}, it suffices to only show
	\[
	w_{t,I^*}^2 - \sum_{j\neq I^*}w_{t,j}^2\MGFto 0.
	\]
	For any $t\in\mathbb{N}_0$, we let
	\[
	G_t \triangleq {\Psi_{t,I^*}^2} - \sum_{j\neq I^*}{\Psi_{t,j}^2},
	\]
	and we are going to show  $\frac{G_t}{t^2}\MGFto 0$.
	
	We argue that it suffices to show that there exists $T\in \MGF$ such that for any $t\geq T$,
	\begin{equation}
		\label{eq:overall_balance_Psi_sufficient_condition}
		|G_{t+1}-G_t|\leq 2 W_2t^{0.6} + 2.
	\end{equation}
	The reason is that given \eqref{eq:overall_balance_Psi_sufficient_condition}, for any $t\geq T$
	\begin{align*}
		|G_{t}| 
		= \left|G_T +\sum_{\ell=T}^{t-1} (G_{\ell+1} -G_\ell)\right| 
		&\leq  \left|G_T\right| + \sum_{\ell = T}^{t-1}|G_{\ell+1} -G_\ell|\\
		&\leq \left|G_T\right| + \sum_{\ell = T}^{t-1}\left(2W_2\ell^{0.6}+2\right)\\
		&\leq \left|G_T\right| + 2(t - T) + 2W_2\int_{T+1}^{t+1}x^{0.6}\mathrm{d}x \\
		&= \left|G_T\right| + 2(t - T) + \frac{2W_2}{1.6}\left[(t+1)^{1.6}-(T+1)^{1.6}\right]\\
		&\leq T^2 + 2(t - T) + \frac{2W_2}{1.6}\left[(t+1)^{1.6}-(T+1)^{1.6}\right],
	\end{align*}
	where in the last inequality,  $\left|G_T\right|\leq T^2$ holds since
	$
	G_T\geq -\sum_{j\neq I^*}\Psi_{T,j}^2 \geq -\left(\sum_{j\neq I^*}\Psi_{T,j}\right)^2\geq -T^2,
	$
	and 
	$
	G_T\leq \Psi_{T,I^*}^2\leq T^2.
	$
	This gives
	$
	\frac{|G_t|}{t^2}\MGFto 0.
	$
	
	Now we are going to prove the sufficient condition \eqref{eq:overall_balance_Psi_sufficient_condition}. We can calculate
	\begin{align*}
		G_{t+1} - G_t &= \left[\left(\Psi_{t,I^*}+\psi_{t,I^*}\right)^2 - \Psi_{t,I^*}^2\right]
		- \sum_{j\neq I^*}\left[\left(\Psi_{t,j}+\psi_{t,j}\right)^2 - \Psi_{t,j}^2\right] \\
		& = 2\left(\psi_{t,I^*}\Psi_{t,I^*} - \sum_{j\neq I^*}\psi_{t,j}\Psi_{t,j}\right) + \left(\psi_{t,I^*}^2 -\sum_{j\neq I^*}\psi_{t,j}^2\right).
	\end{align*}
	Then by Lemma \ref{lem:psi_bounds}, we have
	\begin{align*}
		&G_{t+1} - G_t\\
		\geq& 2\left[ \alpha_{t,I^*}\sum_{j\neq I^*} \frac{\alpha_{t,j}}{1-\alpha_{t,I^*}}\frac{p_{t,j}\Psi_{t,I^*}} {p_{t,I^*}+p_{t,j}}  -  \alpha_{t,I^*}\sum_{j\neq I^*}\frac{\alpha_{t,j}}{1-\alpha_{t,I^*}} \frac{p_{t,I^*}\Psi_{t,j}}{p_{t,I^*}+p_{t,j}} - (1-\alpha_{t,I^*})\sum_{j\neq I^*}\Psi_{t,j} \right] -1 \\
		\geq& 2 \alpha_{t,I^*}\sum_{j\neq I^*} \frac{\alpha_{t,j}}{1-\alpha_{t,I^*}}\frac{p_{t,j}\Psi_{t,I^*} - p_{t,I^*}\Psi_{t,j}} {p_{t,I^*}+p_{t,j}} - 2(1-\alpha_{t,I^*})t - 1,
	\end{align*}
	where the last inequality uses $\sum_{j\neq I^*}\Psi_{t,j}\leq t$,
	and
	\begin{align*}
		&G_{t+1} - G_t\\
		\leq& 2\left[ \alpha_{t,I^*}\sum_{j\neq I^*} \frac{\alpha_{t,j}}{1-\alpha_{t,I^*}}\frac{p_{t,j}\Psi_{t,I^*}} {p_{t,I^*}+p_{t,j}} +  (1-\alpha_{t,I^*})\Psi_{t,I^*}  -  \alpha_{t,I^*}\sum_{j\neq I^*}\frac{\alpha_{t,j}}{1-\alpha_{t,I^*}} \frac{p_{t,I^*}\Psi_{t,j}}{p_{t,I^*}+p_{t,j}}\right] + 1 \\
		\leq& 2 \alpha_{t,I^*}\left(\sum_{j\neq I^*} \frac{\alpha_{t,j}}{1-\alpha_{t,I^*}}\frac{p_{t,j}\Psi_{t,I^*} - p_{t,I^*}\Psi_{t,j}} {p_{t,I^*}+p_{t,j}}\right)  +  2(1-\alpha_{t,I^*})t   + 1,
	\end{align*}
	where the last inequality uses $\Psi_{t,I^*}\leq t$.
	Hence,
	\begin{align*}
		|G_{t+1} - G_t| &\leq 2 \alpha_{t,I^*}\sum_{j\neq I^*} \frac{\alpha_{t,j}}{1-\alpha_{t,I^*}}\left|\frac{p_{t,j}\Psi_{t,I^*} - p_{t,I^*}\Psi_{t,j}} {p_{t,I^*}+p_{t,j}}\right|  +  2(1-\alpha_{t,I^*})t + 1.
	\end{align*}
	By Lemma \ref{lem:posterior_exponential_convergence}, there exists $T_1\in\MGF$ such that for any $t\geq T_1$, $(1-\alpha_{t,I^*})t\leq 1/2$, and thus
	\begin{align*}
		|G_{t+1} - G_t|  &\leq  2 \alpha_{t,I^*}\sum_{j\neq I^*} \frac{\alpha_{t,j}}{1-\alpha_{t,I^*}}\max_{j\neq I^*} \left|\frac{p_{t,j}\Psi_{t,I^*} - p_{t,I^*}\Psi_{t,j}} {p_{t,I^*}+p_{t,j}}\right|  + 2\\
		&\leq 2\max_{j\neq I^*} \left|\frac{p_{t,j}\Psi_{t,I^*} - p_{t,I^*}\Psi_{t,j}} {p_{t,I^*}+p_{t,j}}\right|  +  2 \\
		&=  2\max_{j\neq I^*} \left|\frac{N_{t,j}\Psi_{t,I^*} - N_{t,I^*}\Psi_{t,j}} {N_{t,I^*}+N_{t,j}}\right|  +  2,
	\end{align*}
	where the last inequality applies $\alpha_{t,I^*}\leq 1$ and $\sum_{j\neq I^*}\frac{\alpha_{t,j}}{1-\alpha_{t,I^*}}=1$.
	By Corollary \ref{cor:W2}, there exists $T_{2}\in\MGF$ such that for any $t\geq T_{2}$ and $i\in[K]$, $|N_{t,i}-\Psi_{t,i}|\leq W_2 t^{0.6}$, and thus for $j\neq I^*$
	\begin{align*}
		\left|\frac{N_{t,j}\Psi_{t,I^*} - N_{t,I^*}\Psi_{t,j}} {N_{t,I^*}+N_{t,j}}\right|
		\leq \left|\frac{N_{t,j}\left(N_{t,I^*}+W_2t^{0.6}\right) - N_{t,I^*}\left(N_{t,j}-W_2t^{0.6}\right)}{N_{t,I^*}+N_{t,j}}\right|
		\leq W_2t^{0.6}.
	\end{align*}
	Hence, for $t\geq T\triangleq \max\{T_1,T_2\}$,
	$|G_{t+1} - G_t| \leq 2W_2t^{0.6} + 2$. Note that $T\in\MGF$ by Lemma \ref{lem: closedness of MGF}. This completes the proof of the sufficient condition \eqref{eq:overall_balance_Psi_sufficient_condition}.
\end{proof}

\subsubsection{Implication of empirical overall balance}
Here we present some results that are implied by empirical overall balance (Proposition \ref{prop:empirical overall balance}) and  needed for proving the sufficient condition for optimality in fixed-confidence setting (Proposition \ref{prop:sufficient condition}).
The first one shows that when $t$ is large enough, two measures of average effort are bounded away from 0 and 1.
\begin{lemma}
	\label{lem:strict boundedness}
	Let $b_1 \triangleq \frac{1}{\sqrt{32(K-1)}}$ and $b_2 \triangleq \frac{3}{4}$.
 Under TTTS with IDS, there exists a random time $T\in\MGF$ such that for any $t\geq T$,
	\[
	b_1 \leq w_{t,I^*} \leq b_2
	\quad\text{and}\quad
	b_1 \leq p_{t,I^*} \leq b_2.
	\]
\end{lemma}
Since $w_{t,I^*}$ and $p_{t,I^*}$ are bounded away from 0 when $t$ is large enough,  
we obtain the following corollary by dividing the equations in Proposition \ref{prop:empirical overall balance} by $w_{t,I^*}$ and $p_{t,I^*}$, respectively,
\begin{corollary}
	\label{cor:overall_balance_psi}
	Under TTTS with IDS,
	\[
	\sum_{j\neq I^*}\frac{w_{t,j}^2}{w_{t,I^*}^2}\MGFto 1
	\quad\text{and}\quad
	\sum_{j\neq I^*}\frac{p_{t,j}^2}{p_{t,I^*}^2}\MGFto 1.
	\]
\end{corollary}

\begin{proof}[Proof of Lemma \ref{lem:strict boundedness}]
	We only show the upper and lower bounds for $w_{t,I^*}$ since proving those for $p_{t,I^*}$ is the same. 
	
	By Proposition \ref{prop:overall_balance_Psi}, there exists $T_1\in\MGF$ such that for any $t\geq T_1$,
	\[
	w_{t,I^*}^2 - \sum_{j\neq I^*} w_{t,j}^2\leq \frac{1}{2}.
	\]
	This implies
	\[
	\frac{1}{2} \geq w_{t,I^*}^2 -\left(\sum_{j\neq I^*}w_{t,j}\right)^2 = w_{t,I^*}^2 - \left(1-w_{t,I^*}\right)^2 = 2w_{t,I^*}-1,
	\]
	which gives $w_{t,I^*}\leq b_2\triangleq \frac{3}{4}$.
	
	Using Proposition \ref{prop:overall_balance_Psi} again, there exists $T_2\in\MGF$ such that for any $t\geq T_2$,
	\[
	w_{t,I^*}^2 - \sum_{j\neq I^*} w_{t,j}^2\geq -\frac{1}{32(K-1)}.
	\]
	By Cauchy-Schwarz inequality,
	\begin{align*}
		w_{t,I^*}^2  & \geq -\frac{1}{32(K-1)} + \frac{\left(\sum_{j\neq I^*}w_{t,j}\right)^2}{K-1} \\
		& = -\frac{1}{32(K-1)} + \frac{\left(1-w_{t,I^*}\right)^2}{K-1} \\
		& \geq -\frac{1}{32(K-1)} + \frac{1}{16(K-1)} = \frac{1}{32(K-1)},
	\end{align*}
	where the last inequality uses $w_{t,I^*}\leq b_2 = \frac{3}{4}$. This gives $w_{t,I^*}\geq b_1 \triangleq \frac{1}{\sqrt{32(K-1)}}$. 
	Taking $T\triangleq \max\{T_1,T_2\}$ completes the proof since $T\in\MGF$ by Lemma \ref{lem: closedness of MGF}.
\end{proof}

\subsection{Sufficient exploration under TTTS with IDS}
In this subsection, we show that applied with information-directed selection, TTTS sufficiently explores all arms:
\begin{proposition}\label{prop: sufficient exploration}
	Under TTTS with IDS, there exists $T\in\MGF$ such that for any $t\geq T$,
	\[
	\min_{i\in[K]}N_{t,i}\geq \sqrt{t/K}.
	\]
\end{proposition}

The previous analyses in \citet{Qin2017,Shang2019, jourdan2022top} only work for any sequence of tuning parameters $\{h_t\}_{t\in\mathbb{N}_0}$ that is \emph{uniformly strictly bounded}, i.e., there exists $h_{\min}>0$ such that with probability one,
\[
\inf_{t\in\mathbb{N}_0}\min\{h_t,1-h_t\} \geq h_{\min}.
\]
Our IDS is adaptive to the algorithmic randomness in TTTS in the sense that it depends on the algorithmic randomness in picking top-two candidates,
so it does not enjoy the property above. We need some novel analysis to prove the sufficient exploration of TTTS with IDS.

Following \citet{Shang2019} of analyzing TTTS, we define the following auxiliary arms, the two ``most promising arms'':
\begin{equation}
	\label{eqn:two most promising arms chosen by TTPS}
	J_t^{(1)} \in \argmax_{i\in[K]} \alpha_{t,i} 
	\quad\text{and}\quad
	J_t^{(2)} \in \argmax_{i\neq J_t^{(1)}} \alpha_{t,i}.
\end{equation}
We further define the one that is less sampled as:
\[
J_t \triangleq \argmin_{i\in\left\{J_t^{(1)},J_t^{(2)}\right\}} N_{t,i} = \argmin_{i\in\left\{J_t^{(1)},J_t^{(2)}\right\}} p_{t,i}.
\]
Note that the identity of $J_t$ can change over time.
Next we show that IDS allocates a decent amount of effort to $J_t$.

\begin{lemma}
	\label{lem:explore under-sampled set}
	Under TTTS with IDS,
	for any $t\in \mathbb{N}_0$, 
	\begin{equation*}
		\psi_{t,J_t} \geq \frac{1}{2K(K-1)}.
	\end{equation*}
\end{lemma}
\begin{proof}
	Fix $t\in\mathbb{N}_0$. We have
	\[
	\left(I_t^{(1)}, I_t^{(2)}\right) = \left(J_t^{(1)}, J_t^{(2)}\right)\quad\text{with probability}\quad
	\alpha_{t,J_t^{(1)}} \frac{\alpha_{t,J_t^{(2)}}}{1-\alpha_{t,J_t^{(1)}}} \geq \frac{1}{K(K-1)},
	\]
	where the inequality follows from the definition of $J_t^{(1)}$ and $J_t^{(2)}$.
	Note that
	\[
	h_{J_t^{(1)},J_t^{(2)}}(\bm{p}_t;\thetabf_t) = \frac{p_{t,J_t^{(2)}}}{p_{t,J_t^{(1)}} + p_{t,J_t^{(2)}}}.
	\]
	If $J_t =\argmin_{i\in\left\{J_t^{(1)},J_t^{(2)}\right\}} p_{t,i} = J_t^{(1)}$,  
	\[
	\psi_{t,J_t} \geq \frac{1}{K(K-1)} \frac{p_{t,J_t^{(2)}}}{p_{t,J_t^{(1)}} + p_{t,J_t^{(2)}}}\geq \frac{1}{2K(K-1)}.
	\]
	Otherwise, $J_t =\argmin_{i\in\left\{J_t^{(1)},J_t^{(2)}\right\}} p_{t,i} = J_t^{(2)}$ and
	\[
	\psi_{t,J_t} \geq \frac{1}{K(K-1)} \frac{p_{t,J_t^{(1)}}}{p_{t,J_t^{(1)}} + p_{t,J_t^{(2)}}}\geq \frac{1}{2K(K-1)}.
	\]
	This completes the proof.
\end{proof}

Following \citet{Qin2017, Shang2019}, we define an insufficiently sampled set for any $t\in\mathbb{N}_0$ and $s\geq 0$,:
\[
U_t^s \triangleq \{i\in [K] \,:\, N_{t,i} < s^{1/2}\}.
\]
Proposition \ref{prop: sufficient exploration} can be rephrased as that there exists $T\in\MGF$ such that for all $t\geq T$, $U_t^{t/K}$ is empty.

Lemma 11 in \cite{Shang2019} shows a nice property of TTTS: if TTTS with a selection or tuning rule allocates a decent amount of effort to $J_\ell$ at any time $\ell$, the insufficiently sampled set $U_t^s$ is empty when $t$ is large enough.

\begin{lemma}[Lemma 11 in \cite{Shang2019}]
	\label{lem:insufficiently sampled set becomes empty}
	\footnote{While the original result focuses on $\beta$-tuning, the same proof applies to any selection rule that satisfies the condition.}Under TTTS with any selection rule (e.g., IDS) such that
	\[
	\exists \psi_{\min} > 0 \quad \forall t\in\mathbb{N}_0 \quad \psi_{t,J_t} \geq \psi_{\min},
	\]
	there exists $S\in\MGF$ such that for any $s\geq S$, $U^s_{\lfloor Ks\rfloor} = \emptyset$.
\end{lemma}

Now we can complete the proof of Proposition \ref{prop: sufficient exploration}.
\begin{proof}[Proof of Proposition \ref{prop: sufficient exploration}]
	By Lemma \ref{lem:explore under-sampled set}, IDS satisfies the condition in Lemma \ref{lem:insufficiently sampled set becomes empty}. 
	Take the corresponding $S$ in Lemma \ref{lem:insufficiently sampled set becomes empty} for IDS, and let $T \triangleq KS$. For any $t\geq T$, we let $s = t/K\geq S$, and then by Lemma \ref{lem:insufficiently sampled set becomes empty}, we have $U^s_{\lfloor Ks \rfloor} = U_{t}^{t/K}$ is empty.
\end{proof}

\subsubsection{Implication of sufficient exploration} 
Here we present some results that are implied by sufficient exploration (Proposition \ref{prop: sufficient exploration}) and needed for proving the sufficient condition for optimality in fixed-confidence setting (Proposition \ref{prop:sufficient condition}).
With sufficient exploration, the posterior means strongly converge to the unknown true means, and the probability of any sub-optimal arm being the best decays exponentially. 
\begin{lemma}[Lemmas 6 and 12 in \cite{Shang2019}]
	\label{lem:posterior_exponential_convergence}
	\footnote{While the original result focuses on $\beta$-tuning, the same proof applies to any selection rule that ensures sufficient exploration.}Under TTTS with any selection rule that ensures sufficient exploration (e.g., IDS), i.e.,
	\[
	\exists \tilde{T}\in\MGF \quad \forall t\geq \tilde{T}  \quad  \min_{i\in[K]} N_{t,i}\geq \sqrt{t/K},
	\]
	the following properties hold:
	\begin{enumerate}
		\item $\bm\theta_{t}\MGFto \bm\theta$; and
		\item there exists $T\in\MGF$ such that for any $t\geq T$ and $j\neq I^*$,
		\[
		\alpha_{t,j}\leq  \exp\left(-ct^{1/2}\right),
		\quad\text{where}\quad
		c \triangleq \frac{\Delta_{\min}^2}{16\sigma^2 K^{1/2}}.
		\]
	\end{enumerate}
\end{lemma}

\subsection{Strong convergence to optimal proportions: completing the proof of Proposition \ref{prop:sufficient condition}}
In this subsection, we are going to complete the proof of Proposition \ref{prop:sufficient condition}, the sufficient condition for optimality for fixed-confidence setting, which requires a sequence of results.

\begin{lemma}
	\label{lem:z-score-asymp}
	Under TTTS with IDS, for any arm $j\neq I^*$,
	\[ 
	\frac{C_{t,I^*, j} }{ p_{t,I^*} \cdot f_j\left(\frac{p_{t,j}}{p_{t,I^*}}\right)} \MGFto 1,
	\quad\text{where}\quad 
	f_j(x) \triangleq  \frac{\left(\theta_{I^*} - \theta_j\right)^2   }{2\sigma^2\left(1 + \frac{1}{x}\right) }
	\quad\text{and}\quad
	C_{t,I^*,j} = \frac{\left(\theta_{t,I^*} - \theta_{t,j}\right)^2   }{2\sigma^2\left(\frac{1}{p_{t,I^*}} + \frac{1}{p_{t,j}}\right) }.
	\]
\end{lemma}
\begin{proof}
	Fix $j\neq I^*$. We have
	\[
	p_{t,I^*}f_j\left(\frac{p_{t,j}}{p_{t,I^*}}\right) = p_{t,I^*}\frac{\left(\theta_{I^*} - \theta_j\right)^2   }{2\sigma^2\left(1 + \frac{p_{t,I^*}}{p_{t,j}}\right) }  = \frac{\left(\theta_{I^*} - \theta_{j}\right)^2   }{2\sigma^2\left(\frac{1}{p_{t,I^*}} + \frac{1}{p_{t,j}}\right) }.
	\]
	By Proposition \ref{prop: sufficient exploration}, TTTS with IDS ensures sufficient exploration, and thus by Lemma \ref{lem:posterior_exponential_convergence}, 
	$\bm\theta_t\MGFto\bm\theta$. Then applying the continuous mapping theorem (Lemma \ref{lem:continuous-mapping}), we have
	\[
	\frac{C_{t,I^*, j} }{ p_{t,I^*}\cdot f_j\left(\frac{p_{t,j}}{p_{t,I^*}}\right)} \MGFto 1.
	\]
	This completes the proof.
\end{proof}

\begin{lemma}
	\label{lem:over sampled implies exponentially small}
	Under TTTS with IDS, for any $\epsilon > 0$, there exists a deterministic constant $c_\epsilon>0$ and a random time $T_\epsilon\in\MGF$ such that for any $t\geq T_\epsilon$ and $j\neq I^*$,
	\[
	\frac{w_{t,j}}{w_{t,I^*}} > \frac{p_j^*+\epsilon}{p_{I^*}^*} \quad\implies\quad  \psi_{t,j}\leq \exp\left(-c_\epsilon t\right) + (K-1)\exp\left(-ct^{1/2}\right),
	\]
	where $c$ is defined in Lemma \ref{lem:posterior_exponential_convergence}.
\end{lemma}
\begin{proof}
	Fix $\epsilon > 0$. It suffices to show that for any $j\neq I^*$, there exist a deterministic constant $c^{(j)}_\epsilon>0$
	and a random time $T^{(j)}_\epsilon\in\MGF$ such that for any $t\geq T^{(j)}_\epsilon$,
	\[
	\frac{w_{t,j}}{w_{t,I^*}} > \frac{p_j^*+\epsilon}{p_{I^*}^*} \quad\implies\quad  \psi_{t,j}\leq \exp\left(-c_\epsilon^{(j)} t\right) + (K-1)\exp\left(-ct^{1/2}\right),
	\]
	since this completes the proof by taking $T_\epsilon \triangleq \max_{j\neq I^*} T^{(j)}_\epsilon$ and $c_\epsilon \triangleq \min_{j \neq I^*} c^{(j)}_\epsilon$.
	
	From now on, we fix $j\neq I^*$.  By Corollary \ref{cor:W2}, there exists $T_{\epsilon,1}^{(j)}\in\MGF$ such that for any $t\geq T_{\epsilon,1}^{(j)}$,
	\[
	\frac{w_{t,j}}{w_{t,I^*}} = \frac{\Psi_{t,j}}{\Psi_{t,I^*}} > \frac{p_j^*+\epsilon}{p_{I^*}^*}
	\quad\implies\quad \frac{p_{t,j}}{p_{t,I^*}} = \frac{N_{t,j}}{N_{t,I^*}} > \frac{p_j^*+\epsilon/2}{p_{I^*}^*}.
	\]
	Then by Corollary \ref{cor:overall_balance_psi}, there exists $T_{\epsilon,2}^{(j)}\in\MGF$ such that for any $t\geq T_{\epsilon,2}^{(j)}$,
	\[
	\frac{p_{t,j}}{p_{t,I^*}} > \frac{p_j^*+\epsilon/2}{p_{I^*}^*} 
	\quad\implies\quad
	\exists A_t\neq I^* \,\, : \,\, \frac{p_{t,A_t}}{p_{t,I^*}}   \leq \frac{p_{A_t}^*}{p_{I^*}^*}.
	\]
	
	From now on, we consider $t\geq \max\left\{T_{\epsilon,1}^{(j)},T_{\epsilon,2}^{(j)}\right\}$ and we have
	\begin{equation}
		\label{eq:over_allocated_and_under_allocated}
		\frac{w_{t,j}}{w_{t,I^*}} = \frac{\Psi_{t,j}}{\Psi_{t,I^*}} > \frac{p_j^*+\epsilon}{p_{I^*}^*}
		\quad\implies\quad \frac{p_{t,j}}{p_{t,I^*}} = \frac{N_{t,j}}{N_{t,I^*}} > \frac{p_j^*+\epsilon/2}{p_{I^*}^*}
		\quad\text{and}\quad
		\exists A_t\neq I^* \,\, : \,\, \frac{p_{t,A_t}}{p_{t,I^*}}   \leq \frac{p_{A_t}^*}{p_{I^*}^*}.
	\end{equation}
	By Lemmas \ref{lem:psi_bounds} and \ref{lem:posterior_exponential_convergence},
	\begin{align*}
		\psi_{t,j} 
		\leq
		\alpha_{t,I^*}\frac{\alpha_{t,j}}{1-\alpha_{t,I^*}} \frac{p_{t,I^*}}{p_{t,I^*}+p_{t,j}} + (1-\alpha_{t,I^*})
		\leq  \frac{\alpha_{t,j}}{1-\alpha_{t,I^*}} +  (K-1)\exp\left(-ct^{1/2}\right).
	\end{align*}
	We can upper bound the numerator $\alpha_{t,j}$ as follows,
	\begin{align*}
		\alpha_{t,j}  \leq \Prob_t\left(\tilde{\theta}_{j}\geq \tilde{\theta}_{I^*}\right) & = \Prob\left(\frac{\tilde{\theta}_{I^*}-\tilde{\theta}_{j}-(\theta_{t,I^*}-\theta_{t,j})}{\sigma\sqrt{1/N_{t,I^*}+1/N_{t,j}}} \leq \frac{-(\theta_{t,I^*}-\theta_{t,j})}{\sigma\sqrt{1/N_{t,I^*}+1/N_{t,j}}}\right) \\
		& =\Phi\left(-\sqrt{2tC_{t,I^*,j}}\right),   
	\end{align*}
	where $\tilde{\bm\theta} = \left(\tilde{\theta}_{1},\ldots, \tilde{\theta}_{K}\right)$ is a sample drawn from the posterior distribution $\Pi_t$ and $\Phi(\cdot)$ is Gaussian cumulative distribution function. Similarly, we can lower bound the denominator $1 - \alpha_{t,I^*}$ as follows,
	\[
	1 - \alpha_{t,I^*} = \Prob_t\left(\exists a\neq I^*\,:\, \tilde{\theta}_{a} \geq \tilde{\theta}_{I^*}\right) \geq \Prob_t\left(\tilde{\theta}_{A_t}\geq \tilde{\theta}_{I^*}\right)= \Phi\left(-\sqrt{2tC_{t,I^*,A_t}}\right).
	\]
	Hence,
	\[
	\psi_{t,j}  \leq  \frac{\Phi\left(-\sqrt{2tC_{t,I^*,j}}\right)}{\Phi\left(-\sqrt{2tC_{t,I^*,A_t}}\right)} + (K-1)\exp\left(-ct^{1/2}\right).
	\]
	The remaining task is to show that there exist a deterministic constant $c^{(j)}_\epsilon > 0$ and a random time $T^{(j)}_{\epsilon,3}\in\MGF$ such that for any $t\geq T^{(j)}_{\epsilon,3}$,
	\begin{equation}
		\label{eq:tiny probablity}
		\frac{\Phi\left(-\sqrt{2tC_{t,I^*,j}}\right)}{\Phi\left(-\sqrt{2tC_{t,I^*,A_t}}\right)} \leq \exp\left(-c_\epsilon^{(j)}t\right).
	\end{equation}
	By Lemma \ref{lem:z-score-asymp}, for any $\delta > 0$,
	\begin{align*}
		\frac{\Phi\left(-\sqrt{2tC_{t,I^*,j}}\right)}{\Phi\left(-\sqrt{2tC_{t,I^*,A_t}}\right)} 
		&\leq \frac{\Phi\left(-\sqrt{2tp_{t,I^*}f_j\left(\frac{p_{t,j}}{p_{t,I^*}}\right)(1-\delta)}\right)}{\Phi\left(-\sqrt{2tp_{t,I^*}f_{A_t}\left(\frac{p_{t,A_t}}{p_{t,I^*}}\right)(1+\delta)}\right)} \\
		&\leq \frac{\Phi\left(-\sqrt{2tp_{t,I^*}f_j\left(\frac{p^*_{j}+\epsilon/2}{p^*_{I^*}}\right)(1-\delta)}\right)}{\Phi\left(-\sqrt{2tp_{t,I^*}f_{A_t}\left(\frac{p^*_{A_t}}{p^*_{I^*}}\right)(1+\delta)}\right)} \\
		&= \frac{\Phi\left(-\sqrt{2tp_{t,I^*}f_j\left(\frac{p^*_{j}+\epsilon/2}{p^*_{I^*}}\right)(1-\delta)}\right)}{\Phi\left(-\sqrt{2tp_{t,I^*}f_{j}\left(\frac{p^*_{j}}{p^*_{I^*}}\right)(1+\delta)}\right)},
	\end{align*}
	where the second inequality follows from \eqref{eq:over_allocated_and_under_allocated} the monotonicity of $f_j$ and $f_{A_t}$ and the final equality uses the information balance \eqref{eq:def_C_ij_Gaussian}.
	
	We can pick a sufficiently small $\delta$ as a function of $\epsilon$ such that
	\[
	c_1 \triangleq f_j\left(\frac{p^*_{j}+\epsilon/2}{p^*_{I^*}}\right)(1-\delta) > f_{j}\left(\frac{p^*_{j}}{p^*_{I^*}}\right)(1+\delta) \triangleq c_2.
	\]
	Then \eqref{eq:tiny probablity} follows from Gaussian tail upper and lower bounds (or $\frac{1}{t} \log \Phi\left(-\sqrt{t} x\right) \to - x^2/2$ as $t\to \infty$) and that $p_{t,I^*}$ is bounded away from 0 and 1 when $t$ is large (Lemma \ref{lem:strict boundedness}). 
	This completes the proof.
\end{proof}

\begin{lemma}
	\label{lem:all arms not over sampled}
	Under TTTS with IDS, for any $\epsilon > 0$, there exists $T_\epsilon\in\MGF$ such that for any $t\geq T_\epsilon$,
	\[
	\frac{w_{t,j}}{w_{t,I^*}} \leq  \frac{p_j^*+\epsilon}{p_{I^*}^*},\quad \forall j\neq I^*.
	\]
\end{lemma}
\begin{proof}
	Fix $\epsilon > 0$ and $j\neq I^*$. By Lemma \ref{lem:over sampled implies exponentially small}, there exists $\tilde{T}_\epsilon\in\MGF$ such that for any $t\geq \tilde{T}_\epsilon$ and $j\neq I^*$,
	\[
	\frac{w_{t,j}}{w_{t,I^*}} > \frac{p_j^*+\epsilon/2}{p_{I^*}^*} \quad\implies\quad  \psi_{t,j}\leq \exp\left(-c_\epsilon t\right) + (K-1)\exp\left(-ct^{1/2}\right).
	\]
	Define 
	\[
	\kappa_{\epsilon} \triangleq \sum_{\ell=0}^{\infty} \left[\exp\left(-c_{\epsilon}\ell\right) + (K-1)\exp\left(-c\ell^{1/2}\right)\right] < \infty.
	\]
	
	From now on, we consider $t\geq \tilde{T}_\epsilon$. We need to consider two cases, and we provide an upper bound on $\frac{w_{t,j}}{w_{t,I^*}}$ for each case.
	
	\begin{enumerate}
		\item The first case supposes that 
		\[
		\forall \ell\in \left\{\tilde{T}_{\epsilon},\tilde{T}_{\epsilon}+1,\ldots, t-1\right\}:
		\quad
		\frac{w_{\ell,j}}{w_{\ell,I^*}} \geq \frac{p_j^* + \epsilon/2}{p^*_{I^*}}.
		\] 
		We have
		\begin{align*}
			\Psi_{t,j}
			&= \Psi_{\tilde{T}_{\epsilon},j}  + \sum_{\ell = \tilde{T}_{\epsilon}}^{t-1} \psi_{\ell, j}\\
			&= \Psi_{\tilde{T}_{\epsilon},j}
			+ \sum_{\ell = \tilde{T}_{\epsilon}}^{t-1} \psi_{\ell, j}\mathbf{1}\left(\frac{w_{\ell,j}}{w_{\ell,I^*}} \geq \frac{p_j^* + \epsilon/2}{p^*_{I^*}} \right) \\
			&\leq \Psi_{\tilde{T}_{\epsilon},j} + \sum_{\ell = \tilde{T}_{\epsilon}}^{t-1} \left[\exp\left(-c_{\epsilon}\ell\right)+ (K-1)\exp\left(-c\ell^{1/2}\right) \right]  \\
			&\leq \Psi_{\tilde{T}_{\epsilon},j} + \kappa_{\epsilon}.
		\end{align*}
		Take $T\in \MGF$ in Lemma \ref{lem:strict boundedness}, so for any $t\geq T$, $\Psi_{t,I^*}\geq b_1 t$, and thus
		\begin{equation}
			\label{eq:case 1}
			\frac{w_{t,j}}{w_{t,I^*}} = \frac{\Psi_{t,j}}{\Psi_{t,I^*}}
			\leq \frac{\Psi_{\tilde{T}_{\epsilon},j} + \kappa_{\epsilon}}{\Psi_{t,I^*}} \leq \frac{\tilde{T}_{\epsilon} + \kappa_{\epsilon}}{b_1t}.
		\end{equation}
		\item The alternative case supposes that
		\[
		\exists\ell\in \left\{\tilde{T}_{\epsilon},\tilde{T}_{\epsilon}+1,\ldots, t-1\right\}:
		\quad 
		\frac{w_{\ell,j}}{w_{\ell,I^*}} < \frac{p_j^* + \epsilon/2}{p^*_{I^*}}.
		\]
		We define 
		\[
		L_t \triangleq \max \left\{\ell\in \left\{\tilde{T}_{\epsilon},\tilde{T}_{\epsilon}+1,\ldots, t-1\right\} \,:\, \frac{w_{\ell,j}}{w_{\ell,I^*}} < \frac{p_j^* + \epsilon/2}{p^*_{I^*}}\right\}.
		\]
		We have
		\begin{align*}
			\Psi_{t,j} 
			&= \Psi_{L_t, j} + \psi_{L_t,j} + \sum_{\ell = L_t+1}^{t-1} \psi_{\ell, j} \\
			&= \Psi_{L_t, j} 
			+ \psi_{L_t,j}
			+ \sum_{\ell = L_t + 1}^{t-1} \psi_{\ell, j}\mathbf{1}\left(\frac{w_{\ell,j}}{w_{\ell,I^*}} \geq \frac{p_j^* + \epsilon/2}{p^*_{I^*}}\right) 
			\\
			&\leq \frac{p_j^* + \epsilon/2}{p^*_{I^*}}\Psi_{L_t,I^*}  + 1 +  \sum_{\ell = L_t + 1}^{t-1} \left[\exp\left(-c_{\epsilon}\ell\right)+ (K-1)\exp\left(-c\ell^{1/2}\right) \right]\\
			&\leq \frac{p_j^* + \epsilon/2}{p^*_{I^*}}\Psi_{L_t,I^*} + \left(\kappa_{\epsilon} + 1\right),
		\end{align*}
		where the first inequality uses $\frac{\Psi_{L_t,j}}{\Psi_{L_t,I^*}} = \frac{w_{L_t,j}}{w_{L_t,I^*}} < \frac{p_j^* + \epsilon/2}{p^*_{I^*}}$. 
		Then by Lemma \ref{lem:strict boundedness}, for any $t\geq T$,
		\begin{equation}
			\label{eq:case 2}
			\frac{w_{t,j}}{w_{t,I^*}}  \leq \frac{p_j^* + \epsilon/2}{p^*_{I^*}}\frac{\Psi_{L_t,I^*}}{\Psi_{t,I^*}} + \frac{1 + \kappa_{\epsilon} }{\Psi_{t,I^*}} \leq  \frac{p_j^* + \epsilon/2}{p^*_{I^*}} +  \frac{1 + \kappa_{\epsilon} }{b_1t}.
		\end{equation}
	\end{enumerate}
	
	Putting \eqref{eq:case 1} and \eqref{eq:case 2} together, we find that for any $t\geq T_1$, 
	\[ 
	\frac{w_{t,j}}{w_{t,I^*}} = \frac{\Psi_{t,j}}{\Psi_{t,I^*}} \leq \max\left\{  \frac{\tilde{T}_{\epsilon} + \kappa_{\epsilon}}{b_1t} \, , \,\frac{p_j^* + \epsilon/2}{p^*_{I^*}} +  \frac{\kappa_{\epsilon} + 1}{b_1t} \right\} \leq \frac{p_j^* + \epsilon/2}{p^*_{I^*}} +\frac{\tilde{T}_\epsilon + 2\kappa_{\epsilon} + 1}{t}. 
	\]
	Then we have 
	\[
	t\geq \frac{2\left( \tilde{T}_\epsilon + 2\kappa_{\epsilon} + 1 \right)}{\epsilon}
	\quad\implies\quad
	\frac{w_{t,j}}{w_{t,I^*}} \leq \frac{p_j^*+ \epsilon}{p^*_{I^*}}.
	\]
	Taking
	$
	T_\epsilon \triangleq  \max\left\{\tilde{T}_\epsilon, T, \frac{2\left( \tilde{T}_\epsilon + 2\kappa_{\epsilon} + 1 \right)}{\epsilon}\right\}
	$
	completes the proof since Lemma \ref{lem: closedness of MGF} implies $T_\epsilon\in\MGF$.
\end{proof}

Now we are ready to complete the proof of Proposition \ref{prop:sufficient condition}.
\begin{proof}[Proof of Proposition \ref{prop:sufficient condition}]
	Corollary \ref{cor:overall_balance_psi} and Lemma \ref{lem:all arms not over sampled} give Proposition \ref{prop:sufficient condition}.
\end{proof}

\newpage
\section{Supporting results for Section \ref{sec:existing}: structure of the optimal solution for best-\texorpdfstring{$k$}{k}-arm identification} \label{app:optiaml_allocation_problem}

In this section, we present detailed study of the optimal allocation problem \eqref{eq:optimal_allocation}.
In particular, we characterize the optimality conditions in terms of the allocation vector $\bm p$ \textit{without the presence of the dual variable} $\bm \mu$.
This type of conditions is frequently analyzed for BAI, see \cite{GarivierK16,Chen2000OCBA,Glynn2004,chen2023balancing}.
Conditions based only on $\bm p$ are attractive because they directly yield intuitive and effective algorithms.

The goal of this section is to show that sufficient and necessary conditions based \textit{only on} $\bm p$ can present complicated structure and generalizing algorithms such as \cite{chen2023balancing} is a highly non-trivial task, if not impossible.

The key challenge in solving \eqref{eq:optimal_allocation} is that $\Gamma_{\bm\theta}(\bm p)$ is a minimum among a set of functions. 
Hence, there are multiple $(i,j)$ pairs such that $C_{i,j}(\bm p^*)$ achieves the optimal value $\Gamma^*_{\thetabf}$.
Note that by complementary slackness, the dual variable $\mu_{i,j}$ can only be positive for the pairs $(i,j)$ such that $C_{i,j}(\bm p^*)$ achieves the optimal value $\Gamma^*_{\thetabf}$.
Our KKT analysis greatly simplify the characterization of the optimal solution by directly encoding the solution structure in the complementary slackness conditions.
However, if necessary and sufficient conditions independent of the dual variable $\bm \mu$ were to be derived, the discussion of the detailed structure cannot be bypassed.

\subsection{Properties of the optimal solution}
\label{sec:properties_optimal_solution}
Before discussing the structure results, we present several properties of the optimal solution to \eqref{eq:optimal_allocation}. These properties generalize those for BAI in \cite{GarivierK16}.
\uniqueness*
\begin{lemma}[Positivity]
	\label{lmm:positive_sol}
	The optimal solution $\bm p^* > \bm{0}$. 
\end{lemma}

\begin{lemma}[Monotonicity]
	\label{prop: monotonicity}
	The optimal solution $\bm p^*$ satisfies the following properties:
	\begin{align*}
		\forall i,i'\in\I:\quad \theta_i\geq \theta_{i'} \implies p^*_{i} \le p^*_{i'}
		\quad\text{and}\quad
		\forall j,j'\in\J:\quad \theta_j\geq \theta_{j'} \implies p^*_{j} \ge p^*_{j'}.
	\end{align*}
\end{lemma}

\subsection{Proof of Proposition \ref{prop:information_balance} (A necessary condition -- information balance)}
Note that for any $(i,j)\in\I\times\I^c$, $C_{i,j}(\cdot)$ depends on $\bm p$ only through $p_i,p_j$.
Throughout this section, we write $C_{i,j}(p_i, p_j) = C_{i,j}(\bm p)$ with a slight abuse of notation.
\begin{proof}
	We first prove the second statement of \eqref{eq:information_balance}, i.e.,
	\begin{equation}\label{eq:j_IB}
		\Gamma_{\bm\theta}^* = \min_{i' \in \I} C_{i',j}(\bm p^*), \quad \forall j\in\J.
	\end{equation}
	
	We prove by contradiction.  Suppose that \eqref{eq:j_IB} does not hold. Let $\J_{\text{min}}\subset\J$ be the set of arms $j\in\J$ such that $\min_{i' \in \I} C_{i',j}(\bm p^*) = \Gamma^*_{\bm\theta}$.
	Then $\J_{\text{min}}\neq\J$. By the definition of $\Gamma^*_{\thetabf}$, $\J_{\text{min}}$ is non-empty.
	
	Since $C_{i,j}$'s are continuous functions strictly increasing in $p_{j}$, there exists a sufficiently small $\epsilon > 0$ such that 
	\[\min_{i \in \I} C_{i,j}\left(p^*_i, p^*_j-\epsilon/(|\J| -|\J_{\text{min}}|)\right) > \Gamma^*_{\bm\theta}, \quad \forall j \in \J\backslash \J_{\text{min}}\]
	and 
	\[\min_{i \in \I} C_{i,j'}(p^*_i, p^*_{j'} + \epsilon/|\J_{\text{min}}|) > \min_{i \in \I} C_{i,j'}(p^*_i, p^*_{j'}) = \Gamma^*_{\bm\theta}, \quad  \forall j' \in \J_{\text{min}}.\]
	Hence, $\bm p^*$ cannot be an optimal solution, leading to a contradiction. 
	Consequently, we must have \eqref{eq:j_IB}.
	
	With similar argument, we can show that 
	\[
	\Gamma^*_{\thetabf} = \min_{j' \in \J} C_{i,j}(\bm{p}^*), \quad\forall i \in \I,
	\]
	using the fact that $C_{i,j}$'s are continuous functions strictly increasing in $p_{i}$. 
\end{proof}

\subsection{Overall balance in connected subgraphs: additional necessary conditions}
\label{app:overall_balance_details}
Next, we identify set of necessary condition for optimality based on balancing the allocation between the sets of top arms and bottom arms.
For general reward distributions, this condition appears to have an obscure expression.
However, the result is significantly simplified for Gaussian bandits as we discuss now.

Recall from Proposition \ref{prop:information_balance} that there are multiple $(i,j)$ pairs such that $C_{i,j}(\bm p^*) = \Gamma^*_{\thetabf}$. To facilitate discussion of the solution structure, we require the following definition.
\begin{definition}[Bipartite graph induced by the optimal solution]
	The \emph{bipartite graph} induced by the optimal solution $\bm p^*$ is denoted by $\mathcal G(\I,\J,\mathcal E)$ where the set of edges $\mathcal E$ contains any $(i,j) \in\I\times\J$ such that $C_{i,j}(\bm p^*) = \Gamma^*_{\thetabf}$. 
\end{definition}

\begin{remark}[Connected subgraphs]
	By Proposition \ref{prop:information_balance}, for any arm $i \in \I \cup \J$, there must exist an edge $e \in \mathcal E$ such that $i$ is incident to $e$. Hence, the bipartite graph  $\mathcal G(\I,\J,\mathcal E)$ can be uniquely decomposed into a collection of $L$ \textit{connected subgraphs} $\left\{\mathcal G_{\ell}(\I_{\ell},\J_{\ell},\mathcal E_\ell)\right\}_{\ell\in[L]}$ such that 
	\begin{enumerate}
		\item for any $\ell\in[L]$, $\I_{\ell},\J_{\ell},\mathcal E_\ell \neq \varnothing$;
		\item $\{\I_{\ell}\}_{\ell\in[L]}$ are mutually exclusive with $\bigcup_{\ell\in[L]}\I_{\ell} = \I$; 
		$\{\J_{\ell}\}_{\ell\in[L]}$ are mutually exclusive with $\bigcup_{\ell\in[L]}\J_{\ell} = \J$; 
		$\{\mathcal E_{\ell}\}_{\ell\in[L]}$ are mutually exclusive with $\bigcup_{\ell\in[L]}\mathcal{E}_{\ell} = \mathcal{E}$.
	\end{enumerate}
\end{remark}

We give the following example \ref{ex:disconnected_graph} to illustrate the concept of the bipartite graph and connected subgraphs.
\begin{example}\label{ex:disconnected_graph}
	Consider Gaussian bandits with mean rewards $\bm\theta = \{ 0.51, 0.5, 0, -0.01, -0.092\}$, variance $\sigma^2 = 1/4$  and $k=2$. 
	The optimal allocation\footnote{The optimal allocation can be obtained by numerically solving the three scenarios listed in Remark \ref{rmk:full_characterization_2_5} and selected the best solution among them.} is 
	\[\bm p^* \approx (0.21851674, 0.23713642, 0.21851674, 0.20260818, 0.12322192).\] 
	We have $\Gamma^*_{\thetabf} = C_{1,3}(\bm p^*) = C_{2,4}(\bm p^*) = C_{2,5}(\bm p^*) = 0.0568362041$. Hence, there are two connected subgraphs.  Note that in each subgraph, the sum-of-squared allocations are balanced, i.e., 
	\begin{equation*}
		(p^*_1)^2 = (p^*_3)^2 \quad\text{and}\quad (p^*_2)^2 = (p^*_4)^2 + (p^*_5)^2.
	\end{equation*}
\end{example}

Now, we are ready to present the necessary condition for Gaussian bandits.
\begin{proposition}[Overall balance in connected subgraphs -- Gaussian bandits]
	\label{prop:SS_balance}
	For Gaussian bandits, each connected subgraph of the bipartite graph induced by the optimal solution balances the sum-of-squared allocations, i.e.,
	\begin{equation}
		\label{eq:SS_balance}
		\sum_{i\in\mathcal{I}_\ell}(p^{*}_i)^2 = \sum_{j\in\I^c_\ell}(p^{*}_j)^2, \quad\forall \ell\in[L].
	\end{equation}
\end{proposition}

\begin{remark}[Proposition \ref{prop:SS_balance} implies overall balance in Proposition \ref{prop:overall_balance}] 
	\label{rmk:SS_overall}
	Equation \eqref{eq:SS_balance} is strictly stronger than Equation \eqref{eq:overall_balance}.
	Example \ref{ex:disconnected_graph} provides such an illustration.
\end{remark}

\begin{example}[Failure of $\beta$-tuning]\label{ex:failure_of_fixed_tuning} 
	Consider the Gaussian bandits in Example \ref{ex:disconnected_graph}.
	We now show that top-two algorithms with $\beta^*$-tuning can fail, where $\beta^* = p_1^* + p_2^*$.
	As a counterexample, 
	\[\bm p' \approx (0.21843623, 0.23721697, 0.21859726, 0.20254937, 0.12320016)\] 
	is a feasible solution such that $p'_1 + p'_2 = p_1^* + p_2^* \triangleq \beta^*$ but 
	\[
	\Gamma_{\bm \theta}(\bm p') = C_{1,3}(\bm p') = C_{1,4}(\bm p') = C_{2,4}(\bm p') = C_{2,5}(\bm p') = 0.056836198 < 0.0568362041 = \Gamma^*_{\thetabf}.
	\] 
	This example show that information balance \eqref{eq:information_balance}, together with $p_1+p_2 = \beta^*$ does not guarantee the uniqueness of the solution. 
	More importantly, we show with an numerical example in Figure \ref{fig:fail} that two-two algorithms with $\beta$-tuning is not guaranteed to be optimal, even if supplied with the optimal $\beta^*$.
	In particular, we consider the Gaussian bandits in Example \ref{ex:disconnected_graph}, assuming that the true mean vector is known.  
	Figure \ref{fig:fail} reports the sample path averaged across $100$ replications of $\Gamma_{\bm\theta}(\bm p_t)$ where $\bm p_t$ is the sample allocation under either \emph{TS-KKT+($0$)}\footnote{Since $\bm\theta$ is assumed to be known, the empirical best-$k$ is just the true best-$k$.  We focus on achieving the optimal allocation for this demonstration.} or an $\beta^*$-tuning version of \emph{TS-KKT+($0$)}, i.e., let $\beta_t = \beta^*$ with $\beta^* = p_1^* + p_2^*$. It shows that the \emph{TS-KKT+($0$)} algorithm with $\beta^*$-tuning converges to the solution $\bm p'$, which is sub-optimal.
	We further remark that $O(1/t)$ is the best achievable convergence rate, because we are only allowed to draw one arm in each step, effectively fixing the step size to $1/t$.
	\begin{figure}[htbp]
		\centering
		\includegraphics[width=\textwidth]{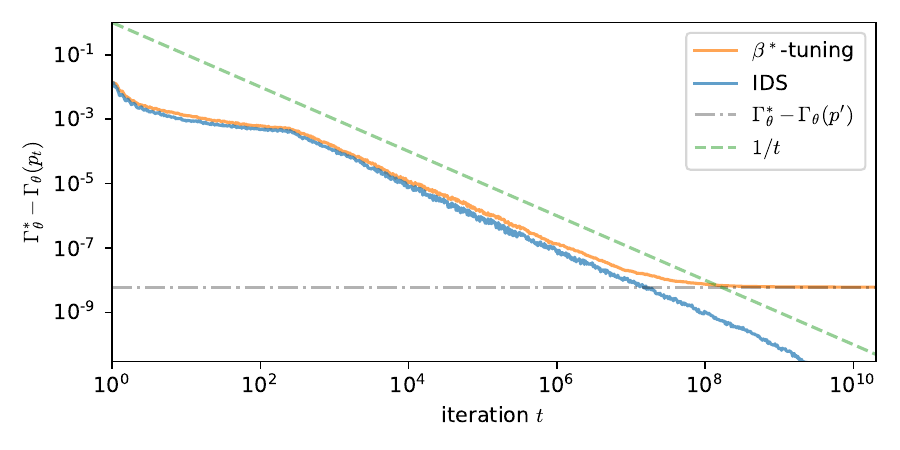}
		\caption{The sample path of $\Gamma_{\bm\theta}(\bm p_t)$ for $\bm p_t$ being the allocation under either TS-KKT+($0$) or an $\beta^*$-tuning version of TS-KKT+($0$), i.e., let $\beta_t \equiv \beta^* = p_1^* + p_2^*$.  Averaged across 100 replications.  For demonstration, we assume that $\bm\theta$ is known so that both algorithms are effectively deterministic. In this setting without learning, all empirical and sampled quantities are replaced by the true mean $\bm\theta$ and the TS-KKT+($0$) algorithm is equivalent to the (best-$k$-arm version of) EB-TC algorithm in \cite{jourdan2022top}.}
		\label{fig:fail}
	\end{figure}
	The rational behind this counterexample is that the optimal solution induces two connected subgraphs, and each subgraph need its dedicated balancing condition. 
	Hence, an overall balance as such induced by $\beta$-tuning (even if set to $\beta^*$) is not sufficient to discern the optimal allocation.
\end{example}

\begin{example}\label{ex:failure_of_sufficiency}
	We point out that \eqref{eq:information_balance} and \eqref{eq:overall_balance} together do not guarantee the uniqueness of the solution nor the optimality.
	Even for choosing the best-$2$ among $4$ arms in a Gaussian bandit, counterexamples are ubiquitous. 
	Consider the Gaussian bandits with mean rewards $\bm\theta = (1,0.7,0,-0.5)$, variance $\sigma^2 = 1/4$ and $k = 2$. The solution $\bm p_1 \approx (0.0482, 0.459, 0.4603, 0.0325)$
	satisfies \eqref{eq:information_balance} and \eqref{eq:overall_balance} with an objective value of $\Gamma_{\thetabf}(\bm p_1) \approx 0.0873$. For $\bm p_1$, we have $C_{1,3}(\bm p_1) = C_{1,4}(\bm p_1) = C_{2,4}(\bm p_1) < C_{2,3}(\bm p_1)$.
	However, the optimal solution $\bm p^* \approx (0.1277, 0.3894, 0.4016, 0.0813)$, with an optimal objective value of $\Gamma^*_{\thetabf} \approx 0.1938 > \Gamma_{\thetabf}(\bm p_1)$.    For the optimal solution, we have $C_{1,3}(\bm p^*) = C_{2,4}(\bm p^*) = C_{2,3}(\bm p^*) < C_{1,4}(\bm p^*)$.
\end{example}

Now we formally prove Proposition \ref{prop:SS_balance}.

\begin{proof}[Proof of Proposition \ref{prop:SS_balance} (Overall balance in connected subgraphs -- Gaussian)]
	Consider any connected subgraph $\mathcal G_\ell = \mathcal G(\I_\ell,\J_\ell,\mathcal E_\ell)$ of the bipartite graph induced by an optimal solution $\bm p$.
	We show that if 
	\begin{equation*}
		\sum_{i\in\mathcal{I}_\ell}p_i^2 \neq \sum_{j\in\I^c_\ell}p_j^2.
	\end{equation*}
	then we can find another $\bm p' \neq \bm p$ with higher objective value, and hence $\bm p$ cannot be an optimal solution.
	
	By Proposition \ref{prop:information_balance}, for any arm $i \in [K]$, there must exist an edge $e \in \mathcal E$ such that $i$ is incident to $e$.
	Hence, we have $\I_\ell\neq \varnothing$ and $\J_\ell \neq \varnothing$.
	Take any $i_1 \in \I_\ell$ as an anchor point.  By the connectedness of $\mathcal G_\ell$, for any $j \in \J_\ell$, there exist a path $p_{i_1,j} = (i_1, j_1, i_2, j_2, \dots, j_r = j)$ connecting $i_1$ and $j$.
	Along the path, we perform Taylor's expansion
	\begin{equation*}
		C_{i,j}(\bm p + \bm\epsilon) = C_{i,j}(\bm p) + \frac{\partial C_{i,j}(\bm p)}{\partial p_i}\epsilon_i + \frac{\partial C_{i,j}(\bm p)}{\partial p_j}\epsilon_j + o(\|\bm\epsilon\|).
	\end{equation*}
	Starting from the first edge $(i_1,j_1)$, we define $\tilde\epsilon_{i_1}$ and $\tilde\epsilon_{j_1}$ such that the first order terms in the Taylor's series vanishes
	\begin{equation}\label{eq:def_tilde_epsilon}
		\frac{\partial C_{i_1,j_1}(\bm p)}{\partial p_{i_1}}\tilde\epsilon_{i_1} + \frac{\partial C_{i_1,j_1}(\bm p)}{\partial p_{j_1}}\tilde\epsilon_{j_1} = 0.
	\end{equation}
	Recall that, for Gaussian bandits with reward distribution $N(\theta_i,\sigma^2)$, we have 
	\[C_{i,j}(\bm p) = \frac{(\theta_i - \theta_j)^2}{2\sigma^2}\frac{p_i p_j}{p_i + p_j}\]
	and
	\begin{align*}
		\frac{\partial C_{i,j}(\bm p)}{\partial p_i} 
		= \frac{(\theta_i - \theta_j)^2}{2\sigma^2}\frac{p_j^2}{(p_i + p_j)^2}, \quad \frac{\partial C_{i,j}(\bm p)}{\partial p_j} 
		= \frac{(\theta_i - \theta_j)^2}{2\sigma^2}\frac{p_i^2}{(p_i + p_j)^2}.   
	\end{align*}
	Hence, we can rewrite \eqref{eq:def_tilde_epsilon} as 
	\begin{equation*}
		\tilde\epsilon_{j_1} = -\tilde\epsilon_{i_1} \frac{p^2_{j_1}}{p^2_{i_1}}.
	\end{equation*}
	Repeat this procedure for $(j_1, i_2)$ on the path, we have
	\begin{equation*}
		\tilde\epsilon_{i_2} = -\tilde\epsilon_{j_1} \frac{p^2_{i_2}}{p^2_{j_1}} = \tilde\epsilon_{i_1} \frac{p^2_{j_1}}{p^2_{i_1}}\frac{p^2_{i_2}}{p^2_{j_1}} = \tilde\epsilon_{i_1} \frac{p^2_{i_2}}{p^2_{i_1}}.
	\end{equation*}
	Repeat for all edges on the path, we have
	\begin{equation*}
		\tilde\epsilon_{j} = -\tilde\epsilon_{i_1} \frac{p^2_{j}}{p^2_{i_1}}, \quad j \in \J_\ell.
	\end{equation*}
	
	Similarly, for all $i \in \I_\ell$, we have
	\begin{equation*}
		\tilde\epsilon_{i} = \tilde\epsilon_{i_1} \frac{p^2_{i}}{p^2_{i_1}}, \quad i \in \I_\ell.
	\end{equation*}
	
	Now, we claim that if $\sum_{i\in\mathcal{I}_\ell}p_i^2 \neq \sum_{j\in\I^c_\ell}p_j^2$, then we can find a $\bm\epsilon$ such that $\Gamma_{\thetabf}(\bm p + \bm\epsilon) > \Gamma_{\thetabf}(\bm p)$.  Consequently, $\bm p$ cannot be an optimal solution.
	
	To this end, let
	\[\delta \equiv \sum_{i \in \I_\ell} \tilde\epsilon_i + \sum_{j \in \J_\ell} \tilde\epsilon_j = \tilde\epsilon_{i_1} \left(\sum_{i \in \I_\ell} \frac{p^2_i}{p^2_{i_1}} - \sum_{j \in \J_\ell} \frac{p^2_j}{p^2_{i_1}}\right).\]
	We consider two cases.
	\begin{enumerate}
		\item If $\sum_{i\in\mathcal{I}_\ell}p_i^2 > \sum_{j\in\I^c_\ell}p_j^2$, then let $\epsilon_{i_1} < 0$ be sufficiently small such that the first order terms dominates the remainder in the Taylor series.  We have $\delta < 0$. Define
		\begin{align*}
			\epsilon_i &= \tilde\epsilon_i - \frac{\delta}{|\I_\ell|} > \tilde \epsilon_i, \quad i \in \I_\ell, \\
			\epsilon_j &= \tilde\epsilon_j, \quad i \in \I_\ell.
		\end{align*}
		Note that $\frac{\partial C_{i,j}(\bm p)}{\partial p_i}$ is strictly positive, then the first order term is strictly positive for any $(i,j)$ pairs, and hence $\Gamma_{\thetabf}(\bm p + \bm\epsilon) > \Gamma_{\thetabf}(\bm p)$.
		\item If $\sum_{i\in\mathcal{I}_\ell}p_i^2 < \sum_{j\in\I^c_\ell}p_j^2$, then let $\epsilon_{i_1} > 0$ be sufficiently small such that the first order terms dominates the remainder in the Taylor series.  We have $\delta < 0$. Define
		\begin{align*}
			\epsilon_i &= \tilde\epsilon_i - \frac{\delta}{|\I_\ell|} > \tilde \epsilon_i, \quad i \in \I_\ell, \\
			\epsilon_j &= \tilde\epsilon_j, \quad i \in \I_\ell.
		\end{align*}
		Then the first order term is strictly positive for any $(i,j)$ pairs, and hence $\Gamma_{\thetabf}(\bm p + \bm\epsilon) > \Gamma_{\thetabf}(\bm p)$.
	\end{enumerate}
	
	Finally, we remark that it is straightforward to generalize the above result to Gaussian bandits with heterogeneous variances.  
	To this end, note that 
	\[C_{i,j}(\bm p) = \frac{(\theta_i - \theta_j)^2}{2}\frac{p_i p_j}{\sigma_j^2p_i + \sigma_i^2p_j}\]
	and
	\begin{align*}
		\frac{\partial C_{i,j}(\bm p)}{\partial p_i} 
		= \frac{(\theta_i - \theta_j)^2}{2}\frac{\sigma_i^2p_j^2}{(\sigma_j^2p_i + \sigma_i^2p_j)^2}, \quad \frac{\partial C_{i,j}(\bm p)}{\partial p_j} 
		= \frac{(\theta_i - \theta_j)^2}{2}\frac{\sigma_j^2p_i^2}{(\sigma_j^2p_i + \sigma_i^2p_j)^2}. 
	\end{align*}
	Consequently, we have
	\[\frac{\frac{\partial C_{i,j}(\bm p)}{\partial p_i}}{\frac{\partial C_{i,j}(\bm p)}{\partial p_j} } = \frac{p_j^2/\sigma_j^2}{p_i^2/\sigma_i^2}.\]
	The rest follows exactly as the case with equal variances.  We omit the details. 
\end{proof}

In Proposition \ref{prop:SS_balance}, we show that for Gaussian bandits, the optimal solution must balance the top and bottom arms in all connected subgraphs in terms of the sum-of-squared allocations.
For general reward distribution, we have similar results for each connected subgraph.
However, the form of this necessary condition cannot be easily simplified.

To derive the general result, we mirror the proof of Proposition \ref{prop:SS_balance}.
Consider any connected subgraph $
\mathcal G_\ell = \mathcal G(\I_\ell,\J_\ell,\mathcal E_\ell)$ of the bipartite graph induced by the optimal solution $\bm p^*$.
By Proposition \ref{prop:information_balance}, for any arm $i \in \I \cup \J$, there must exist an edge $e \in \mathcal E$ such that $i$ is incident to $e$.
Hence, we have $\I_\ell\neq \varnothing$ and $\J_\ell \neq \varnothing$.

Take any $i_1 \in \I_\ell$ as an \textit{anchor point}.  By the connectedness of $G_\ell$, for any $j \in \J_\ell$, there exists a path $x_{i_1,j} = (i_1, j_1, i_2, j_2, \dots, j_r = j)$ connecting $i_1$ and $j$.
Along the path, we perform Taylor's expansion and derive that
\begin{equation*}
	C_{i,j}(\bm p + \bm\epsilon) = C_{i,j}(\bm p) + \frac{\partial C_{i,j}(\bm p)}{\partial p_i}\epsilon_i + \frac{\partial C_{i,j}(\bm p)}{\partial p_j}\epsilon_j + o(\|\bm\epsilon\|).
\end{equation*}
By Lemma \ref{lm:C_ij_derivative}, we have 
\[
\frac{\partial C_{i,j}(\bm p^*)}{\partial p_i} = d(\theta_i, \bar{\theta}_{i,j})
\quad\text{and}\quad 
\frac{\partial C_{i,j}(\bm p^*)}{\partial p_j} = d(\theta_j, \bar{\theta}_{i,j}).
\]
Define $\tilde\epsilon_{i_1}$ and $\tilde\epsilon_{j_1}$ such that the first order terms in the Taylor's series vanish
\begin{equation*}
	d(\theta_{i_1}, \bar{\theta}_{i_1,j_1})\tilde\epsilon_{i_1} + d(\theta_{j_1}, \bar{\theta}_{i_1,j_1})\tilde\epsilon_{j_1} = 0.
\end{equation*}
Hence, we can write
\[\tilde\epsilon_{j_1} = -\tilde\epsilon_{i_1} \frac{d(\theta_{i_1}, \bar{\theta}_{i_1,j_1})}{d(\theta_{j_1}, \bar{\theta}_{i_1,j_1})}.\]
Repeating this procedure along the path connecting $i_1$ and $j$, we have
\begin{equation*}
	\tilde\epsilon_{j} = \tilde\epsilon_{j_r} = - \tilde\epsilon_{i_1} S_{i_1\to j},
\end{equation*}
where
\begin{equation}\label{eq:S_i1_to_j}
	S_{i_1\to j}  = \frac{d(\theta_{i_r}, \bar{\theta}_{i_r,j_r})}{d(\theta_{j_r}, \bar{\theta}_{i_r,j_r})} \cdots\frac{d(\theta_{j_1}, \bar{\theta}_{i_2,j_1})}{d(\theta_{i_2}, \bar{\theta}_{i_2,j_1})} \frac{d(\theta_{i_1}, \bar{\theta}_{i_1,j_1})}{d(\theta_{j_1}, \bar{\theta}_{i_1,j_1})}.
\end{equation}

Similarly, we have for any $i \in \I_\ell$, there exists a path $p_{i_1,i} = (i_1, j'_1, i'_2, j'_2, \dots, i'_{r'} = i)$ connecting $i_1$ and $i$.  We can then derive 
\begin{equation*}
	\tilde\epsilon_{i} = \tilde\epsilon_{i'_{r'}} = \tilde\epsilon_{i_1} S_{i_1\to i},
\end{equation*}
where
\begin{equation}\label{eq:S_i1_to_i}
	S_{i_1\to i} = \frac{d(\theta_{j'_{r'-1}}, \bar{\theta}_{i'_{r'},j'_{r'-1}})}{d(\theta_{i'_{r'}}, \bar{\theta}_{i'_{r'},j'_{r'-1}})} \cdots\frac{d(\theta_{j'_1}, \bar{\theta}_{i'_2,j'_1})}{d(\theta_{i'_2}, \bar{\theta}_{i'_2,j'_1})} \frac{d(\theta_{i'_1}, \bar{\theta}_{i'_1,j'_1})}{d(\theta_{j'_1}, \bar{\theta}_{i'_1,j'_1})}.
\end{equation}
We adopt the convention that $S_{i_1\to i_1} = 1$.
For Gaussian bandits, this simplifies to
\[
S_{i_1\to j} = \frac{(p^*_j)^2}{(p^*_{i_1})^2}
\quad\text{and}\quad
S_{i_1\to i} = \frac{(p^*_i)^2}{(p^*_{i_1})^2}.
\]
For general distribution, we cannot easily simplify $S_{i_1\to j}$ and $S_{i_1\to i}$.
With the same argument as in the proof of Proposition \ref{prop:SS_balance}, we have the following proposition.

\begin{proposition}[Overall balance in connected subgraphs] \label{prop:general_SS}
	Consider a bandit with general exponential family rewards.
	Let $\mathcal G$ be the bipartite graph induced by the optimal solution $\bm p^*$ and let $\mathcal G_{\ell}$ be any connected subgraph of $\mathcal G$.
	For any anchor point $i_1 \in \I_\ell$, $\mathcal G_\ell$ balances the top and bottom arms
	\begin{equation}\label{eq:SS_general}
		\sum_{i \in \I_\ell} S_{i_1\to i} = \sum_{j \in \J_\ell} S_{i_1\to j}.
	\end{equation}
\end{proposition}

\subsection{Proofs of properties of the optimal solution in Appendix \ref{sec:properties_optimal_solution} } \label{app:proof_uniqueness}
\begin{proof}[Proof of Lemma \ref{lm:uniqueness} (Uniqueness)] 
	By Lemma \ref{lm:C_ij_derivative}, $C_{i,j}(\cdot)$ is a concave function. Hence the optimal allocation problem \eqref{eq:optimal_allocation} is a concave programming.
	Consequently, the optimal set is convex, see \citet[Chapter 2]{boyd2004convex}.
	
	We prove by contradiction. Suppose there exist two optimal solutions $\bm p' \neq \bm p''$, then 
	\[\bm p(\varepsilon) =  (1-\varepsilon)\bm p' +  \varepsilon \bm p''\]
	is also an optimal solution for any $\varepsilon \in (0,1)$.
	To facilitate discussion, we define the index set where $\bm p'$ differs from $\bm p''$
	\[\mathcal{D} = \{i\in[K]: p_i' \neq p_i''\} \neq \varnothing.\]
	
	Now, consider the optimal solution $\bm p'$. 
	By the information balance \eqref{eq:information_balance}, there exist a set $\mathcal{B} \subset \I\times\J$ such that 
	\[C_{i,j}(\bm p') = \Gamma_{\bm\theta}(\bm p'), \quad \forall (i,j) \in \mathcal{B}, \quad \text{and} \quad C_{i',j'}(\bm p') > \Gamma_{\bm\theta}(\bm p'), \quad \forall (i',j') \notin \mathcal{B}.\]
	For any collection $\mathcal{C}$ of $(i,j)$ pairs, we define
	\[\mathcal{C}_{\cap \mathcal{D}} \triangleq \{(i,j) \in \mathcal{C} : \{i,j\} \cap  \mathcal{D} \neq \varnothing\}.\]
	
	\begin{enumerate}
		\item For any $(i,j) \in (\mathcal{B}\backslash \mathcal{B}_{\cap\mathcal{D}})$, we have
		\begin{equation}\label{eq:unique_helper_1}
			C_{i,j}(\bm p(\varepsilon)) = C_{i,j}(\bm p') = \Gamma_{\bm \theta}(\bm p'). 
		\end{equation}
		\item Next, consider any $(i,j) \in \mathcal{B}^c$. Since $|\mathcal{B}^c| < \infty$, we can define
		\begin{equation*}
			\Delta^{\Gamma_{\bm\theta}(\bm p')}_{\text{min}} \triangleq \min_{(i,j) \notin \mathcal{B}}C_{i,j}(\bm p') - \Gamma_{\bm\theta}(\bm p') > 0.
		\end{equation*}
		By the continuity of $C_{i,j}(\cdot)$, there exists a sufficiently small $\varepsilon$ such that 
		\begin{equation*}
			\min_{(i,j) \notin \mathcal{B}}C_{i,j}(\bm p(\varepsilon)) - \Gamma_{\bm\theta}(\bm p') > \Delta^{\Gamma_{\bm\theta}(\bm p')}_{\text{min}}/2 > 0,
		\end{equation*}
		then
		\begin{equation}\label{eq:unique_helper_2}
			\min_{(i,j) \notin \mathcal{B}}C_{i,j}(\bm p(\varepsilon)) > \Gamma_{\bm\theta}(\bm p') = \Gamma_{\bm\theta}(\bm p(\varepsilon)),
		\end{equation}
		where the last equality follows from the fact that both $\bm p'$ and $\bm p(\varepsilon)$ are optimal solutions.
		\item Finally, consider any $(i,j) \in \mathcal{B}_{\cap\mathcal{D}}$. Invoking Lemma \ref{lmm:contour}, we conclude that $\varepsilon$ can be selected such that
		\begin{equation}\label{eq:unique_helper_3}
			C_{i,j}(\bm p(\varepsilon)) \neq C_{i,j}(\bm p'), \quad \forall (i,j)\in\mathcal{B}_{\cap\mathcal{D}}.
		\end{equation}
	\end{enumerate}
	We have the following contradictions.
	\begin{enumerate}
		\item Suppose that there exists some $(i_1,j_1) \in \mathcal{B}_{\cap\mathcal{D}}$ such that 
		\[C_{i_1,j_1}(\bm p(\varepsilon)) < C_{i_1,j_1}(\bm p') = \Gamma_{\bm\theta}(\bm p').\]
		Then
		\[\Gamma_{\bm\theta}(\bm p(\varepsilon)) \le C_{i_1,j_1}(\bm p(\varepsilon)) < \Gamma_{\bm\theta}(\bm p').\]
		This contradicts with the fact that $\bm p(\varepsilon)$ is an optimal solution.
		\item Suppose that for all $(i,j) \in \mathcal{B}_{\cap\mathcal{D}}$, we have 
		\[C_{i_1,j_1}(\bm p(\varepsilon)) > C_{i_1,j_1}(\bm p') = \Gamma_{\bm\theta}(\bm p').\] 
		Recall that $\mathcal{D} \neq \varnothing$, then there exist $i\in\mathcal{D}$ or $j\in\mathcal{D}$. Suppose $i_2 \in\I\cap\mathcal{D}\neq \varnothing$. Combining \eqref{eq:unique_helper_1},\eqref{eq:unique_helper_2} and \eqref{eq:unique_helper_3}, we have
		\[\min_{j \in \J} C_{i_2,j}(\bm p(\varepsilon)) > \Gamma_{\bm\theta}(\bm p').\]
		Invoking the information balance in Proposition \ref{prop:information_balance} we conclude that
		\[\Gamma_{\bm\theta}(\bm p(\varepsilon)) > \Gamma_{\bm\theta}(\bm p').\]
		This contradicts with the fact that $\bm p'$ is an optimal solution.  
		Suppose $j_2 \in\J\cap\mathcal{D}\neq \varnothing$, the proof follows similarly.
	\end{enumerate}
	
	In summary, we have uniqueness of the optimal solution to \eqref{eq:optimal_allocation}.
\end{proof}

\begin{proof}[Proof of Lemma \ref{lmm:positive_sol} (Positivity)]
	Consider any fixed $i\in \I$ and any feasible solution $\bm p$. If $p_i = 0$ for some $i \in \I$, then $C_{i,j}(\bm p) = p_i d(\theta_i, \bar{\theta}_{i,j}) + p_j d\left(\theta_j, \frac{p_i \theta_i + p_j \theta_j}{p_i + p_j}\right) = 0$ holds for all $j\in \I^c$. Similarly, if $p_j = 0$ for some $j \in \I^c$, then $C_{i,j} = 0$ holds for all $i\in\I$. However, there exists $\bm p = (1/K, 1/K, \dots, 1/K)$ such that $C_{i,j} = \frac{1}{K}d\left (\theta_i, \frac{\theta_i + \theta_j}{2}\right) +  \frac{1}{K}d\left (\theta_j, \frac{\theta_i + \theta_j}{2}\right)>0$ holds for all $i\in\I$ and $j\in\I^c$. Thus, the optimal solution cannot have $p_i = 0$ for any $i\in[K]$.
\end{proof}

\begin{proof}[Proof of Lemma \ref{prop: monotonicity} (Monotonicity)]
	For the ease of exposition, we introduce the following function
	\begin{equation*}
		g_{i,j}(x) =d\left(\theta_i, \frac{\theta_i + x\theta_j}{1+x}\right) + x d\left(\theta_j, \frac{\theta_i + x\theta_j}{1+x}\right), \quad i \in \I, j \in \J.
	\end{equation*}
	Then $C_{i,j}(\bm p)$ can be written as
	\begin{equation*}
		C_{i,j}(\bm p) = p_i g_{i,j}\left(\frac{p_j}{p_i}\right).
	\end{equation*}
	
	At the optimal solution $\bm p^*$, consider any $j$, $j'\in \J$ such that $\theta_{j'} \geq \theta_{j}$. 
	Proposition \ref{prop:information_balance} implies that there exist a $i\in\I$ such that 
	\begin{equation*}
		C_{i, j}(\bm p^*) = \Gamma^*_{\thetabf} \leq C_{i, j'}(\bm p^*).
	\end{equation*}
	Note that we must have $p^*_{i} > 0$, the inequality above implies that
	\begin{equation}
		\label{helper1}
		g_{i,j}\left(\frac{p^*_j}{p^*_i}\right) \leq  g_{i,j'}\left(\frac{p^*_{j'}}{p^*_i}\right).
	\end{equation}
	
	By Lemma \ref{lm:gij}, for any $i, j$ and $j'$ such that $\theta_i > \theta_{j'} \geq \theta_{j}$, we have 
	\begin{equation}
		\label{eq: gij decreasing}
		g_{i,j}(x) \geq g_{i,j'}(x) \quad\text{for all } x.
	\end{equation}
	Plugging $x=\frac{p^*_j}{p^*_i}$ into \eqref{eq: gij decreasing}, we have, 
	\begin{equation}
		\label{helper2}
		g_{i,j}\left( \frac{p^*_j}{p^*_i}\right) \geq g_{i,j'}\left(\frac{p^*_j}{p^*_i}\right).
	\end{equation}
	Combining  \eqref{helper1}, \eqref{helper2} and $g_{i,j'}(x) > 0$, we conclude that
	$p_{j'}^*  \geq p_{j}^*$.
	This shows the monotonicity of $\bm p^*$ within the bottom arms $\J$.
	
	For the top arms $\I$, we introduce
	\begin{equation*}
		g_{j,i}(x) = d\left(\theta_j, \frac{\theta_j + x\theta_i}{1+x}\right) + x d\left(\theta_i, \frac{\theta_j + x\theta_i}{1+x}\right).
	\end{equation*}
	Then $C_{i,j}$ can be written as 
	\begin{equation*}
		C_{i,j}(\bm p) = p_j g_{j,i}\left(\frac{p_i}{p_j}\right).
	\end{equation*}
	From the proof of Lemma \ref{lm:gij} (See \eqref{helper: h monontonicty} and replace $\theta_i$ with $\theta_j$ in $h$), we have 
	\[g_{j,i}'(x) = d \left(\theta_i, \frac{\theta_j + x\theta_i}{1+x}\right)> 0\] 
	and, for any $\theta_{i} \geq \theta_{i'} > \theta_j$, we have
	\begin{equation}
		\label{helper: gji increasing}
		g_{j, i}(x) \geq g_{j,i'}(x) \quad\text{for all } x. 
	\end{equation}
	The rest of the proof mirrors the proof of the monotonicity within the bottom arms. 
	At the optimal solution $\bm p^*$, consider any $i$, $i'\in \I$ such that $\theta_{i} \geq \theta_{i'}$. Proposition \ref{prop:information_balance} also implies that there exists a $j\in \J$ such that 
	\begin{equation*}
		C_{i, j}(\bm p^*) = \Gamma^*_{\thetabf} \le C_{i', j}(\bm p^*).
	\end{equation*}
	Note that we must have $p^*_{j} > 0$, the inequality above implies that
	\begin{equation}
		\label{helper3}
		g_{j, i}\left(\frac{p^*_i}{p^*_{j}}\right) \leq  g_{j,i'}\left(\frac{p^*_{i'}}{p^*_{j}}\right).
	\end{equation}
	Plugging $x=\frac{p^*_i}{p^*_j}$ into \eqref{helper: gji increasing}, we have, 
	\begin{equation}
		\label{helper4}
		g_{j,i}\left( \frac{p^*_i}{p^*_j}\right) \geq g_{j,i'}\left(\frac{p^*_i}{p^*_j}\right).
	\end{equation}
	Combining  \eqref{helper3}, \eqref{helper4} and $g_{j,i'}(x) > 0$, we conclude that
	$p_{i'}^*  \geq p_{i}^*$.
	This shows the monotonicity of $\bm p^*$ within the top arms $\I$. The proof is complete. 
\end{proof}

\newpage
\section{Optimality conditions based only on the allocation vector}
\label{app:optimality_condition_on_p}
Optimality conditions that depends only on the allocation vector $\bm p$ can be derived for the best-$k$-arm identification problem.
However, we show in this section that these optimality conditions are complicated and impractical for algorithm design.

\subsection{A (conditional) sufficient condition}
To begin, we present a sufficient condition for a large set of problem instances $\bm\theta$.
For the ease of notation, we re-label the arm with the $\ell$-th largest mean as arm $i_\ell$, for $\ell = 1,\dots, k$ and re-label the arm with the $(\ell+k)$-th largest mean as arm $j_\ell$, for $\ell = 1,\dots, K-k$.
It may be intriguing to think that the optimal solution should gather equal statistical evidence to distinguish (i) the arm $i_k$ from any bottom arm $j$; and (ii) the arm $j_1$ from any top arm $i$.
The next proposition show that this intuitive balancing rule (together with the balance of sum-of-squared allocation) is optimal under certain condition.
To simplify notation, we define $\bar \theta^*_{i,j} \triangleq\bar \theta_{i,j}(\bm p^*)$ for all $(i,j) \in \I\times\J$.
\begin{proposition}
	\label{prop:necessary_sufficient_condition}
	Consider any problem instance such that the optimal solution $\bm p^*$ satisfies	\begin{equation}\label{eq:key_assumption}
		1-\sum_{j \in \J\backslash \{j_1\}}\frac{d(\theta_{i_k}, \bar \theta^*_{i_k,j})}{d(\theta_{j}, \bar \theta^*_{i_k,j})}  > 0.
	\end{equation}
	Then $\bm p^*$ is the optimal solution to (\ref{eq:optimal_allocation})
	if 
	\begin{align}
		&\Gamma^*_{\thetabf} = C_{i_k,j}(\bm{p}^*) = C_{i,j_{1}}(\bm p^*), \quad \forall i\in \I, \forall j \in \J, \quad \text{and}\label{eq:optimality_condition_balancing} \\
		&\frac{d(\theta_{i_k}, \bar\theta^*_{i_k,j_1})}{d(\theta_{j_1}, \bar\theta^*_{i_k,j_1}) }\sum_{i\in\I} \frac{d(\theta_{j_1}, \bar\theta^*_{i,j_1}) }{d(\theta_{i}, \bar\theta^*_{i,j_1})}
		= \sum_{j \in \J}	\frac{d(\theta_{i_k}, \bar\theta^*_{i_k,j}) }{d(\theta_{j}, \bar\theta^*_{i_k,j})}. \label{eq:ss_balancing}
	\end{align}
\end{proposition}
Complete characterization of sufficient and necessary conditions based only on $\bm p$ can be derived following the lines of Proposition \ref{prop:necessary_sufficient_condition}, see Appendix \ref{sec:sufficient}.
However, we caution that such results rely on assumptions similar to (\ref{eq:key_assumption}), which is non-trivial to check.
Hence, developing a sampling rule based on tracking the sufficient condition can be challenging.
We address this challenge by introducing the dual variable, and analyze the KKT conditions.

\begin{remark}
	Condition \eqref{eq:optimality_condition_balancing} implies that the bipartite graph is connected.
	By setting the anchor point as $i_k$ in Proposition \ref{prop:general_SS}, we see that \eqref{eq:ss_balancing} is equivalent to the balancing of top and bottom arms.
\end{remark}

\begin{remark}
	For general bandits, condition (\ref{eq:key_assumption}) always holds when (i) $|\I| \le 2$ and $|\J| \le 2$; and (ii) $|\I| = 1$ or $|\J| = 1$. 
	When condition (\ref{eq:key_assumption}) holds, there is only one connected subgraph.
	For Gaussian bandits with equal variance, the assumption (\ref{eq:key_assumption}) reduces to $(p^*_{i_k})^2 > \sum_{j \in \J\backslash \{j_1\}} (p^*_j)^2$; and (\ref{eq:ss_balancing}) reduces to the balancing of the sum-of-squared allocations.
	Recall the monotonicity in Lemma \ref{prop: monotonicity}, we have $p_{j_1} \ge p_j$ for $j \neq j_1$.
	In general, $p^*_{j_1} - p^*_j$ increases as the gap $(\theta_{j_1} - \theta_j)$ increases. 
	Roughly speaking, when $(\theta_{j_1} - \theta_j)$ is large enough with respect to $(\theta_{i_k} - \theta_{j_1})$, condition (\ref{eq:key_assumption}) will hold.
	This suggest that problem instances closer to the slippage configuration are the most subtle instances in terms of the structure of the optimal solution.
\end{remark}

\subsection{A Case study for 5 choose 2: A pathway to complete characterization of optimality based on the allocation vector} \label{sec:sufficient}

In Proposition \ref{prop:necessary_sufficient_condition}, we characterized the sufficient conditions for optimality under the assumption of (\ref{eq:key_assumption}).
We now discuss how to obtain complete characterization.

We assume that the mean rewards are mutually different without loss of generality (for the optimal allocation problem).
Otherwise, we can combine arms with the same mean rewards and evenly distribute the allocation among them.

To facilitate our discussion, we present $C_{i,j}(\bm p^*)$ as a matrix in $\mathbb R^{|\I|\times|\J|}$, where the $(i,j)$-th entry is ``$=$'' if $C_{i,j}(\bm p^*) = \Gamma^*_{\thetabf}$ and ``$>$'' if $C_{i,j}(\bm p^*) > \Gamma^*_{\thetabf}$.
The following matrix represents the case discussed in Proposition \ref{prop:necessary_sufficient_condition}.
\begin{center}
	\begin{tabular}{ c|cccc } 
		arms    & $j_1$ & $j_2$ & $\cdots$  & $j_{K-k}$   \\
		\hline
		$i_1$     & $=$   & $>$   & $\cdots$  & $>$   \\
		$i_2$     & $=$   & $>$   & $\cdots$  & $>$   \\    
		$\vdots$   & $\vdots$   & $\vdots$  & $\vdots$    & $\vdots$      \\
		$i_{k-1}$   & $=$   & $>$   & $\cdots$  & $>$   \\
		$i_k$     & $=$   & $=$   & $\cdots$  & $=$   \\
	\end{tabular}
\end{center}
Recall the assumption in Proposition \ref{prop:necessary_sufficient_condition}, i.e.,
\begin{equation*}
	1-\sum_{j \in \J\backslash \{j_1\}}\frac{d(\theta_{i_k}, \bar \theta^*_{i_k,j})}{d(\theta_{j}, \bar \theta^*_{i_k,j})}  > 0.
\end{equation*}
In general, the $=$'s locations can be quite different from the clean structure in the matrix above.

We illustrate with an example and discuss the sufficient conditions for it.
Consider choosing the best-$2$ arms among $5$ competing arms. 
Without loss of generality, we assume that the arms are sorted in descending order of the mean rewards, i.e., $\I = \{1,2\}$ and $\J = \{3,4,5\}$.
We may have the following scenario.
\begin{center}
	\begin{tabular}{ c|cccc } 
		arms    & $3$   & $4$   & $5$   \\
		\hline
		$1$     & $=$   & $=$   & $>$   \\
		$2$     & $>$   & $=$   & $=$   \\    
	\end{tabular}
\end{center}
Mirroring the proof of Proposition \ref{prop:necessary_sufficient_condition}, we have the following sufficient conditions for this case.

\begin{proposition}\label{thm:sufficiency_2_5}
	Consider any problem instance such that the optimal solution $\bm p^*$ satisfies
	\begin{equation}\label{eq:assumption_2_5}
		\begin{split}
			&1 - \frac{d(\theta_{2}, \bar\theta^*_{2,5})}{d(\theta_{5}, \bar\theta^*_{2,5}) } > 0, \\
			&1 - \frac{d(\theta_{4}, \bar\theta^*_{2,4})}{d(\theta_{2}, \bar\theta^*_{2,4}) } + \frac{d(\theta_{4}, \bar\theta^*_{2,4})}{d(\theta_{2}, \bar\theta^*_{2,4}) }\frac{d(\theta_{2}, \bar\theta^*_{2,5})}{d(\theta_{5}, \bar\theta^*_{2,5}) } > 0.
		\end{split}
	\end{equation}
	Then $\bm p^*$ is the optimal solution to (\ref{eq:optimal_allocation})
	if 
	\begin{equation} \label{eq:balance_C_2_5}
		\Gamma^*_{\thetabf} = C_{1,3}(\bm p^*) = C_{1,4}(\bm p^*) = C_{2,4}(\bm p^*) = C_{2,5}(\bm p^*)
	\end{equation}
	and
	\begin{equation} \label{eq:balance_SS_2_5}
		1 + \frac{d(\theta_{1}, \bar\theta^*_{1,4}) }{d(\theta_{4}, \bar\theta^*_{1,4})}\frac{d(\theta_{4}, \bar\theta^*_{2,4})}{d(\theta_{2}, \bar\theta^*_{2,4}) } = \frac{d(\theta_{1}, \bar\theta^*_{1,3})}{d(\theta_{3}, \bar\theta^*_{1,3}) } + \frac{d(\theta_{1}, \bar\theta^*_{1,4}) }{d(\theta_{4}, \bar\theta^*_{1,4})} + \frac{d(\theta_{1}, \bar\theta^*_{1,4}) }{d(\theta_{4}, \bar\theta^*_{1,4})}\frac{d(\theta_{4}, \bar\theta^*_{2,4})}{d(\theta_{2}, \bar\theta^*_{2,4}) }\frac{d(\theta_{2}, \bar\theta^*_{2,5})}{d(\theta_{5}, \bar\theta^*_{2,5}) }.
	\end{equation}  
\end{proposition}
\begin{remark}
	For this scenario, the bipartite graph is connected, and the condition (\ref{eq:balance_SS_2_5}) is equivalent to the balancing of top and bottom arms (by setting arm $1$ as the anchor point in Proposition \ref{prop:general_SS}).
\end{remark}

\begin{remark}[Complete characterization of the case with $(K,k) = (5,2)$]\label{rmk:full_characterization_2_5}
	We consider Gaussian bandit with $(K,k) = (5,2)$.
	For simplicity, we consider the problem instances where the mean rewards are mutually different.
	Recall that we have the balancing of top and bottom arms in the entire bipartite graph in \eqref{eq:overall_balance}, i.e., 
	\begin{align*}
		(p^*_1)^2 + (p^*_2)^2 = (p^*_3)^2 + (p^*_4)^2 + (p^*_5)^2.
	\end{align*}
	By the monotonicity in Lemma \ref{prop: monotonicity}, we must have 
	\begin{align*}
		(p^*_2)^2 > (p^*_5)^2,
	\end{align*}
	otherwise, $(p^*_3)^2 \ge (p^*_4)^2 \ge (p^*_5)^2 \ge (p^*_2)^2 \ge (p^*_1)^2$, which contradicts with \eqref{eq:overall_balance}.
	
	Note that the condition \eqref{eq:assumption_2_5} is equivalent to 
	\begin{align*}
		(p^*_2)^2 > (p^*_5)^2 \quad\text{and}\quad (p^*_2)^2 < (p^*_4)^2 + (p^*_5)^2.
	\end{align*}
	Since we must have $(p^*_2)^2 > (p^*_5)^2$, condition \eqref{eq:assumption_2_5} reduces to 
	\begin{align*}
		(p^*_2)^2 < (p^*_4)^2 + (p^*_5)^2 .
	\end{align*}
	Recall that condition (\ref{eq:key_assumption}) reduces to
	\begin{align*}
		(p^*_2)^2 > (p^*_4)^2 + (p^*_5)^2.
	\end{align*}   
	Hence the two scenarios are mutually exclusive.
	Finally, recall that Example \ref{ex:disconnected_graph} corresponds to $\Gamma^*_{\thetabf} = C_{1,3}(p^*_{1}, p^*_3) = C_{2,4}(p^*_{2}, p^*_4) = C_{2,5}(p^*_{2}, p^*_5)$ and
	\begin{align}\label{eq:assumption_2_5_equal}
		(p^*_2)^2 = (p^*_4)^2 + (p^*_5)^2 \quad\text{and}\quad (p^*_1)^2 = (p^*_3)^2,
	\end{align}   
	which corresponds to
	\begin{center}
		\begin{tabular}{ c|cccc } 
			arms    & $3$   & $4$   & $5$   \\
			\hline
			$1$     & $=$   & $>$   & $>$   \\
			$2$     & $>$   & $=$   & $=$   \\    
		\end{tabular}
	\end{center}
	Combining (\ref{eq:key_assumption}), (\ref{eq:assumption_2_5}) and (\ref{eq:assumption_2_5_equal}), we have now fully characterized the sufficient conditions for the case of choosing $2$ arms among $5$, which we summarize in Table \ref{tab:5_2}.
	
	\begin{table}[htbp]
		\centering
		\begin{tabular}{c|c|c|c}
			\toprule
			Condition & $(p^*_2)^2 > (p^*_4)^2 + (p^*_5)^2$ & $ (p^*_2)^2 < (p^*_4)^2 + (p^*_5)^2$ & $ (p^*_2)^2 = (p^*_4)^2 + (p^*_5)^2$\\
			\midrule
			Information balance & \begin{tabular}{ c|cccc } 
				arms    & $3$   & $4$   & $5$   \\
				\hline
				$1$     & $=$   & $>$   & $>$   \\
				$2$     & $=$   & $=$   & $=$   \\    
			\end{tabular}
			&   \begin{tabular}{ c|cccc } 
				arms    & $3$   & $4$   & $5$   \\
				\hline
				$1$     & $=$   & $=$   & $>$   \\
				$2$     & $>$   & $=$   & $=$   \\    
			\end{tabular}
			&    \begin{tabular}{ c|cccc } 
				arms    & $3$   & $4$   & $5$   \\
				\hline
				$1$     & $=$   & $>$   & $>$   \\
				$2$     & $>$   & $=$   & $=$   \\    
			\end{tabular} \\
			\midrule
			\multirow{2}{*}{Overall balance} & \multirow{2}{*}{$\displaystyle\sum_{i\in\mathcal{I}}(p^{*}_i)^2 = \sum_{j\in\I^c}(p^{*}_j)^2$} & \multirow{2}{*}{$\displaystyle\sum_{i\in\mathcal{I}}(p^{*}_i)^2 = \sum_{j\in\I^c}(p^{*}_j)^2$} & $(p^*_1)^2 = (p^*_3)^2$
			\\
			&&&       $ (p^*_2)^2 = (p^*_4)^2 + (p^*_5)^2$ \\
			\bottomrule
		\end{tabular}
		\caption{Complete characterization of the optimal solution for Gaussian bandits, choose 2 from 5 arms, with mutually distinctive mean rewards. }
		\label{tab:5_2}
	\end{table}
	
\end{remark}

We omit the discussion for general $(K,k)$ pair and instances with arms sharing the same mean. 
A similar analysis can be carried out.  
However, we are cautious that any elegant presentation can be derived due to the \emph{combinatorial} number of scenarios, even though conditions such as (\ref{eq:key_assumption}) and (\ref{eq:assumption_2_5}) can significantly reduce the number of scenarios to be discussed.

\subsection{Proof of Proposition \ref{prop:necessary_sufficient_condition}}
\label{proof of thm1}

\begin{proof}
	We restate the KKT conditions from Section \ref{app:proof_KKT} here for easier reference:
	\begin{subequations}
		\label{ap_eq:KKT}
		\begin{align}
			& -1 + \sum_{i\in \I} \sum_{j\in \I^c} \mu_{i,j} = 0\label{ap_eq:KKT_stationarity0},\\
			& \lambda - \sum_{j\in \I^c} \mu_{i,j}d(\theta_i, \bar{\theta}_{i,j}) = 0, \quad\forall  i\in \I, 
			\label{ap_eq:KKT_stationarity1}
			\\
			& \lambda - \sum_{i\in \I} \mu_{i,j}d(\theta_j, \bar{\theta}_{i,j}) = 0,  \quad\forall  j\in \I^c, 
			\label{ap_eq:KKT_stationarity2}
			\\
			&\sum_{i\in[K]} p_i  - 1 = 0, \quad p_i \geq 0, \quad \forall i\in[K]\label{ap_eq:primal_simplex},
			\\
			& \phi - C_{i,j}(\bm p) \le 0, \quad\forall i\in \I, j\in \I^c, \label{ap_eq:Cij}
			\\ 
			& \mu_{i,j} \geq 0, \quad\forall i\in \I, j\in \I^c\label{ap_eq:dual_feasibility}, 
			\\
			& \sum_{i\in\I}\sum_{j\in\I^c} \mu_{i,j}(\phi - C_{i,j}(\bm p)) = 0. 
			\label{ap_eq:KKT_CS} 
		\end{align}
	\end{subequations}
	
	Let $\bm p$ be the optimal primal solution so that (\ref{ap_eq:primal_simplex}) holds automatically. 
	We aim to find a KKT solution such that 
	\begin{equation}\label{eq:KKT_sol_structure}
		\mu_{i,j} = 0, \quad \text{if  $i\neq i_k$ or $j\neq j_1$}.
	\end{equation}
	In particular, under (\ref{eq:KKT_sol_structure}), we find that the solution to (\ref{ap_eq:KKT_stationarity1}) and (\ref{ap_eq:KKT_stationarity2}) satisfies
	\begin{align}
		\mu_{i,j} &= 0, \quad \text{if  $i\neq i_k$ or $j\neq j_1$}, \notag\\
		\mu_{i_k, j}& = \frac{\lambda}{d(\theta_{j}, \bar{\theta}_{i_k,j})}, \quad \text{for $j\neq j_{1}$}, \notag\\
		\mu_{i, j_1} &= \frac{\lambda}{d(\theta_{i}, \bar{\theta}_{i,j_1})}, \hspace{13pt} \text{for $i\neq i_k$},\notag\\
		\mu_{i_k, j_1} &= \lambda \left(
		1-\sum_{j\in \I^c\backslash \{ j_1\}}
		\frac{d(\theta_{i_k}, \bar{\theta}_{i_k,j})}{d(\theta_{j}, \bar{\theta}_{i_k,j})}
		\right)\bigg/ 
		d(\theta_{i_k}, \bar{\theta}_{i_k, j_1}) \notag\\
		& = 
		\lambda \left(
		1-\sum_{i\in \mathcal{I}\backslash \{ i_k\}}
		\frac{d(\theta_{j_1}, \bar{\theta}_{i,j_1})}{d(\theta_{i}, \bar{\theta}_{i,j_1})}
		\right)\bigg/ 
		d(\theta_{j_1}, \bar{\theta}_{i_k, j_1}). \label{eq:mu_ij_expression}
	\end{align}
	where the multiplier $\lambda$ is determined by (\ref{ap_eq:KKT_stationarity0}), i.e.,
	\begin{equation}
		\sum_{j\neq j_1}\mu_{i_k, j} + \sum_{i\neq i_k}\mu_{i, j_1} +\mu_{i_k, j_1} = 1.
	\end{equation}
	
	For the dual feasibility condition (\ref{ap_eq:dual_feasibility}) to hold, we require condition (\ref{eq:key_assumption}) to hold, i.e.,
	\begin{equation}
		1-\sum_{j \in \J\backslash \{j_1\}}\frac{d(\theta_{i_k}, \bar \theta_{i_k,j})}{d(\theta_{j}, \bar \theta_{i_k,j})}  > 0,
	\end{equation}
	
	For (\ref{eq:mu_ij_expression}) to hold, we require (\ref{eq:ss_balancing}), i.e.,
	\begin{equation}
		\frac{d(\theta_{i_k}, \bar\theta_{i_k,j_1})}{d(\theta_{j_1}, \bar\theta_{i_k,j_1}) }\sum_{i\in\I} \frac{d(\theta_{j_1}, \bar\theta_{i,j_1}) }{d(\theta_{i}, \bar\theta_{i,j_1})}
		= \sum_{j \in \J}	\frac{d(\theta_{i_k}, \bar\theta_{i_k,j}) }{d(\theta_{j}, \bar\theta_{i_k,j})}.
	\end{equation}
	
	We have now checked that equations (\ref{ap_eq:KKT_stationarity0}), (\ref{ap_eq:KKT_stationarity1}), (\ref{ap_eq:KKT_stationarity2}), (\ref{ap_eq:primal_simplex}) and (\ref{ap_eq:dual_feasibility}) hold.
	
	For the primal variable (equivalent to objective value) $\phi$, 
	we let 
	\begin{equation}
		\phi = C_{i_k,j}(\bm p) = C_{i,j_1}(\bm p), \quad \forall (i,j)\in\I\times\J.
	\end{equation}
	By noting that $\mu_{i,j} = 0$ for $i\neq i_k$ and $j\neq j_1$, we have (\ref{ap_eq:KKT_CS}). 
	At last, when $\Gamma^*_{\thetabf} = \phi$, the inequality in (\ref{ap_eq:Cij}) also holds automatically.
	
	Hence, all the conditions in (\ref{ap_eq:KKT}) holds.
	The proof is complete. 
\end{proof}

\newpage
\section{Proof of Theorem \ref{thm:explicit_opt}: fixed-confidence sample complexity}
\label{app:fixed_confidence_LB}
For problem instance $\thetabf\in\Theta$, we define the set of alternative problem instances with different best-$k$-arm sets,
\begin{align*}
	\mathrm{Alt}(\thetabf) 
	&\triangleq 
	\left\{\varthetabf\in \Theta \,:\, \I(\varthetabf) \neq \I(\thetabf)\right\}\\
	&= \bigcup_{(i,j) \in \I(\bm\theta)\times\I^c(\bm\theta)}\left\{\varthetabf\in\Theta \,:\, \vartheta_i < \vartheta_j\right\}.
\end{align*}
For $(i,j) \in \I(\bm\theta)\times\I^c(\bm\theta)$, let $\mathrm{Alt}_{i,j}(\thetabf) \triangleq  \left\{\varthetabf\in\Theta \,:\, \vartheta_i < \vartheta_j\right\}$, and thus
\[
\mathrm{Alt}(\thetabf)  = \bigcup_{(i,j) \in \I(\bm\theta)\times\I^c(\bm\theta)} \mathrm{Alt}_{i,j}(\thetabf).
\]
We further define the \emph{discriminative information} between $\thetabf,\varthetabf\in\Theta$ under allocation $\bm{p}\in\mathcal{S}_K$
\[
D_{\bm{p}}(\bm{\theta},\bm{\vartheta}) = \sum_{i=1}^K p_i d(\theta_i,\vartheta_i).
\]
Then we can write each generalized Chernoff information $C_{i,k}(\bm p)$ defined in Equation \eqref{eq:def_C_ij} as follows,
\begin{lemma}
	\label{lm:C_ij_and_LB}
	For any $(i,j) \in \I(\thetabf)\times \I^c(\thetabf)$ and $\bm{p}\in \mathcal{S}_K$, 
	\[
	C_{i,j}(\bm p) = \inf_{\bm \vartheta \in \mathrm{Alt}_{i,j}(\thetabf) } D_{\bm p}(\bm \theta, \bm \vartheta).
	\]
\end{lemma}

We hence write $\Gamma_{\thetabf}^*$ defined in Equation \eqref{eq:optimal_allocation} as
\begin{align*}
	\Gamma^*_{\thetabf} &= \max_{\bm p \in\mathcal{S}_K}\min_{(i,j)\in\I(\thetabf)\times \I^c(\thetabf)}\inf_{\bm\vartheta \in \mathrm{Alt}_{i,j}(\thetabf)} D_{\bm p}(\bm\theta, \bm\vartheta), 
\end{align*}

\begin{proof}[Proof of Lemma \ref{lm:C_ij_and_LB}]
	Recall that 
	\begin{equation*}
		\inf_{\bm \vartheta: \vartheta_j > \vartheta_i} D_{\bm p}(\bm\theta, \bm\vartheta) = \inf_{\bm \vartheta: \vartheta_j > \vartheta_i} \sum_{i=1}^K p_i d(\theta_i, \vartheta_i). 
	\end{equation*}
	
	We have
	\begin{equation*}
		\begin{aligned}
			\inf_{\bm\vartheta: \vartheta_j > \vartheta_i} \sum_{\ell=1}^K p_\ell d(\theta_\ell, \vartheta_\ell)
			& = \inf_{\vartheta_j > \vartheta_i} \left\{
			p_i d(\theta_i, \vartheta_i)  + p_j d(\theta_j, \vartheta_j) \right\}+
			\sum_{\ell\neq i, j} \inf_{\vartheta_\ell}p_\ell d(\theta_\ell, \vartheta_\ell) \\
			&= \inf_{\vartheta_j > \vartheta_i} \left\{
			p_i d(\theta_i, \vartheta_i)  + p_j d(\theta_j, \vartheta_j) \right\}\hspace{23pt}\text{[setting $\vartheta_\ell = \theta_\ell$]}\\
			&=\inf_{\tilde{\theta}} \left\{
			p_i d(\theta_i, \tilde{\theta}) + p_j d(\theta_j, \tilde{\theta}) \right\}.\hspace{34pt}\text{[monotonicity of KL-divergence]}\\
		\end{aligned}
	\end{equation*}
	Taking derivative of the function inside the infimum with respect to $\tilde{\theta}$ and equating it to $0$, we have
	\begin{equation}
		\label{min C derivative}
		\frac{\mathrm{d}\eta(\tilde{\theta})}{\mathrm{d}\theta} \left(p_i (\tilde{\theta} - \theta_i)+p_j (\tilde{\theta} - \theta_j)\right)=0,
	\end{equation}
	where we use $d(\theta_i, \tilde{\theta}) = A(\eta(\tilde{\theta})) - A(\eta(\theta_i)) - \theta_i (\eta(\tilde{\theta})-\eta(\theta_i))$.
	We find that the infimum is achieved at $\bar{\theta}_{i,j} = (p_i \theta_i + p_j \theta_j)/(p_i + p_j)$. Hence, 
	\begin{equation*}
		\inf_{\bm \vartheta \in \mathrm{Alt}_{i,j}(\thetabf)} D_{\bm p}(\bm \theta, \bm \vartheta)= p_i d(\theta_i, \bar{\theta}_{i,j}) + p_j d(\theta_j, \bar{\theta}_{i,j}) = C_{i,j}(\bm p).
	\end{equation*}
\end{proof}

To complete the proof of Theorem \ref{thm:explicit_opt}, we also need to utilize Lemma 1 in \cite{kaufmann2016complexity}, which is rephrased below.
\begin{lemma}[Lemma 1 in \cite{kaufmann2016complexity}]
	Let $\bm \theta$ and $\bm \vartheta$ be two bandit models with $K$ arms such that for all $i\in[K]$, the distributions $\theta_i$ and $\vartheta_i$ are mutually absolutely continuous. For any almost-surely finite stopping time $\tau_\sigma$ with respect to $\mathcal{F}_t=\sigma\left(I_0, Y_{1,I_0},\ldots, I_{t-1}, Y_{t, I_{t-1}}\right)$,
	\begin{equation} \label{eq:LB_kaufmann}
		\sum_{i=1}^K \mathbb{E}_{\bm \theta} [N_{\tau_{\delta}, i}]d(\theta_i, \vartheta_i) \geq \sup_{\mathcal{E}\in \mathcal{F}_{\tau_\delta}}\mathrm{kl}(\mathbb{P}_{\bm \theta}(\mathcal{E}), \mathbb{P}_{\bm \vartheta} (\mathcal{E})),
	\end{equation}
	where $\mathrm{kl}(\cdot, \cdot)$ is the KL divergence of two Bernoulli distributions. 
\end{lemma}
\begin{proof}[Proof of Theorem \ref{thm:explicit_opt}]
	The $\delta$-correctness of an algorithm implies that, for any $\bm\theta\in\Theta$, $\bm\vartheta \in \mathrm{Alt}(\thetabf)$ and $\mathcal{E}\in\mathcal{F}_{\tau_\delta} $, we have $\mathbb{P}_{\bm \theta}(\mathcal{E}) > 1-\delta$ and $\mathbb{P}_{\bm \vartheta}(\mathcal{E})\leq \delta$. 
	Hence, (\ref{eq:LB_kaufmann}) implies that
	\begin{equation*}
		\sum_{i=1}^K \mathbb{E}_{\bm\theta}[N_{\tau_{\delta}, i}] d(\theta_i, \vartheta_i) \geq \mathrm{kl}(1-\delta, \delta),
	\end{equation*}
	where we use the monotonicity of the KL divergence of Bernoulli distribution, and thus
	\begin{equation*}
		\E_{\bm\theta}[\tau_\delta] \inf_{\bm\vartheta \in \mathrm{Alt}(\thetabf)}\sum_{i=1}^K \frac{\E_{\bm\theta}[N_{\tau_\delta, i}]}{\E_{\bm\theta}[\tau_\delta]} d(\theta_i, \theta'_i) \geq \mathrm{kl}(1-\delta, \delta).
	\end{equation*} 
	the lower bound of sample complexity $\E_{\bm\theta}[\tau_\delta] $ is
	\begin{equation*}
		\E_{\bm\theta}[\tau_\delta]  \geq \mathrm{kl}(1-\delta, \delta) \left(\inf_{\bm\vartheta\in \mathrm{Alt}(\thetabf)}\sum_{i=1}^K p_i d(\theta_i, \vartheta_i) \right)^{-1} .
	\end{equation*}
	It can be verify that $\lim_{\delta\to 0}\frac{\mathrm{kl}(1-\delta, \delta)}{\log(1/\delta)} = 1$.
	Hence, we have the asymptotic lower bound
	\begin{equation*}
		\liminf_{\delta\to 0}\frac{\E_{\bm\theta}[\tau_\delta] }{\log(1/\delta)} \geq  \left(\inf_{\bm\vartheta\in \mathrm{Alt}(\thetabf)}\sum_{i=1}^K p_i d(\theta_i, \vartheta_i) \right)^{-1}. 
	\end{equation*}
	Minimizing this lower bound is equivalent to the following optimization problem,
	\begin{align*}
		\max_{\bm p\in\mathcal{S}_K}\inf_{\bm\vartheta\in \mathrm{Alt}(\thetabf)} \sum_{i=1}^K p_i d(\theta_i, \vartheta_i) 
		& = \max_{\bm p\in\mathcal{S}_K}\inf_{\bm\vartheta\in \mathrm{Alt}(\thetabf)} D_{\bm p}(\bm\theta, \bm\vartheta) \\
		& = \max_{\bm p \in\mathcal{S}_K}\min_{(i,j)\in\I(\thetabf)\times \I^c(\thetabf)}\inf_{\bm\vartheta \in \mathrm{Alt}_{i,j}(\thetabf)} D_{\bm p}(\bm\theta, \bm\vartheta) \\
		& = \Gamma^*_{\thetabf}.
	\end{align*}
	This completes the proof.
\end{proof}

\newpage
\section{Proof of Theorem \ref{thm:KKT}} \label{app:proof_KKT}

\begin{proof}
	Let $\lambda \in \R$, $\bm \iota = \{\iota_i: i\in[K]\}$ and $\bm \mu = \{\mu_{i,j}: i \in \I, j \in \I^c\} $ be the Lagrangian multipliers corresponding to the equality constraint \eqref{eq:convex_formulation_01}, inequality constraints \eqref{eq:convex_formulation_03} and inequality constraints \eqref{eq:convex_formulation_02}, respectively.
	The Lagrangian is
	\begin{equation}\label{helper:Lagrangian}
		\mathcal{L}(\phi, \bm p, \lambda, \bm \iota, \bm \mu) = \phi - \lambda\left( \sum_{i\in[K]}p_i - 1\right)  + \sum_{i\in[K]} \iota_i p_i- \sum_{i\in\I} \sum_{j\in\I^c} \mu_{i,j} (\phi - C_{i,j}(\bm p)).
	\end{equation}
	Lemma \ref{lmm:positive_sol} simplifies the Lagrangian in the sense that $\bm \iota = \bm 0$, which follows from the complementary slackness condition for \eqref{eq:convex_formulation_03}.
	
	Using Lemma \ref{lm:C_ij_derivative}, the KKT conditions are given by
	\begin{subequations}
		\label{eq:KKT}
		\begin{align}
			& -1 + \sum_{i\in \I} \sum_{j\in \I^c} \mu_{i,j} = 0\label{eq:KKT_stationarity0},\\
			& \lambda - \sum_{j\in \I^c} \mu_{i,j}d(\theta_i, \bar{\theta}_{i,j}) = 0, \quad\forall  i\in \I, 
			\label{eq:KKT_stationarity1}
			\\
			& \lambda - \sum_{i\in \I} \mu_{i,j}d(\theta_j, \bar{\theta}_{i,j}) = 0,  \quad\forall  j\in \I^c, 
			\label{eq:KKT_stationarity2}
			\\
			&\sum_{i\in[K]} p_i  - 1 = 0, \quad p_i \geq 0, \quad \forall i\in[K]\label{eq:primal_simplex},
			\\
			& \phi - C_{i,j}(\bm p) \le 0, \quad\forall i\in \I, j\in \I^c, \label{eq:Cij}
			\\ 
			& \mu_{i,j} \geq 0, \quad\forall i\in \I, j\in \I^c\label{eq:dual_feasibility}, 
			\\
			& \sum_{i\in\I}\sum_{j\in\I^c} \mu_{i,j}(\phi - C_{i,j}(\bm p)) = 0. 
			\label{eq:KKT_CS} 
		\end{align}
	\end{subequations}

	The KKT conditions in \eqref{eq:KKT} is not immediately insightful for the design of optimal adaptive selection in that the key stationarity condition \eqref{eq:KKT_stationarity1} and \eqref{eq:KKT_stationarity2} are not directly written in terms of the allocation proportion $\bm p$.
	To obtain insights on algorithmic design and optimal adaptive selection, we reformulate the KKT conditions into a set of necessary and sufficient conditions that is explicitly expressed by the dual variables $\bm \mu$ and primal variables $\bm p$.
	
	To show the equivalent between (\ref{eq:KKT_equiv}) and (\ref{eq:KKT}), we first show that (\ref{eq:KKT}) implies (\ref{eq:KKT_equiv}).
	
	We claim that there must exist a $i'\in\I$ and a $j'\in\I^c$ such that $\phi = C_{i',j'}(\bm p)$ and hence $\phi = \min_{i\in\I}\min_{j\in\I^c} C_{i,j}(\bm p)$. 
	Together with \eqref{eq:KKT_CS}, this implies \eqref{eq:KKT_equiv_complementary_slackness}.
	Suppose otherwise, we have $\phi < C_{i,j}(\bm p)$ for all $i\in\I$ and $j\in\I^c$. 
	By the complementary slackness condition (\ref{eq:KKT_CS}), we have $\mu_{i,j} = 0$ for all $i\in\I$ and $j\in\I^c$, which yields contradiction with \eqref{eq:KKT_stationarity0}. Furthermore, by the complementary slackness condition (\ref{eq:KKT_CS}), $\phi = C_{i.j}(\bm p)$ whenever $\mu_{i,j} > 0$. Hence, we can write the stationarity conditions (\ref{eq:KKT_stationarity1}) and (\ref{eq:KKT_stationarity2}) as
	\begin{subequations}
		\label{helper: kkt}
		\begin{align}
			\frac{\lambda}{\phi} p_i & = \sum_{j\in\I^c} \mu_{i,j}\frac{p_i d(\theta_i, \bar{\theta}_{i,j})}{C_{i,j}(\bm p)} = \sum_j \mu_{i,j} h_{i,j},\\
			\frac{\lambda}{\phi} p_j & = \sum_{i\in\I} \mu_{i,j}\frac{p_j d(\theta_j, \bar{\theta}_{i,j})}{C_{i,j}(\bm p)} = \sum_i \mu_{i,j} h_{j,i}.
		\end{align}
	\end{subequations}
	Summing the above over all $i\in\I$ and $j\in\I^c$ and using the fact that $h_{i,j} + h_{j,i} = 1$, we have $\lambda = \phi$. 
	Plugging $\lambda = \phi$ into (\ref{helper: kkt}), we obtain (\ref{eq:KKT_equiv_stationarity}).
	
	Next we show that (\ref{eq:KKT_equiv}) also implies (\ref{eq:KKT}). 
	Notice that \eqref{eq:KKT_stationarity0}, \eqref{eq:Cij}, \eqref{eq:dual_feasibility} and \eqref{eq:KKT_CS} holds automatically.
	Using (\ref{eq:selection_function}) and (\ref{eq:KKT_equiv_complementary_slackness}), we can rewrite (\ref{eq:KKT_equiv_stationarity}) as
	\begin{equation}
		\begin{aligned}
			p_i &= \sum_{j: \mu_{i,j}>0} \mu_{i,j}\frac{p_i d(\theta_i, \bar{\theta}_{i,j})}{C_{i,j}(\bm p)}  = \sum_{j: \mu_{i,j}>0} \mu_{i,j}\frac{p_i d(\theta_i, \bar{\theta}_{i,j})}{\phi},\\
			p_j &= \sum_{i: \mu_{i,j}>0} \mu_{i,j}\frac{p_j d(\theta_j, \bar{\theta}_{i,j})}{C_{i,j}(\bm p)}  = \sum_{i: \mu_{i,j}>0} \mu_{i,j}\frac{p_j d(\theta_j, \bar{\theta}_{i,j})}{\phi}.
		\end{aligned}
	\end{equation}
	Thus, we have
	\begin{equation}
		\phi = \sum_{j\in \I^c} \mu_{i,j} d(\theta_i, \bar{\theta}_{i,j}) = \sum_{i\in \I} \mu_{i,j} d(\theta_j, \bar{\theta}_{i,j}).
	\end{equation}
	Let $\lambda = \phi$, we have 
	\begin{align*}
		\lambda -  \sum_{j\in \I^c} \mu_{i,j} d(\theta_i, \bar{\theta}_{i,j}) &= 0, \text{for all } i\in\I,\\
		\lambda -  \sum_{i\in \I} \mu_{i,j} d(\theta_j, \bar{\theta}_{i,j}) &= 0, \text{for all } j\in\I^c,
	\end{align*}
	which coincide with (\ref{eq:KKT_stationarity1}) and (\ref{eq:KKT_stationarity2}). 
	At last, (\ref{eq:primal_simplex}) can be seen from directly summing (\ref{eq:KKT_equiv_stationarity}) over all $i\in\I$ and $j\in\I^c$ and noting that $\bm\mu \in \mathcal{S}_{k(K-k)}$.
	This completes the proof. 
\end{proof}

\newpage
\section{Stopping rule for best-\texorpdfstring{$k$}{k}-arm identification}\label{app:stopping}

We consider the Chernoff stopping rule from the perspective of generalized hypothesis testing. 
Recall that at the end of each round $t$, the information available to the player is $\mathcal{H}_t$. If $\tau_{\delta} = t$, conditional on $\mathcal{H}_t$, the player recommend top-$k$ arm set $\hat{\I}_t$ as 
\[
\hat{\mathcal I}_t = \I(\bm\theta_t) = \argmax_{\mathcal{I} \subset [K], |\mathcal I| = k}\left\{\sum_{i \in \mathcal I} \theta_{t,i}\right\}.
\]
Define the rejection region as
\begin{equation}
	R_{t} = \left\{\bm \theta\in\Theta: \min_{i\in\hat{\I}_t} \theta_i < \max_{j\notin\hat{\I}_t}\theta_j \right\}.
\end{equation}
Then the generalized likelihood ratio test statistic is given by
\begin{equation}
	Z_t  = \log\frac{\sup_{\bm \theta}L_t(\bm \theta)}{\sup_{\bm \theta'\in R_t}L_t(\bm \theta')},
\end{equation}
where the likelihood is given by $L_t(\bm \theta)  = \prod_{i=1}^K \exp\left\{N_{t,i} \eta(\theta_i) \theta_{t,i} - N_{t,i}A(\eta(\theta_i))  \right\}$.
To ensure the $\delta$-correctness, we only need to identify a threshold $\gamma(t, \delta)$ such that, 
\begin{equation}\label{threshold beta}
	\mathbb{P}\left(Z_t \geq \gamma(t, \delta)\right) \leq \delta.
\end{equation}

To this end, we show a convenient closed-form formula for $Z_t$. First, 
it is clear that
\begin{equation}
	\begin{aligned}
		\sup_{\bm \theta}L_t(\bm \theta) 
		&= \sup_{\bm \theta}  \prod_{i=1}^K \exp\left\{N_{t,i} \eta(\theta_i) \theta_{t,i} - N_{t,i}A(\eta(\theta_i))  \right\} \\  
		&= \prod_{i=1}^K \exp\left\{N_{t,i} \eta(\theta_{t,i}) \theta_{t,i} - N_{t,i}A(\eta(\theta_{t, i}))  \right\}.
	\end{aligned}
\end{equation}
Then $Z_t$ can be simplified as 
\begin{equation} \label{eq:GLR}
	\begin{aligned}
		Z_t 
		&=\log\frac{\sup_{\bm \theta}  \prod_{i=1}^K \exp\left\{N_{t,i} \eta(\theta_i) \theta_{t,i} - N_{t,i}A(\eta(\theta_i))  \right\}   }{\sup_{\bm \theta'\in R_t} \prod_{i=1}^K \exp\left\{N_{t,i} \eta(\theta'_i) \theta_{t,i} - N_{t,i}A(\eta(\theta'_i))  \right\}    }\\
		&= -\log \sup_{\bm \theta'\in R_t} \prod_{i=1}^K \exp \left\{
		-N_{t,i} \left(
		A(\eta(\theta'_i)) -A(\eta(\theta_{t,i}))  - \theta_{t,i} (\eta(\theta'_i)  - \eta(\theta_{t,i}) 
		\right)
		\right\}\\
		&=-\log \sup_{\bm \theta'\in R_t} \exp \left\{-\sum_{i=1}^K N_{t,i} d(\theta_{t,i}, \theta'_i) \right\}\\
		&=\inf_{\bm \theta'\in R_t} \sum_{i=1}^K N_{t,i} d(\theta_{t,i}, \theta'_i)\\
		&= \min_{(i, j)\in \hat\I_t\times\hat\I_t^c} 
		N_{t,i} d\left(\theta_{t,i}, \frac{N_{t,i}\theta_{t,i}+N_{t,j}\theta_{t,j}}{N_{t,i}+N_{t,j}}\right)+N_{t,j} d\left(\theta_{t,j}, \frac{N_{t,i}\theta_{t,i}+N_{t,j}\theta_{t,j}}{N_{t,i}+N_{t,j}}\right) \\
		& = \min_{(i, j)\in \hat\I_t\times\hat\I_t^c}t \cdot C_{t, i,j}.
	\end{aligned}
\end{equation}

Thus, for the fixed-confidence setting, at the end of each round $t$, the algorithm compute $Z_t$. If $Z_t > \gamma_{t, \delta}$, then it stops and recommend $\hat{\I}_t$.

\newpage
\section{Facts about the exponential family and properties of \texorpdfstring{$C_{i,j}(\cdot)$}{the generalized Chernoff information}}\label{app:facts}

We restrict our attention to reward distribution following an one-dimensional natural exponential family. In particular, a natural exponential family has identity natural statistic in canonical form, i.e.,
\begin{equation}
	p(y|\eta) = b(y) \exp\{\eta y - A(\eta)\}, \quad \eta\in \mathcal{T}=\left\{\eta: \int b(y)e^{\eta y}\mathrm{d}y < \infty\right\}
\end{equation}
where $\eta$ is the natural parameter and $A(\eta)$ is assumed to be twice differentiable. The distribution is called non-singular if $\text{Var}_{\eta}(Y_{t, i})>0 $ for all $\eta\in \mathcal{T}^o$, where $\mathcal{T}^o$ denotes the interior of the parameter space $\mathcal{T}$. Denote by $\theta$ the mean reward  $\theta(\eta)=\int y p(y|\eta) dy$. 
By the properties of the exponential family, we have, for any $\eta \in \mathcal{T}^o$, 
\[
\theta(\eta) = \mathbb{E}_\eta[Y_{t, i}] = A'(\eta), \quad\text{and}\quad 
A''(\eta)=\text{Var}_\eta(Y_{t,i}) > 0.
\]
We immediately have the following lemma.
\begin{lemma}\label{lm:theta_eta}
	At any $\eta\in \mathcal{T}^o$, 
	$\theta(\eta)$ is strictly increasing and
	$A(\eta)$ is strictly convex. 
\end{lemma}

Lemma \ref{lm:theta_eta} implies that $\theta(\cdot)$ is a one-to-one mapping.
Consequently, we can use $\theta$ to denote the distribution $p(\cdot|\eta)$ for convenience. 
Let $\theta_1$ and $\theta_2$ be the mean reward of the distribution $p(\cdot|\eta_1)$ and $p(\cdot|\eta_2)$. 
Denote by $d(\theta_1, \theta_2)$ the  Kullback–Leibler (KL) divergence from $p(\cdot|\eta_1)$ to $p(\cdot|\eta_2)$. We have convenient closed-form expressions of $d(\cdot, \cdot)$ and its partial derivative.
\begin{lemma}
	The partial derivatives of the KL divergense of a one-dimensional natural exponential family satisfies
	\begin{equation}
		\label{kl: exponential family}
		d(\theta_1, \theta_2) = A(\eta_2) - A(\eta_1)  - \theta_1 (\eta_2 - \eta_1),
	\end{equation}
	\begin{equation}
		\label{kl: derivative theta}
		\frac{\partial d(\theta_1, \theta_2)}{\partial \theta_1} = \eta_1 - \eta_2, \quad 
		\frac{\partial d(\theta_1, \theta_2)}{\partial \theta_2} = (\theta_2 - \theta_1) \frac{\mathrm{d} \eta(\theta_2)}{\mathrm{d} \theta_2}.
	\end{equation}
\end{lemma}

\subsection{Properties of \texorpdfstring{$C_{i,j}(\cdot)$}{the generalized Chernoff information}}

\begin{lemma}\label{lm:C_ij_derivative}
	For any $(i,j)\in\I\times\I^c$, $C_{i,j}(\bm p)$ is concave in $\bm p$, strictly increasing in $p_i$ and $p_j$, and
	\[
	\frac{\partial C_{i,j}(\bm p)}{\partial p_i} = d(\theta_i, \bar{\theta}_{i,j})
	\quad\text{and}\quad 
	\frac{\partial C_{i,j}(\bm p)}{\partial p_j} = d(\theta_j, \bar{\theta}_{i,j})
	\]
\end{lemma}
where
$
\bar{\theta}_{i,j} = \bar{\theta}_{i,j}(\bm p) = \frac{p_i \theta_i + p_j \theta_j}{p_i + p_j}.
$
\begin{proof}
	To see the concavity of $C_{i,j}(\bm p)$, note that $C_{i,j}(\bm p) = \inf_{\tilde{\theta}} \left\{
	p_i d(\theta_i, \tilde{\theta})  +p_j d(\theta_j, \tilde{\theta}) \right\}$, which is the minimum over a family of linear functions in $\bm p$. This directly implies that $C_{i,j}(\bm p)$ is concave in $\bm p$ (See Chapter 3.2 of \cite{boyd2004convex}).
	
	The partial derivatives of $C_{i,j}(\bm p)$ are derived as follows
	\begin{equation*}
		\begin{aligned}
			\frac{\partial C_{i,j}(\bm p)}{\partial p_i} 
			&= d(\theta_i, \bar{\theta}_{i,j}) + p_i   \frac{\partial d(\theta_i, \bar{\theta})}{\partial p_i}  + p_j   \frac{\partial d(\theta_j, \bar{\theta})}{\partial p_i}\\
			&=d(\theta_i, \bar{\theta}_{i,j}) + \frac{\partial \bar{\theta}}{\partial p_i} \cdot\left[p_i  \frac{\partial d(\theta_i, \bar{\theta})}{\partial \bar{\theta}}  + p_j   \frac{\partial d(\theta_j, \bar{\theta})}{\partial \bar{\theta}}\right]\\
			&=d(\theta_i, \bar{\theta}_{i,j}).
		\end{aligned}
	\end{equation*}
	The last equality comes from (\ref{min C derivative}). 
	The calculation of $\frac{\partial C_{i,j}(\bm p)}{\partial p_j}$ follows similarly. \end{proof}

\begin{lemma}\label{lmm:contour}
	For any $(i,j)\in\I\times\I^c$, the contour of $C_{i,j}(\cdot)$ does not contain any line segment.
\end{lemma}
\begin{proof}
	To prove the result, we need to basic facts.
	\begin{enumerate}
		\item We first give the Hessian matrix of $C_{i,j}(\bm p)$ as follows
		\begin{equation*}
			H_{i,j} = 
			\begin{bmatrix}
				\frac{\partial^2C_{i,j}(\bm p)}{\partial^2 p_i} & \frac{\partial^2C_{i,j}(\bm p)}{\partial p_i \partial p_j} \\
				\frac{\partial^2C_{i,j}(\bm p)}{\partial p_j \partial p_i} & \frac{\partial^2C_{i,j}(\bm p)}{\partial^2 p_j} 
			\end{bmatrix}
			=
			\frac{(\theta_i - \theta_j)^2}{(p_i + p_j)^3}\frac{\mathrm{d} \eta(\bar{\theta}_{i,j})}{\mathrm{d}\theta}
			\begin{bmatrix}
				-p_j^2 & p_i p_j \\
				p_i p_j & - p_i^2
			\end{bmatrix},
		\end{equation*}
		where $\eta$ is the natural parameter of the exponential family; see Appendix \ref{app:facts}.
		By Lemma \ref{lm:theta_eta}, $\frac{\mathrm{d} \eta(\bar{\theta})}{\mathrm{d}\theta}$ is strictly positive. 
		Note that $H_{i,j}$ has an eigenvalue $0$.
		The other eigenvalue can be calculated correspondingly,
		\begin{equation*}
			\lambda_{\min }(H_{i,j}) = -\frac{p_i^2 + p_j^2}{(p_i + p_j)^3}\frac{\mathrm{d} \eta(\bar{\theta}_{i,j})}{\mathrm{d}\theta}(\theta_i - \theta_j)^2.
		\end{equation*}
		Note that $\lambda_{\min }(H_{i,j}) < 0$ whenever $p_i$ or $p_j$ is greater than $0$.
		\item Recall that $C_{i,j}$ satisfies the partial differential equation in \eqref{eq:PDE}, which has a general solution in the form of 
		\[C_{i,j}(\bm p) = p_i g_{i,j}\left(\frac{p_j}{p_i}\right),\]
		where $g_{i,j}(\cdot)$ is any differentiable function. This implies that the graph\footnote{The graph of a function $f(x)$ is defined as $G(f) = \{(x,f(x)): x \in D\}$, where $D$ is the domain of $f(x)$.} of $C_{i,j}(\bm p)$ is the boundary of a cone. In particular, for any point on the surface, the line segment connecting the point and the origin belongs to the graph of $C_{i,j}(\bm p)$.
	\end{enumerate}

	We prove the result by contradiction. Suppose that there exist a line segment in the contour of $C_{i,j}(p_i, p_j)$. Then the triangle specified by the two end points of the line segment and the origin belongs to the graph of $C_{i,j}$.
	This implies that at any point in the interior of this triangle, $C_{i,j}$ has a Hessian of $\bm 0$.
	This contradicts with the fact that the Hessian of $C_{i,j}$ at any point $(p_i,p_j)$ such that $p_i$ or $p_j$ is greater than $0$ is semi-negative definite, with a eigenvalue strictly smaller than $0$.
\end{proof}

\begin{lemma}
	\label{lm:gij}
	For any $(i,j)\in\I\times\I^c$, $C_{i,j}(\bm p)$ can be written as
	\begin{equation*}
		C_{i,j}(\bm p) = p_i g_{i,j}\left(\frac{p_j}{p_i}\right),
	\end{equation*}
	where 
	\begin{equation*}
		g_{i,j}(x) =d\left(\theta_i, \frac{\theta_i + x\theta_j}{1+x}\right) + x d\left(\theta_j, \frac{\theta_i + x\theta_j}{1+x}\right).
	\end{equation*}
	Furthermore, we have 
	\begin{equation*}
		g'_{i,j}(x) = d\left(\theta_j, \frac{\theta_i + x\theta_j}{1+x}\right) > 0.
	\end{equation*}
	Moreover, for any $i, j$ and $j'$ such that $\theta_i > \theta_{j'} \geq \theta_{j}$, we have 
	\begin{equation*}
		g_{i,j}(x) \geq g_{i,j'}(x) \quad\text{for all } x.
	\end{equation*}
\end{lemma}
\begin{proof}
	We start by identifying that 
	\begin{equation*}
		C_{i,j}(p_i, p_j) = p_i g_{i,j} \left(
		\frac{p_j}{p_i}\right).
	\end{equation*}
	Fix $p_i$ and let $x = \frac{p_j}{p_i}$,
	\begin{equation*}
		\frac{\partial C_{i,j}}{\partial p_j }  = p_i g_{i,j}'(x) \frac{\partial x}{\partial p_j} = g_{i,j}'(x).
	\end{equation*}
	By Lemma \ref{lm:C_ij_derivative}, we have 
	\begin{equation*}
		\frac{\partial C_{i,j}}{\partial p_j } = d \left(\theta_j, \frac{p_i \theta_i + p_j \theta_j}{p_i + p_j}\right) = d\left(\theta_j, \frac{\theta_i + x\theta_j}{1+x}\right).
	\end{equation*}
	Hence, 
	\begin{equation*}
		g_{i,j}'(x) = d\left(\theta_j, \frac{\theta_i + x\theta_j}{1+x}\right) > 0.
	\end{equation*}
	To this end, we fix $\theta_i$ and $x$ and define 
	\begin{equation*}
		h(\alpha) = d\left(\theta_i, \frac{\theta_i + x\alpha}{1+x}\right) + x d\left(\alpha, \frac{\theta_i + x\alpha}{1+x}\right).
	\end{equation*}
	It follows that $h(\theta_j) = g_{i,j}(x)$. To complete the proof, we need to show that, for any $\theta_i > \theta_{j'} \geq \theta_{j}$, $g_{i, j}(x) \geq g_{i, j'}(x)$. It suffices to verify that $h'(\alpha) < 0$ when $\theta_i > \alpha$.
	To see this, let 
	\[
	m = \frac{\theta_i + x\alpha }{1+x}.
	\]
	We recall the partial derivative of $d(\cdot, \cdot)$ in (\ref{kl: derivative theta}) and calculate $h'(\alpha)$ as follows.
	\begin{equation*}
		\begin{aligned}
			h'(\alpha)
			&= \frac{\partial d(\theta_i, m)}{\partial m} \frac{x}{1+x}  + x \left[
			\frac{\partial d(\alpha, m)}{\partial \alpha} + \frac{\partial d(\alpha, m)}{\partial m }\frac{x}{1+x}
			\right]\\
			& = (m - \theta_i) \frac{\mathrm{d} \eta}{\mathrm{d} m} \frac{x}{1+x} + x \left[
			\eta(\alpha) - \eta(m) + (m-\alpha) \frac{\mathrm{d} \eta}{\mathrm{d} m} \frac{x}{1+x}
			\right]\\
			& = x(\eta(\alpha) - \eta(m))  + ((m-\theta_i) + x (m-\alpha))\frac{\mathrm{d} \eta}{\mathrm{d} m} \frac{x}{1+x}\\
			& = x (\eta(\alpha) - \eta(m)).
		\end{aligned}
	\end{equation*}
	Noticing that $\eta'(\theta) = 1/\theta'(\eta) > 0$ and $m - \alpha = (\theta_i - \alpha) / (1+x) $, we have 
	\begin{equation}
		\label{helper: h monontonicty}
		h'(\alpha)
		\begin{cases}
			< 0, \quad\text{if }\theta_i > \alpha,\\
			> 0, \quad\text{if }\theta_i < \alpha.
		\end{cases}
	\end{equation}
	This completes the proof. 
\end{proof}

\newpage
\section{Existing algorithms}\label{app:existing_algs}

\paragraph{KL-LUCB \citep{kaufmann2013information}.} At each round $t$, the algorithm computes the upper confidence bound $U_i(t)$ and lower confidence bound $L_i(t)$ of the mean reward for each arm $i$:  	 	
\begin{equation}\label{KL-LUCB}
	\begin{aligned}
		U_i(t) &= \max\{q\in[\theta_{t,i}, \infty): T_{t,i} d(\theta_{t,i}, q) \leq \beta_{t, \delta}\},\\
		L_i(t) &= \min\{q\in(-\infty, \theta_{t,i}]: T_{t,i} d(\theta_{t,i}, q) \leq \beta_{t, \delta}\},
	\end{aligned}
\end{equation}
where $\beta_{t, \delta}$ is the exploration rate we specify to be $ \log \left((\log t + 1)/\delta\right)$. Then it samples two critical arms $u_t$ and $l_t$:
\[
u_t = \argmax_{i\notin \hat{\I}_t} U_i(t)\quad\text{and}\quad l_t = \argmin_{i\in  \hat{\I}_t} L_i(t),
\]
The algorithm stops when $U_{u_t}(t) > L_{l_t}(t)$.

\paragraph{UGapE \citep{gabillon2012best}.}  At each round $t$, the algorithm first compute the same $U_i(t)$ and $L_i(t)$ as (\ref{KL-LUCB}). Then it computes the gap index $B_i(t)=\max_{i'\neq i}^k U_{i'}(t) - L_i(t)$ for each arm $i$, where $\max_{}^k$ denotes the operator returning the $k$-th largest value. Based on $B_i(t)$, it evaluates a set of $k$ arms: $J(t) = \{j:\argmin^j B_j(t)\}$. From $J(t)$, it computes two critical arms $u_t$ and $l_t$:
\[
u_t = \argmax_{i\notin J_t} U_i(t)\quad\text{and}\quad l_t = \argmin_{i\in  J_t} L_i(t),
\]  and samples one with larger $\beta_{t, \cdot}$. The algorithm stops when $\max_{i\in J(t)} B_i(t) > 0$.

\paragraph{$m$-LinGapE \citep{reda2021top}.} At each round $t$, the algorithm first compute the same $U_i(t)$ and $L_i(t)$ as (\ref{KL-LUCB}). Then it computes the gap index $B_{i, j}(t)=U_{i}(t) - L_j(t)$ for each arm $i$. It choose $J(t)$ as the set of $k$ arms with largest empirical means. From $J(t)$, it computes two critical arms $u_t$ and $l_t$:
\[
u_t = \argmax_{j\in J_t} \max_{i\notin J_t} B_{i,j}(t)\quad\text{and}\quad l_t = \argmax_{i\notin  J_t} B_{i, u_t}(t),
\]  and samples one which would reduce the variance of the empirical mean difference  between them. (This is so-called "greedy selection rule" in \cite{reda2021top}.)  The algorithm stops when 
\[B_{l_t, u_t}(t) > 0.\] 

We remark that, in \cite{reda2021top}, the authors introduce a meta algorithm called ``Gap-Index-Focused" Algorithm (GIFA), which includes KL-LUCB, UGapE and $m$-LinGapE in its framework. Refer to Algorithm 1 and Table 1 in \cite{reda2021top}. 

\paragraph{MisLid \citep{reda2021dealing}.} The algorithm is designed from the lower bound maximin problem and achieves the asymptotic optimality.  It maintains a learner over the probability simplex of number of arms. At each round $t$,  it computes a gain function and updates the learner. Then it perform tracking-rule to sample an arm. We adopt AdaHedge as the learner and use empirical gain function $g_t$ as 
\[
\argmin_{\bm \lambda \in \text{Alt}(\bm \mu^t)} \sum_{i=1}^K \bm p^t_i (\bm \mu^t_i - \bm \lambda)^2  
\]
where $\bm p^t$ is probability distribution returned by AdaHedge, $\bm \mu_t$ is the empirical mean estimate and $\text{Alt}(\bm \mu^t)$ defines the set of bandit model returning the top-$k$ arm set different from $(\bm \mu^t)$. As suggested by \cite{reda2021dealing}, the empirical gain function reduced the sample complexity practically compared with the optimistic gain function suggested by theory.	 For the tracking rule, we adopt C-Tracking in our experiments. For the stopping rule, we adopt the same General Likelihood Ratio Test statistic with threshold $\log \left((\log (t) + 1)/\delta\right)$.

\paragraph{FWS \citep{wang2021fast}.} The algorithm track the lower bound maximin problem. At each round $t$, it performs forced exploration or Frank-Wolfe update. Due to the non-smoothness of objective function, it construct an $r_t$-sub-differential subspaces and then solve a zero-sum game. Note $r_t$ is a tuning parameter. We choose $r_t = t^{-0.9}/K$ as suggested in \cite{wang2021fast}. For the stopping rule, we adopt the same General Likelihood Ratio Test statistic with threshold $\log \left((\log(t) + 1)/\delta\right)$.

\paragraph{Lazy-Mirror-Ascent \citep{menard2019gradient}. } The algorithm treats the lower bound problem as \[\max_{\bm w} F(\bm w).\] It uses a sampling rule based on an online lazy mirror ascent. At each round $t$, it tracks the allocation weights $\bm w(t)$ and updates it by
\[
\tilde{\bm w}(t + 1)  = \argmax_{\bm w} \eta_{t+1} \sum_{s=K}^t\bm w \cdot \nabla  F(\tilde{\bm w}(s); \bm \mu(s)) - \text{kl}(\bm w, \bm \pi);
\] 
\[
{\bm w}(t + 1) = (1- \gamma_t) \tilde{\bm w}(t + 1) + \gamma_t \bm \pi ,
\]
where $\bm \mu(s)$ is the empirical mean estimates and $\bm \pi$ is the uniform distribution. There exists two tuning parameter $\eta_t$ and $\gamma_t$.  Based on the code released by Wouter M. Koolen\footnote{\url{https://bitbucket.org/wmkoolen/tidnabbil/src/master/}}
and the suggestion in \cite{menard2019gradient},  we choose $\eta_t = {\sqrt{\log K/t}}/{F^*}$ and $\gamma_t = 1/(4\sqrt{t})$. Wouter M. Koolen comments that, although setting $F^*$ relies on the knowledge of true arm means, it is critical for the performance. In appendix of \cite{wang2021fast}, the authors also comment that this method is very sensitive to the learning rate $\eta_t$, the current choice is only for experimental comparison and cannot be used in real-world scenarios.

\end{document}